# A Likelihood Ratio Framework for High Dimensional Semiparametric Regression

Yang Ning*   Tianqi Zhao†   Han Liu‡

**Abstract**

We propose a likelihood ratio based inferential framework for high dimensional semiparametric generalized linear models. This framework addresses a variety of challenging problems in high dimensional data analysis, including incomplete data, selection bias, and heterogeneous multitask learning. Our work has three main contributions. (i) We develop a regularized statistical chromatography approach to infer the parameter of interest under the proposed semiparametric generalized linear model without the need of estimating the unknown base measure function. (ii) We propose a new framework to construct post-regularization confidence regions and tests for the low dimensional components of high dimensional parameters. Unlike existing post-regularization inferential methods, our approach is based on a novel directional likelihood. In particular, the framework naturally handles generic regularized estimators with nonconvex penalty functions and it can be used to infer least false parameters under misspecified models. (iii) We develop new concentration inequalities and normal approximation results for U-statistics with unbounded kernels, which are of independent interest. We demonstrate the consequences of the general theory by using an example of missing data problem. Extensive simulation studies and real data analysis are provided to illustrate our proposed approach.

## 1 Introduction

Modern data are characterized by their high dimensionality, complexity and heterogeneity (Fan et al., 2014). More specifically, the datasets usually contain (1) a large number of explanatory variables, (2) complex sampling and missing value schemes due to design or incapability of contacting study subjects, and (3) heterogeneity due to the combination of different data sources.

In the literature, regularization based procedures are commonly used to handle data with high dimensionality. For instance, the $L_1$-regularized maximum likelihood estimation for linear models is proposed by Tibshirani (1996) and the nonconvex penalized maximum likelihood estimation is considered by Fan and Li (2001). During the past decades, the regularization based methods enjoy great success in handling high dimensional data. However, the existing framework is not flexible enough to handle more challenging settings with incomplete data, complex sampling, and multiple datasets with heterogeneity. To motivate our study, consider the following two examples.

---
*Department of Operations Research and Financial Engineering, Princeton University, Princeton, NJ 08544, USA; e-mail: yning@princeton.edu.
†Department of Operations Research and Financial Engineering, Princeton University, Princeton, NJ 08544, USA; e-mail: tianqi@princeton.edu.
‡Department of Operations Research and Financial Engineering, Princeton University, Princeton, NJ 08544, USA; e-mail: hanliu@princeton.edu.

**Example 1: Missing Data and Selection Bias.** Given a univariate random variable $Y$ and a $d$ dimensional random vector $\boldsymbol{X}$, assume that $Y$ given $\boldsymbol{X}$ follows from a generalized linear model with the canonical link,

$$p(y \mid \boldsymbol{x}) = \exp\left\{\boldsymbol{x}^T\boldsymbol{\beta} \cdot y - b(\boldsymbol{x}^T\boldsymbol{\beta}, f) + \log f(y)\right\}, \tag{1.1}$$

where $\boldsymbol{\beta}$ is a $d$-dimensional unknown parameter, $f(\cdot)$ is a known base measure function and $b(\cdot,\cdot)$ is a normalizing function. Let $(Y_1, \boldsymbol{X}_1), ..., (Y_n, \boldsymbol{X}_n)$ denote $n$ independent copies of $(Y, \boldsymbol{X})$. In high dimensional data analysis, the samples $(Y_1, \boldsymbol{X}_1), ..., (Y_n, \boldsymbol{X}_n)$ often contain missing values or they are observed after some unknown complex selection process. To account for the effect of missingness or selection bias, we introduce an indicator variable $\delta_i$, whose value is 1 if $(Y_i, \boldsymbol{X}_i)$ is completely observed or selected, and 0 otherwise. Due to the selection effect, the standard penalized maximum likelihood estimator under model (1.1) with only selected data (i.e., $\delta_i = 1$) is often inconsistent for $\boldsymbol{\beta}$. To account for the missing data and selection bias, we need to develop a new framework to infer the high dimensional parameter $\boldsymbol{\beta}$.

**Example 2: Multitask Learning with Heterogeneity.** Modern datasets are often collected by aggregating multiple data sources. Analysis of such types of data has been studied in the fields of multitask learning in machine learning (Argyriou et al., 2008; Maurer, 2006; Obozinski et al., 2011; Lounici et al., 2011) and seemingly unrelated regression in econometrics (Srivastava and Giles, 1987). In the multitask learning setting, each dataset corresponds to a learning task. More specifically, assume that the data in the $t$th task, $t = 1, ..., T$ are i.i.d. copies of $(Y_{(t)}, \boldsymbol{X}_{(t)})$, which follows from (1.1), i.e.,

$$p(y_{(t)} \mid \boldsymbol{x}_{(t)}) = \exp\left\{\boldsymbol{x}_{(t)}^T\boldsymbol{\beta}_t \cdot y_{(t)} - b(\boldsymbol{x}_{(t)}^T\boldsymbol{\beta}_t, f_t) + \log f_t(y_{(t)})\right\}, \tag{1.2}$$

where $\boldsymbol{\beta}_t$ is a task-specific regression parameter. Most of the existing literature only focuses on the homogeneous tasks, that means $f_t(\cdot) = f(\cdot)$ for any $t = 1, ..., T$. However, the aggregated data are often highly heterogeneous. For instance, the learning tasks obtained from different areas may contain classification for binary responses as well as regression for continuous and count responses, which implies different forms of $f_t(\cdot)$ in (1.2). Thus, to take into account of data heterogeneity, we need a new inferential procedure for $\boldsymbol{\beta}_t$ that does not depend on the knowledge of $f_t(\cdot)$.

To handle the challenges raised above, we propose a new semiparametric model, which takes the form (1.1) but with both $\boldsymbol{\beta}$ and $f(\cdot)$ as unknown parameters. It naturally handles data with missing values, complex sampling and heterogeneity. This paper contains the following three major contributions.

Our first contribution is to provide a new statistical chromatography procedure to directly estimate the finite dimensional regression parameter $\boldsymbol{\beta}$ and leave the nonparametric component $f(\cdot)$ as a nuisance. In particular, we model the data at a more refined granularity of rank and order statistics, so that sophisticated conditioning arguments can be applied to separate the parameter of interest and nuisance component. This step is called the statistical chromatography. Once the parameter of interest and nuisance parameter are separated, we eliminate the nuisance component to construct a pseudo-likelihood of rank statistics and exploit lower order approximation to speed up computation.

Our second contribution is to develop a new inferential framework for parameters under the high dimensional semiparametric model. In particular, we propose a directional likelihood ratio statistic for hypothesis testing and a maximum directional likelihood estimator for confidence regions in the high dimensional regime. Compared to the existing post-regularization inferential methods, our procedure has three important features: (1) we allow general regularized estimators including nonconvex regularized estimators; (2) we do not need any signal strength assumption for model selection consistency; and (3) our framework tolerates possibly misspecified model.

Our third contribution is to develop some new technical tools for studying the high dimensional inference related to U-statistics. First, we prove a general concentration inequality in Lemma 5.3 for U-statistics with unbounded kernels. Second, to apply the central limit theorem for U-statistics, we provide the theoretical justification of the H$á$jek projection in increasing dimensions for normal approximation. More details are provided in Lemma 5.5.

**Comparison with Related Works:** The proposed model is closely related to the proportional likelihood ratio model in Luo and Tsai (2012); Chan (2012). However, unlike their model we do not require the density assumption for the nonparametric function, and we focus on the theoretical properties in high dimensional regimes, which have not been studied before. The proposed estimation procedure is related to the permutation based test (Kalbfleisch, 1978) and can be also interpreted as the composite likelihood method (Lindsay, 1988; Besag, 1974; Varin et al., 2011), but we motivate our method from a different point of view. To handle high dimensional data with missing values, Städler and Bühlmann (2012) proposed an expectation-maximization algorithm. When the explanatory variables are missing completely at random (MCAR), Loh and Wainwright (2012) developed the theory of a nonconvex optimization approach. Compared with the assumptions in these works, we consider a much broader class of missing data mechanisms.

In the linear and generalized linear models, the estimation, prediction error bounds and variable selection consistency for the $L_1$-regularized estimator have been well studied by Bickel et al. (2009); Bunea et al. (2007); van de Geer (2008); Zhang (2009); Meinshausen and Yu (2009); Meinshausen and Bühlmann (2006); Zhao and Yu (2006); Wainwright (2009). More recently, the estimation bounds and oracle properties for the nonconvex regularized estimator are established by Zhang (2010); Fan et al. (2012); Loh and Wainwright (2013); Wang et al. (2013b) and Wang et al. (2013a), among others. In addition to these estimation results, significant progress has been made towards understanding the post-regularization inference (e.g., constructing confidence intervals or testing hypotheses) under the generalized linear models. For instance, Zhang and Zhang (2014) proposed a novel low dimensional projection method and laid down the theoretical framework for inference in high dimension linear models. Some alternative procedures are proposed by van de Geer et al. (2014); Javanmard and Montanari (2013); Belloni et al. (2013, 2012). All these procedures lead to asymptotically normally distributed estimators that can be used to construct Wald type statistics. Other related inferential procedures include the data-splitting method (Wasserman and Roeder, 2009; Meinshausen et al., 2009), stability selection (Meinshausen and Bühlmann, 2010; Shah and Samworth, 2013), $L_2$ confidence set (Nickl and van de Geer, 2013) and the post selection inference (Lockhart et al., 2014; Taylor et al., 2014). Under the oracle property, the asymptotic normality of nonconvex regularized estimators has been established by Fan and Li (2001); Bradic et al. (2011).

This paper proposes a directional likelihood based method for constructing confidence regions and testing hypotheses in high dimensions. Compared to the existing work (Zhang and Zhang, 2014; van de Geer et al., 2014; Javanmard and Montanari, 2013), our method and theory are different in the following four aspects. First, the proposed semiparametric model is much more sophisticated than the generalized linear model. The U-statistic structure due to the statistical chromatography leads to additional technical complexity (see the third contribution above). Therefore, it requires more refined analysis to control the variability of the estimated nuisance parameters in the proposed likelihood function. Second, we consider the inference on local solutions of a nonconvex regularized problem. However, the method in van de Geer et al. (2014) based on inverting the Karush-Kuhn-Tucker condition cannot be directly applied to nonconvex problems. Third, from the hypothesis testing perspective, our main inferential tool is a new directional likelihood ratio test, whereas the existing methods only study the Wald type tests. Fourth, under misspecified models, due to the likelihood based interpretation, our inference framework can be used to infer the least false parameter. This type of results extends the classical results for misspecified models in White (1982) and is new for high dimensional models.

This paper is organized as follows. In Section 2, we formally define the proposed semiparametric model. In Section 3, we introduce the main ideas of statistical chromatography, along with the directional likelihood based inference for hypothesis tests and confidence regions. In Section 4, we establish the asymptotic distributions of the Wald and directional likelihood ratio test statistics. In Section 5, we present the proofs of main results. In Section 6, the extensions to missing data and selection bias problems are considered. Section 7 contains both simulation and real data analysis results.

**Notation:** For positive sequences $a_n$ and $b_n$, we write $a_n \lesssim b_n$, if $a_n/b_n = O(1)$. We denote $a_n \asymp b_n$ if $a_n \lesssim b_n$ and $b_n \lesssim a_n$. Denote $X_n \rightsquigarrow X$ for some random variable $X$ if $X_n$ converges weakly to $X$. For $\mathbf{v} = (v_1, ..., v_d)^T \in \mathbb{R}^d$, and $1 \leq q \leq \infty$, we define $||\mathbf{v}||_q = (\sum_{i=1}^d |v_i|^q)^{1/q}$, $||\mathbf{v}||_0 = |\text{supp}(\mathbf{v})|$, where $\text{supp}(\mathbf{v}) = \{j : v_j \neq 0\}$ and $|A|$ is the cardinality of a set $A$. Denote $||\mathbf{v}||_\infty = \max_{1 \leq i \leq d} |v_i|$ and $\mathbf{v}^{\otimes 2} = \mathbf{v}\mathbf{v}^T$. For a matrix $\mathbf{M}$, let $||\mathbf{M}||_2, ||\mathbf{M}||_\infty, ||\mathbf{M}||_1$ and $||\mathbf{M}||_{L_1}$ be the spectral, elementwise supreme, elementwise $L_1$ and matrix $L_1$ norms of $\mathbf{M}$. For two matrices $\mathbf{M}_1$ and $\mathbf{M}_2$, we write $\mathbf{M}_1 \leq \mathbf{M}_2$ if $\mathbf{M}_2 - \mathbf{M}_1$ is positive semi-definite. Denote $\mathbb{B}_0(s) = \{\boldsymbol{\beta} \in \mathbb{R}^d : ||\boldsymbol{\beta}||_0 \leq s\}$ to be the $L_0$-ball with radius $s$, and $\mathbb{B}_q(R) = \{\boldsymbol{\beta} \in \mathbb{R}^d : ||\boldsymbol{\beta}||_q \leq R\}$ to be the $L_q$ ball with radius $R$. For $S \subseteq \{1, ..., d\}$, let $\mathbf{v}_S = \{v_j : j \in S\}$ and $S^c$ be the complement of $S$. The gradient and subgradient of a function $f(\boldsymbol{x})$ are denoted by $\nabla f(\boldsymbol{x})$ and $\partial f(\boldsymbol{x})$ respectively. For a univariate function $f(x)$, its derivative can also be represented by $f'(x)$. Let $\nabla_S f(\boldsymbol{x})$ denote the gradient of $f(\boldsymbol{x})$ with respect to $\boldsymbol{x}_S$. Let $\mathbf{I}_d$ be the $d$ by $d$ identity matrix. Let $\lfloor k \rfloor$ denote the largest integer less than $k$.

## 2 The Semiparametric Generalized Linear Model

To introduce the proposed model, we first define a semiparametric natural exponential family model. We then introduce the semiparametric generalized linear model.

**Definition 2.1 (Semiparametric natural exponential family).** A random variable $Y \in \mathcal{Y} \subseteq \mathbb{R}$ satisfies the semiparametric natural exponential family (spEF) with parameters $(\theta, f)$, if its density satisfies

$$p(y; \theta, f) = \exp\left\{\theta \cdot y - b(\theta, f) + \log f(y)\right\}, \tag{2.1}$$

where $f(\cdot)$ is an unknown base measure, $\theta$ is an unknown canonical parameter, and

$$b(\theta, f) = \log \int_{\mathcal{Y}} \exp\left(\theta \cdot y\right) \cdot f(y) \cdot dy < \infty$$

is the log-partition function.

The spEF extends the classical natural exponential family by treating the base measure $f(y)$ as an infinite dimensional parameter. By choosing a suitable base measure, the spEF recovers the whole class of natural exponential family distributions. However, the spEF suffers from the identifiability issue. For instance, $\text{spEF}(\theta, f)$ is identical to $\text{spEF}(\theta, c \cdot f)$, where $c$ is any positive constant. To address this problem, we need to impose some identifiability conditions, such as $f(y_0) = 1$, for some $y_0 \in \mathcal{Y}$, or $\int_{\mathcal{Y}} f(y) \cdot dy = 1$ if $f(y)$ is integrable. Later, we can see that these identifiability conditions will not affect our inference procedures. We now define the semiparametric generalized linear model.

**Definition 2.2** (**Semiparametric generalized linear model**). Given a vector of $d$-dimensional covariates $\boldsymbol{X} = (X_1, ..., X_d)^T$ and response $Y \in \mathbb{R}$, assume that $Y$ given $\boldsymbol{X}$ follows the semiparametric natural exponential family

$$p(y \mid \boldsymbol{x}) = \exp\left\{\theta(\boldsymbol{x}) \cdot y - b(\theta(\boldsymbol{x}), f) + \log f(y)\right\}, \quad \text{and} \quad \theta(\boldsymbol{x}) = \boldsymbol{\beta}^T \boldsymbol{x}, \tag{2.2}$$

where $b(\cdot, \cdot)$ is the log-partition function and $\boldsymbol{\beta}$ is a $d$-dimensional parameter. We say that $Y$ given $\boldsymbol{X}$ follows the semiparametric generalized linear model with parameters $(\boldsymbol{\beta}, f)$.

Note that, we directly set $\theta(\boldsymbol{x}) = \boldsymbol{\beta}^T \boldsymbol{x}$ in (2.2), because we implicitly adopt the canonical link, i.e., we choose a link function $g$ such that $g^{-1}(\cdot) = b'(\cdot, f)$. Compared with the generalized linear models (GLMs), the proposed model contains unknown parameters $\boldsymbol{\beta}$ and $f(\cdot)$, where $\boldsymbol{\beta}$ characterizes the covariate effect, and $f(\cdot)$ determines the distribution in the natural exponential family. For instance, the linear regression with standard Gaussian noise has $f(y) = \exp(-y^2/2)$; the logistic regression has $f(y) = 1$; and the Poisson regression has $f(y) = 1/y!$. Thus, these GLMs are parametric submodels of the semiparametric generalized linear model.

**Remark 2.1.** Some exponential family distributions, such as the normal distribution, involve dispersion parameters. In this case, the semiparametric natural exponential family can be written as

$$p(y; \boldsymbol{\theta}, \tau, f) = \exp\left\{\frac{\theta \cdot y - b(\theta, f)}{a(\tau)} + \log f(y; \tau)\right\},$$

where $f(\cdot; \cdot)$ is an unknown positive function, $\theta$ is the natural parameter, $a(\tau)$ is a known function of the dispersion parameter $\tau$ and $b(\theta, f)$ is the log-partition function. Then, with $\theta(\boldsymbol{x}) = \boldsymbol{\beta}^T \boldsymbol{x}$, the semiparametric generalized linear model reduces to

$$p(y \mid \boldsymbol{x}; \boldsymbol{\beta}, \tau, f) = \exp\left\{\bar{\boldsymbol{\beta}}^T \boldsymbol{x} \cdot y - \bar{b}(\bar{\boldsymbol{\beta}}^T \boldsymbol{x}, \tau, f) + \log f(y; \tau)\right\},$$

where $\bar{\boldsymbol{\beta}} = \boldsymbol{\beta}/a(\tau)$ and $\bar{b}(\bar{\boldsymbol{\beta}}^T \boldsymbol{x}, \tau, f) = b(a(\tau)\bar{\boldsymbol{\beta}}^T \boldsymbol{x}, f)/a(\tau)$. Hence, with the new reparametrization $\bar{\boldsymbol{\beta}}$, the proposed model is identical to (2.2), except that we allow $\bar{b}(\cdot)$ and $f(\cdot; \cdot)$ to depend on the dispersion parameter $\tau$. Later, we will see that this dependence does not lead to any extra level of difficulty in terms of inference on $\bar{\boldsymbol{\beta}}$.

The semiparametric generalized linear model has broad applicability to address the challenging problems involving complex and heterogeneous data. In the following, we illustrate how the semiparametric model can be used to handle the missing data and selection bias problems in Example 1 and heterogeneous multitask learning problem in Example 2.

## 2.1  Revisit of Example 1: Missing Data and Selection Bias.

Recall that $Y_i$ and $\boldsymbol{X}_i$ follow the GLM in (1.1) and we are interested in making inference on $\boldsymbol{\beta}$. To account for the missing data and selection effect, we assume that the selection indicator $\delta_i$ given $Y_i$ and $\boldsymbol{X}_i$ satisfies the following decomposable selection model.

**Definition 2.3 (Decomposable selection model).** The missing data or selection model is decomposable, if there exist two nonnegative functions $g_1(\cdot)$ and $g_2(\cdot)$ such that $\mathbb{P}(\delta_i = 1 \mid Y_i, \boldsymbol{X}_i) = g_1(Y_i) \cdot g_2(\boldsymbol{X}_i)$, where $\int g_1(y) \cdot dy = 1$ and $\int g_2(\boldsymbol{x}) \cdot d\boldsymbol{x} = 1$.

Under the assumption of MCAR, the missing data model satisfies $\mathbb{P}(\delta_i = 1|Y_i, \boldsymbol{X}_i) = \mathbb{P}(\delta_i = 1)$, which implies that MCAR is decomposable. Indeed, the decomposable model is much more general. Consider the following partition of covariates $\boldsymbol{X}_i = (\boldsymbol{X}_{io}, \boldsymbol{X}_{im})$, and assume that $(Y_i, \boldsymbol{X}_{im})$ are subject to missingness. It is seen that the missing at random (MAR) in Little and Rubin (1987), defined by $\mathbb{P}(\delta_i = 1|Y_i, \boldsymbol{X}_i) = \mathbb{P}(\delta_i = 1|\boldsymbol{X}_{io})$, is also decomposable. So is the outcome dependent sampling model in Lawless et al. (1999); Breslow et al. (2003). In addition, the decomposable model can be missing not at random (MNAR) (Little and Rubin, 1987). For instance, if $Y_i$ is subject to missingness and the missing mechanism only depends on the potentially unobserved value of $Y_i$, then the missing data pattern is not at random but is still decomposable. Thus, the decomposable selection model is a very flexible nonparametric model for missing data and selection bias. In general, the functions $g_1(\cdot)$ and $g_2(\cdot)$ in Definition 2.3 may not be identifiable. Later, we will see that this nonidentifiability issue can be handled by using the proposed method.

To specify the likelihood based on the selected data, we derive the probability density function of $Y_i$ given $\boldsymbol{X}_i$ and $\delta_i = 1$. Using the Bayes formula, we have

$$p(y_i \mid \boldsymbol{x}_i, \delta_i = 1) = \frac{1}{T_i(\boldsymbol{x}_i)} \cdot \mathbb{P}(\delta_i = 1 \mid y_i, \boldsymbol{x}_i) \cdot p(y_i \mid \boldsymbol{x}_i),$$

where $T_i(\boldsymbol{x}_i) = \int \mathbb{P}(\delta_i = 1 \mid y_i, \boldsymbol{x}_i) \cdot p(y_i \mid \boldsymbol{x}_i) \cdot dy_i$ and $(y_i, \boldsymbol{x}_i)$ is the observed value of $(Y_i, \boldsymbol{X}_i)$. Under the generalized linear model in (1.1) and the decomposable selection model, we obtain

$$p(y_i \mid \boldsymbol{x}_i, \delta_i = 1) = \exp\left\{\boldsymbol{x}_i^T \boldsymbol{\beta} \cdot y_i - b(\boldsymbol{x}_i^T \boldsymbol{\beta}, f^m) + \log f^m(y_i)\right\}, \tag{2.3}$$

where $f^m(y) = g_1(y) \cdot f(y)$. Hence, if $Y_i$ given $\boldsymbol{X}_i$ follows the GLM (1.1) or more generally the semiparametric version (2.2) and the selection model is decomposable, then $Y_i$ given $\boldsymbol{X}_i$ and $\delta_i = 1$ satisfies (2.2) with the same unknown parameter $\boldsymbol{\beta}$ and the unknown based measure $f^m(y) = g_1(y) \cdot f(y)$. We call this the invariance property of semiparametric GLMs under the decomposable selection model. Hence, the inference on $\boldsymbol{\beta}$ with missing data and selection bias can reduce to the inference problem under the semiparametric GLM (2.2).

## 2.2 Revisit of Example 2: Multitask Learning with Heterogeneity.

In Example 2 of Section 1, to take into account of data heterogeneity, we can assume that the based measure function $f_t(\cdot)$ is a task-specific unknown function. Thus, the multitask learning with heterogeneity can be handled by the semiparametric GLM framework, and an inferential method that is invariant to $f(\cdot)$ under the model (2.2) is needed.

# 3 Semiparametric Inference

In this section, we consider how to construct confidence intervals and perform hypothesis tests for some low dimensional component of $\boldsymbol{\beta}$ under the semiparametric GLM.

## 3.1 Regularized Statistical Chromatography

Due to the presence of the unknown function $f(\cdot)$, the likelihood of the semiparametric GLM is complicated, making likelihood based inference of $\boldsymbol{\beta}$ intractable. To handle this problem, we propose a new procedure called statistical chromatography to extract information on $\boldsymbol{\beta}$.

For $i = 1, ..., n$, suppose that the data $(Y_i, \boldsymbol{X}_i)$ are i.i.d. By the discriminative modeling approach, the probability distribution of the data is $p(\boldsymbol{y}, \mathbf{x}; \boldsymbol{\beta}, f) = p(\boldsymbol{y} \mid \mathbf{x}; \boldsymbol{\beta}, f) \cdot p(\mathbf{x})$, where $\boldsymbol{y} = (y_1, ..., y_n)$ and $\mathbf{x} = (\boldsymbol{x}_1, ..., \boldsymbol{x}_n)$ are the observed values of $\boldsymbol{Y} = (Y_1, ..., Y_n)$ and $\mathbf{X} = (\boldsymbol{X}_1, ..., \boldsymbol{X}_n)$. Since the marginal distribution of $\mathbf{X}$ does not involve $\boldsymbol{\beta}$ or $f$, we only focus on the first conditional distribution $p(\boldsymbol{y} \mid \mathbf{x}; \boldsymbol{\beta}, f)$. However, its dependence on $\boldsymbol{\beta}$ and $f$ is still intertwined and the inference on $\boldsymbol{\beta}$ is hindered by the nuisance parameter $f$. To tackle this problem, we need to further separate the parameters $\boldsymbol{\beta}$ and $f$ in the conditional likelihood. To this end, we decompose $\boldsymbol{Y} = (Y_1, ..., Y_n)$ into $\mathbf{R} = (R_1, ..., R_n)$ and $\boldsymbol{Y}_{(\cdot)} = (Y_{(1)}, ..., Y_{(n)})$, which denote the rank and order statistics of $\boldsymbol{Y}$, respectively. Let $\boldsymbol{r}$ and $\boldsymbol{y}_{(\cdot)}$ denote the observed values of $\mathbf{R}$ and $\boldsymbol{Y}_{(\cdot)}$, respectively. Thus, we have

$$p(\boldsymbol{y} \mid \mathbf{x}; \boldsymbol{\beta}, f) = \mathbb{P}(\mathbf{R} = \boldsymbol{r} \mid \mathbf{x}, \boldsymbol{y}_{(\cdot)}; \boldsymbol{\beta}) \cdot p(\boldsymbol{y}_{(\cdot)} \mid \mathbf{x}; \boldsymbol{\beta}, f), \tag{3.1}$$

where by the definition of conditional probabilities we can show that

$$\mathbb{P}(\mathbf{R} = \boldsymbol{r} \mid \mathbf{x}, \boldsymbol{y}_{(\cdot)}; \boldsymbol{\beta}) = \frac{\prod_{i=1}^n p(y_i \mid \boldsymbol{x}_i; \boldsymbol{\beta}, f)}{\sum_{\pi \in \Xi} \prod_{i=1}^n p(y_{\pi(i)} \mid \boldsymbol{x}_i; \boldsymbol{\beta}, f)} = \frac{\exp(\sum_{i=1}^n \boldsymbol{\beta}^T \boldsymbol{x}_i \cdot y_i)}{\sum_{\pi \in \Xi} \exp(\sum_{i=1}^n \boldsymbol{\beta}^T \boldsymbol{x}_i \cdot y_{\pi(i)})}, \tag{3.2}$$

where $\Xi$ is the set of all one-to-one maps from $\{1, ..., n\}$ to $\{1, ..., n\}$. The intuition behind the data decomposition is that the rank statistic given the order statistic has no information on $f$. Mathematically, the product $\prod_{i=1}^n f(y_i)$ appearing in both numerator and denominator of (3.2) only depends on $\boldsymbol{Y}_{(\cdot)}$ and is eliminated. Since we separate parameters $\boldsymbol{\beta}$ and $f$ at a more refined granularity of rank and order statistics, we call this procedure as statistical chromatography.

Given the chromatography decomposition in (3.1), one may opt to only keep the conditional probability (3.2) for estimation and inference of $\boldsymbol{\beta}$. However, the probability in (3.2) is computationally intensive due to the combinatorial nature of permutations. To this end, we consider a surrogate of $\mathbb{P}(\mathbf{R} = \boldsymbol{r} \mid \mathbf{x}, \boldsymbol{y}_{(\cdot)}; \boldsymbol{\beta})$ using the $k$th order information. For notational simplicity, we only present $k = 2$, and leave the discussion for $k > 2$ to the appendix. For any $i$ and $j$, let $\mathbf{R}_{ij}^L$ denote the local rank statistic of $Y_i$ and $Y_j$ among the pair $(Y_i, Y_j)$ (i.e., $\mathbf{R}_{ij}^L = (1, 2)$ or $(2, 1)$).

Instead of considering the full conditional probability in (3.2), we study the product of all possible combinations of the local rank conditional probability, i.e.,

$$\prod_{i<j} \mathbb{P}(\mathbf{R}_{ij}^L = r_{ij}^L \mid \boldsymbol{x}_i, \boldsymbol{x}_j, \boldsymbol{y}_{(i,j)}^L; \boldsymbol{\beta}) = \prod_{i<j} \frac{\exp(\boldsymbol{\beta}^T \boldsymbol{x}_i y_i + \boldsymbol{\beta}^T \boldsymbol{x}_j y_j)}{\exp(\boldsymbol{\beta}^T \boldsymbol{x}_i y_i + \boldsymbol{\beta}^T \boldsymbol{x}_j y_j) + \exp(\boldsymbol{\beta}^T \boldsymbol{x}_i y_j + \boldsymbol{\beta}^T \boldsymbol{x}_j y_i)}, \quad (3.3)$$

where $\boldsymbol{Y}_{(i,j)}^L = (\min(Y_i, Y_j), \max(Y_i, Y_j))$, and $\boldsymbol{y}_{(i,j)}^L$ and $r_{ij}^L$ are the observed values of $\boldsymbol{Y}_{(i,j)}^L$ and $\mathbf{R}_{ij}^L$, respectively. Applying the logarithmic transformation to (3.3), we obtain the function,

$$\ell(\boldsymbol{\beta}) = -\binom{n}{2}^{-1} \sum_{1 \leq i < j \leq n} \log\left(1 + R_{ij}(\boldsymbol{\beta})\right), \text{ where } R_{ij}(\boldsymbol{\beta}) = \exp\left\{-(y_i - y_j) \cdot \boldsymbol{\beta}^T(\boldsymbol{x}_i - \boldsymbol{x}_j)\right\}, \quad (3.4)$$

which can be viewed as a composite log-likelihood (Lindsay, 1988). In high dimensions, we may add a regularization term to $\ell(\boldsymbol{\beta})$, which leads to the regularized chromatography approach.

### 3.2 Confidence Interval and Hypothesis Test: A Likelihood Ratio Approach

We consider the problem of testing some pre-specified low dimensional components of $\boldsymbol{\beta}$. Without loss of generality, assume that $\boldsymbol{\beta}$ can be partitioned as $\boldsymbol{\beta} = (\alpha, \boldsymbol{\gamma})$, where $\alpha \in \mathbb{R}$ and $\boldsymbol{\gamma} \in \mathbb{R}^{d-1}$. Now, we consider the null hypothesis $H_0 : \alpha = \alpha_0$, and treat $\boldsymbol{\gamma}$ as a $(d-1)$-dimensional nuisance parameter. Let $\boldsymbol{\beta}^*$ be the true value of $\boldsymbol{\beta}$, which is usually assumed to be sparse in high dimensional problems. In what follows, we propose a new likelihood ratio test for $H_0$, which is applicable to the high dimensional regime. Before presenting the details of the proposed test, we first briefly explain the challenges of the classical maximum likelihood ratio test (LRT) in high dimensions.

Recall that in the presence of nuisance parameters, the classical LRT statistic is

$$\Lambda_n^L = 2n\{\ell(\widehat{\boldsymbol{\beta}}) - \ell(\alpha_0, \widehat{\boldsymbol{\gamma}}_0)\},$$

where $\widehat{\boldsymbol{\gamma}}_0 := \operatorname{argmax}_{\boldsymbol{\gamma}} \ell(\alpha_0, \boldsymbol{\gamma})$ and $\widehat{\boldsymbol{\beta}} := \operatorname{argmax}_{\boldsymbol{\beta}} \ell(\boldsymbol{\beta})$ are the maximum likelihood estimators (MLEs) under the null hypothesis and full model, respectively. If $\ell(\cdot)$ is a true log-likelihood function, then under certain regularity conditions, one has $\Lambda_n^L \rightsquigarrow \chi_1^2$, under $H_0$, where $\chi_1^2$ denotes the chi-squared distribution of degree of freedom 1.

In what follows, we briefly sketch the proofs of this result in low dimensional linear models and emphasize why the same arguments cannot be applied to regularized estimators in the high dimensional regime. For simplicity, consider the linear model $Y_i = \boldsymbol{X}_i^T \boldsymbol{\beta} + \epsilon_i$, where $\mathbb{E}(\epsilon_i) = 0$ and $\operatorname{var}(\epsilon_i) = 1$. In this case, $\ell(\boldsymbol{\beta}) = -1/(2n) \sum_{i=1}^n (Y_i - \boldsymbol{X}_i^T \boldsymbol{\beta})^2$. Suppose $\boldsymbol{X}_i = (Z_i, \boldsymbol{T}_i^T)^T$, where $\boldsymbol{T}_i \in \mathbb{R}^{d-1}$. Define $\widehat{\boldsymbol{\Sigma}} = (1/n) \sum_{i=1}^n \boldsymbol{X}_i \boldsymbol{X}_i^T$ and $\widehat{\boldsymbol{\Sigma}}_{\boldsymbol{TT}} = (1/n) \sum_{i=1}^n \boldsymbol{T}_i \boldsymbol{T}_i^T$. We have

$$\Lambda_n^L = \sum_{i=1}^n \left\{(Y_i - Z_i \alpha_0 - \boldsymbol{T}_i^T \widehat{\boldsymbol{\gamma}}_0)^2 - (Y_i - \boldsymbol{X}_i^T \boldsymbol{\beta}^*)^2\right\} - \sum_{i=1}^n \left\{(Y_i - \boldsymbol{X}_i^T \widehat{\boldsymbol{\beta}})^2 - (Y_i - \boldsymbol{X}_i^T \boldsymbol{\beta}^*)^2\right\}$$
$$= n(\widehat{\boldsymbol{\beta}} - \boldsymbol{\beta}^*)^T \widehat{\boldsymbol{\Sigma}} (\widehat{\boldsymbol{\beta}} - \boldsymbol{\beta}^*) - n(\widehat{\boldsymbol{\gamma}}_0 - \boldsymbol{\gamma}^*)^T \widehat{\boldsymbol{\Sigma}}_{\boldsymbol{TT}} (\widehat{\boldsymbol{\gamma}}_0 - \boldsymbol{\gamma}^*), \quad (3.5)$$

where the second equality is due to the fact that $(1/n) \sum_{i=1}^n (Y_i - Z_i \alpha_0 - \boldsymbol{T}_i^T \widehat{\boldsymbol{\gamma}}_0) \boldsymbol{T}_i = \mathbf{0}$ and $(1/n) \sum_{i=1}^n (Y_i - \boldsymbol{X}_i^T \widehat{\boldsymbol{\beta}}) \boldsymbol{X}_i = \mathbf{0}$, as $\widehat{\boldsymbol{\gamma}}_0$ and $\widehat{\boldsymbol{\beta}}$ are MLEs. In the low dimensional regime, $\widehat{\boldsymbol{\gamma}}_0$ and $\widehat{\boldsymbol{\beta}}$ can be further approximated by the average of their influence functions, i.e., $\widehat{\boldsymbol{\gamma}}_0 - \boldsymbol{\gamma}^* =$

$(1/n) \sum_{i=1}^n \widehat{\boldsymbol{\Sigma}}_{\boldsymbol{TT}}^{-1} \boldsymbol{T}_i \epsilon_i$ and $\widehat{\boldsymbol{\beta}} - \boldsymbol{\beta}^* = (1/n) \sum_{i=1}^n \widehat{\boldsymbol{\Sigma}}^{-1} \boldsymbol{X}_i \epsilon_i$. Plugging them into (3.5) and applying the central limit theorem, we can easily show that $\Lambda_n \rightsquigarrow \chi_1^2$.

However, the above arguments break down if $\widehat{\gamma}_0$ and $\widehat{\boldsymbol{\beta}}$ are the regularized maximum likelihood estimators. In particular, we encounter the following three challenges: (1) The gradient of the likelihood function evaluated at regularized estimators does not vanish and therefore (3.5) does not hold. (2) The influence function of regularized estimators is ill-posed, because the Hessians $\widehat{\boldsymbol{\Sigma}}$ and $\widehat{\boldsymbol{\Sigma}}_{\boldsymbol{TT}}$ are not invertible in high dimensions. (3) In low dimensions, Knight and Fu (2000) obtained the influence function of regularized estimators and further showed that their limiting distributions contain discrete point masses at 0. This makes the limiting distribution of $\Lambda_n$ intractable. Therefore, the log-likelihood function $\ell(\boldsymbol{\beta})$ and the corresponding likelihood ratio test need to be modified to take into account of high dimensional nuisance parameters $\boldsymbol{\gamma}$.

To address these issues, we propose a new directional likelihood function, which is shown to be a valid inferential tool for $\alpha$ in high dimensional settings. The directional likelihood function for $\alpha$ is defined as
$$\widehat{\ell}(\alpha) = \ell(\alpha, \widehat{\boldsymbol{\gamma}} + (\widehat{\alpha} - \alpha)\widehat{\mathbf{w}}), \tag{3.6}$$
where $\widehat{\boldsymbol{\beta}} := (\widehat{\alpha}, \widehat{\boldsymbol{\gamma}})$ is a first-stage regularized estimator for $\boldsymbol{\beta}^*$, and $\widehat{\mathbf{w}}$ is an estimator for
$$\mathbf{w}^{*T} := \mathbf{H}_{\alpha\boldsymbol{\gamma}}(\mathbf{H}_{\boldsymbol{\gamma\gamma}})^{-1} \in \mathbb{R}^{d-1}, \quad \text{and} \quad \mathbf{H} = -\mathbb{E}\{\nabla^2 \ell(\boldsymbol{\beta}^*)\}. \tag{3.7}$$
Here, the estimators $\widehat{\boldsymbol{\beta}}$ and $\widehat{\mathbf{w}}$ will be introduced later and $\mathbf{H}_{\alpha\boldsymbol{\gamma}}$ and $\mathbf{H}_{\boldsymbol{\gamma\gamma}}$ are the corresponding partitions of $\mathbf{H}$. We can show that the directional likelihood function $\widehat{\ell}(\alpha)$ can be treated as a standard likelihood function for a single unknown parameter $\alpha$. For instance, we define the maximum directional likelihood estimator as
$$\widehat{\alpha}^P = \underset{\alpha \in \mathbb{R}}{\operatorname{argmax}} \, \widehat{\ell}(\alpha). \tag{3.8}$$
To test the null hypothesis $H_0 : \alpha^* = \alpha_0$, we define the maximum directional likelihood ratio test (DLRT) statistic as
$$\Lambda_n = 2n\{\widehat{\ell}(\widehat{\alpha}^P) - \widehat{\ell}(\alpha_0)\}. \tag{3.9}$$

In the following we explain the intuition behind the directional likelihood (3.6) based on the geometry of the likelihood function. The likelihood function $\ell(\boldsymbol{\beta})$ defines a parametrization for a surface $S \subset \mathbb{R}^{d+1}$, in which the coordinates of points can be expressed as $(\boldsymbol{\beta}, \ell(\boldsymbol{\beta})) \in \mathbb{R}^{d+1}$. Consider two smooth functions $\alpha(\cdot) \in \mathbb{R}$ and $\boldsymbol{\gamma}(\cdot) \in \mathbb{R}^{d-1}$, satisfying $\alpha(0) = \alpha^*$, $\alpha'(0) \neq 0$ and $\boldsymbol{\gamma}(0) = \boldsymbol{\gamma}^*$. Define a smooth curve $\boldsymbol{\delta} : I \to \mathbb{R}^{d+1}$, which maps $t \in I$ to $(\alpha(t), \boldsymbol{\gamma}(t), \ell_c(t))$, where $I$ is an interval in $\mathbb{R}$ containing a small neighborhood of 0 and
$$\ell_c(t) = \ell(\alpha(t), \boldsymbol{\gamma}(t)).$$
Note that the curve $\boldsymbol{\delta}$ is within the surface $S$ and passes through the true values $(\alpha^*, \boldsymbol{\gamma}^*, \ell(\boldsymbol{\beta}^*))$ when $t = 0$. Since the curve $\boldsymbol{\delta}$ is determined by the functional forms of $(\alpha(t), \boldsymbol{\gamma}(t))$, we need to decide how to choose $(\alpha(t), \boldsymbol{\gamma}(t))$ such that the likelihood $\ell_c(t)$ along the curve has desired properties locally around $t = 0$. Taking the derivative with respect to $t$, the score function of $\ell_c(t)$ at $t = 0$ is given by
$$S(\alpha^*, \boldsymbol{\gamma}^*) := \frac{d\ell_c(t)}{dt}\bigg|_{t=0} = \alpha'(0) \cdot \nabla_\alpha \ell(\alpha^*, \boldsymbol{\gamma}^*) + \boldsymbol{\gamma}'(0) \cdot \nabla_{\boldsymbol{\gamma}} \ell(\alpha^*, \boldsymbol{\gamma}^*).$$

To construct a valid test based on $\ell_c(t)$, the key step is to ensure that $S(\alpha, \gamma)$ is robust to the perturbation of the high dimensional nuisance parameter $\gamma$. To this end, we require $\mathbb{E}[\nabla_\gamma S(\alpha^*, \gamma^*)] = 0$. This implies $\alpha'(0) \mathbf{H}_{\alpha\gamma} + \gamma'(0) \mathbf{H}_{\gamma\gamma} = 0$, which is equivalent to $\gamma'(0)/\alpha'(0) = -\mathbf{w}^*$ by (3.7). For $t$ in a small neighborhood of 0, the Taylor theorem implies

$$\alpha(t) = \alpha^* + \alpha'(0)t + o(t) \quad \text{and} \quad \gamma(t) = \gamma^* - \alpha'(0)\mathbf{w}^* t + o(t).$$

Ignoring the higher order terms, this gives $\ell_c(t) = \ell(\alpha^* + \alpha'(0)t, \gamma^* - \alpha'(0)\mathbf{w}^* t)$. Finally, a reparametrization of $\ell_c(t)$ with $\alpha := \alpha^* + \alpha'(0)t$ yields a function $\bar{\ell}_c(\alpha)$ of $\alpha$, defined as

$$\bar{\ell}_c(\alpha) := \ell_c\Big(\frac{\alpha - \alpha^*}{\alpha'(0)}\Big) = \ell(\alpha, \gamma^* + (\alpha^* - \alpha)\mathbf{w}^*).$$

Replacing $\alpha^*$, $\gamma^*$ and $\mathbf{w}^*$ by the corresponding estimators $\widehat{\alpha}$, $\widehat{\gamma}$ and $\widehat{\mathbf{w}}$, the function $\bar{\ell}_c(\alpha)$ becomes the directional likelihood in (3.6). Note that, to reduce the perturbation of the high-dimensional nuisance parameter $\gamma$, the key is to find the direction $(1, -\mathbf{w}^{*T})$ of the curve $(\alpha(t), \gamma(t))$ at $t = 0$ in the $d$ dimensional parameter space. Thus, we name $\widehat{\ell}(\alpha)$ as the directional likelihood.

In the following, we consider how to obtain the estimators $\widehat{\alpha}$, $\widehat{\gamma}$ and $\widehat{\mathbf{w}}$ in the directional likelihood (3.6). To estimate $\beta^*$, our inference framework allows a wide class of estimators $\widehat{\beta} = (\widehat{\alpha}, \widehat{\gamma})$ including the regularized estimators with nonconvex (or folded concave) penalty functions; see Remark 4.1. To estimate the $(d-1)$-dimensional vector $\mathbf{w}^*$, we use the following Dantzig-type estimator,

$$\widehat{\mathbf{w}} = \arg\min \|\mathbf{w}\|_1 \quad \text{subject to} \quad \left\|\nabla^2_{\alpha\gamma}\ell(\widehat{\beta}) - \mathbf{w}^T \nabla^2_{\gamma\gamma}\ell(\widehat{\beta})\right\|_\infty \leq \lambda_s, \qquad (3.10)$$

where $\lambda_s$ is a tuning parameter.

To analyze the semiparametric GLM, one technical challenge is that $\nabla \ell(\beta)$ is a high dimensional U-statistic with a possibly unbounded kernel function, i.e.,

$$\nabla \ell(\beta) = \frac{2}{n(n-1)} \cdot \sum_{1 \leq i < j \leq n} \frac{R_{ij}(\beta) \cdot (y_i - y_j) \cdot (\boldsymbol{x}_i - \boldsymbol{x}_j)}{1 + R_{ij}(\beta)}.$$

To decouple the correlation among summands in $\nabla \ell(\beta)$, we extend the Hájek projection (Hoeffding, 1948) to prove the asymptotic normality of a score function in high dimensions. In particular, denote

$$\widehat{\mathbf{U}}_n = \frac{2}{n} \cdot \sum_{i=1}^n \mathbf{g}(y_i, \boldsymbol{x}_i, \beta^*), \quad \text{where} \quad \mathbf{g}(y_i, \boldsymbol{x}_i, \beta) = \frac{n}{2} \cdot \mathbb{E}\Big\{\nabla \ell(\beta) \mid y_i, \boldsymbol{x}_i\Big\}. \qquad (3.11)$$

By definition, $2 \cdot n^{-1} \cdot \mathbf{g}(y_i, \boldsymbol{x}_i, \beta^*)$ is the projection of $\nabla \ell(\beta^*)$ onto the $\sigma$-field generated by $(y_i, \boldsymbol{x}_i)$, and we sum over all subjects to construct $\widehat{\mathbf{U}}_n$. We therefore approximate the U-statistic $\nabla \ell(\beta^*)$ by the sum of independent random variables $\widehat{\mathbf{U}}_n$. Let $\boldsymbol{\Sigma} = \mathbb{E}\{(\mathbf{g}_i^*)^{\otimes 2}\}$ denote the variance of $\mathbf{g}_i^*$, where $\mathbf{g}_i^* = \mathbf{g}(y_i, \boldsymbol{x}_i, \beta^*)$.

In Theorem 4.1, we prove $n^{1/2} \cdot (\widehat{\alpha}^P - \alpha^*) \rightsquigarrow N(0, 4 \cdot \sigma^2 \cdot H_{\alpha|\gamma}^{-2})$, where $\sigma^2 = \boldsymbol{\Sigma}_{\alpha\alpha} - 2\mathbf{w}^{*T}\boldsymbol{\Sigma}_{\gamma\alpha} + \mathbf{w}^{*T}\boldsymbol{\Sigma}_{\gamma\gamma}\mathbf{w}^*$, $H_{\alpha|\gamma} = H_{\alpha\alpha} - \mathbf{H}_{\alpha\gamma}\mathbf{H}_{\gamma\gamma}^{-1}\mathbf{H}_{\gamma\alpha}$ and $\boldsymbol{\Sigma}_{\alpha\alpha}$, $\boldsymbol{\Sigma}_{\gamma\alpha}$ and $\boldsymbol{\Sigma}_{\gamma\gamma}$ are corresponding partitions of $\boldsymbol{\Sigma}$. To construct confidence intervals and Wald-type hypothesis test, one needs to estimate the asymptotic variance, which depends on the unknown covariance and Hessian matrices $\boldsymbol{\Sigma}$ and $\mathbf{H}$. By exploiting the U-statistic structure of $\nabla \ell(\beta)$, we can estimate $\boldsymbol{\Sigma}$ by

$$\widehat{\boldsymbol{\Sigma}} = \frac{1}{n} \cdot \sum_{i=1}^n \Big\{\frac{1}{n-1} \sum_{j=1, j \neq i}^n \frac{R_{ij}(\widehat{\beta}) \cdot (y_i - y_j) \cdot (\boldsymbol{x}_i - \boldsymbol{x}_j)}{1 + R_{ij}(\widehat{\beta})}\Big\}^{\otimes 2}. \qquad (3.12)$$

Thus, we define $\widehat{\sigma}^2 = \widehat{\boldsymbol{\Sigma}}_{\alpha\alpha} - 2\widehat{\mathbf{w}}^T\widehat{\boldsymbol{\Sigma}}_{\gamma\alpha} + \widehat{\mathbf{w}}^T\widehat{\boldsymbol{\Sigma}}_{\gamma\gamma}\widehat{\mathbf{w}}$. Moreover, we can estimate $H_{\alpha|\gamma}$ by $\widehat{H}_{\alpha|\gamma} = -\nabla^2_{\alpha\alpha}\ell(\widehat{\boldsymbol{\beta}}) + \widehat{\mathbf{w}}^T\nabla^2_{\gamma\alpha}\ell(\widehat{\boldsymbol{\beta}})$. Therefore, a confidence interval for $\alpha^*$ with $(1-\omega)$ coverage probability is given by $[\widehat{\alpha}^P - \zeta \cdot n^{-1/2}, \widehat{\alpha}^P + \zeta \cdot n^{-1/2}]$, where $\zeta = 2 \cdot \widehat{\sigma} \cdot \widehat{H}^{-1}_{\alpha|\gamma} \cdot \Phi^{-1}(1-\omega/2)$.

In addition, to test the null hypothesis $H_0 : \alpha^* = \alpha_0$, Theorem 4.2 shows that the maximum directional likelihood ratio test statistic $\Lambda_n$ in (3.9) satisfies $(4 \cdot \sigma^2)^{-1} \cdot H_{\alpha|\gamma} \cdot \Lambda_n \rightsquigarrow \chi^2_1$. Hence our test with the significance level $\omega$ is given by

$$\psi_{\text{DLRT}}(\omega) = \mathbb{1}\{(4 \cdot \widehat{\sigma}^2)^{-1} \cdot \widehat{H}_{\alpha|\gamma} \cdot \Lambda_n \geq \chi^2_{1\omega}\}, \tag{3.13}$$

where $\chi^2_{1\omega}$ is the $(1-\omega)$-th quantile of a $\chi^2_1$ random variable. The null hypothesis is rejected if and only if $\psi_{\text{DLRT}}(\omega) = 1$, and the associated p-value is given by $P_{\text{DLRT}} = 1 - \chi^2_1((4 \cdot \widehat{\sigma}^2)^{-1} \cdot \widehat{H}_{\alpha|\gamma} \cdot \Lambda_n)$, where $\chi^2_1(\cdot)$ is the c.d.f of a chi-squared distribution with degree of freedom 1. In Corollary 4.2, we prove that the directional likelihood ratio test can control the type I error asymptotically in high dimensions, i.e., $\lim_{n\to\infty} \mathbb{P}(\psi_{\text{DLRT}}(\omega) = 1 \mid H_0) = \omega$ and the p-value is asymptotically uniformly distributed, i.e., $P_{\text{DLRT}} \rightsquigarrow \text{Uniform}[0,1]$, under $H_0$.

## 4 Main Results

We first prove the asymptotic normality of the maximum directional likelihood estimator $\widehat{\alpha}^P$ in (3.8). Then, we prove the limiting distribution of $\Lambda_n$ as well as the validity of the maximum directional likelihood ratio test in (3.13) under the null hypothesis $H_0 : \alpha^* = \alpha_0$. In the following, we first present some regularity conditions.

**Assumption 4.1.** Assume that $Y$ given $\boldsymbol{X}$ has the sub-exponential tail, i.e., for any $\delta > 0$, $\mathbb{P}(|Y| \geq \delta \mid \boldsymbol{X}) \leq C' \cdot \exp(-C \cdot \delta)$, for some constants $C, C' > 0$. We also assume that the covariates are uniformly bounded, i.e., $\|\boldsymbol{X}\|_\infty = \mathcal{O}(1)$.

It is easily seen that the sub-exponential condition holds for most commonly used GLMs in practice. Following van de Geer et al. (2014), we also assume the bounded covariates for simplicity. It can be further relaxed to some sub-Gaussian or sub-exponential type assumptions.

**Assumption 4.2.** Recall that $\mathbf{g}(y_i, \boldsymbol{x}_i, \boldsymbol{\beta}^*)$ and $\mathbf{H}$ are defined in (3.11) and (3.7), respectively. Denote

$$\boldsymbol{\Sigma} = \mathbb{E}\{\mathbf{g}(y_i, \boldsymbol{x}_i, \boldsymbol{\beta}^*)^{\otimes 2}\}, \quad H_{\alpha|\gamma} = H_{\alpha\alpha} - \mathbf{H}_{\alpha\gamma}\mathbf{H}^{-1}_{\gamma\gamma}\mathbf{H}_{\gamma\alpha}.$$

Assume that $\lambda_{\min}(\boldsymbol{\Sigma}) \geq c$, $\lambda_{\min}(\mathbf{H}) \geq c$ and $\lambda_{\max}(\mathbf{H}) \leq c'$, for some constants $c, c' > 0$, $\|\mathbf{H}\|_\infty = \mathcal{O}(1)$, $H_{\alpha|\gamma} = \mathcal{O}(1)$ and $H^{-1}_{\alpha|\gamma} = \mathcal{O}(1)$.

Note that $\boldsymbol{\Sigma}$ can be interpreted as the second moment of the Hájek projection, which approximates the asymptotic variance of $\nabla\ell(\boldsymbol{\beta}^*)$, and $H_{\alpha|\gamma}$ is known as the partial information matrix for $\alpha$ in the literature. This condition merely assumes that the minimum eigenvalues of $\boldsymbol{\Sigma}$ and $\mathbf{H}$ are lower bounded by a constant, and $H_{\alpha|\gamma}$ is $\mathcal{O}(1)$ and invertible. These assumptions are standard for low dimensional models to guarantee the existence of the asymptotic variance of MLEs.

**Assumption 4.3.** Let $s_1 = \|\mathbf{w}^*\|_0$, where $\mathbf{w}^*$ is defined in (3.7). Assume that $s_1^{3/2} \cdot n^{-1/2} \cdot M^3 = o_\mathbb{P}(1)$, where $M := \max_{1 \leq i < j \leq n} \|(y_i - y_j) \cdot (\boldsymbol{x}_i - \boldsymbol{x}_j)\|_\infty$.

Note that if $Y$ is a bounded random variable, then together with Assumption 4.1, we have $M = \mathcal{O}_\mathbb{P}(1)$. If $Y$ is unbounded with the sub-exponential tail, we can take $M = \mathcal{O}_\mathbb{P}(\log n)$. This assumption essentially requires that $\mathbf{w}^*$ is sufficiently sparse (i.e., $s_1^3 = o(n)$ up to polynomials of $\log n$), such that the approximation error in the Hájek projection is controllable. As shown later in Theorem 4.1, the asymptotic variance of $\widehat{\alpha}^P$ depends on $\mathbf{w}^*$. Without the sparsity assumption on $\mathbf{w}^*$, the asymptotic variance seems unlikely to be consistently estimated. We note that, in van de Geer et al. (2014), one of their key assumptions is that the inverse of the Fisher information matrix $\mathbf{\Omega} = \mathbf{H}^{-1}$ is sparse. Let $\mathbf{\Omega}_{*\alpha}$ and $\mathbf{\Omega}_{*\gamma}$ denote the columns of $\mathbf{\Omega}$ corresponding to $\alpha$ and $\boldsymbol{\gamma}$. To see the connections, consider the following block matrix inverse formula, $\mathbf{\Omega}_{*\alpha} = H_{\alpha|\gamma}^{-1}(1, -\mathbf{H}_{\alpha\gamma}\mathbf{H}_{\gamma\gamma}^{-1})^T$, where $H_{\alpha|\gamma} = H_{\alpha\alpha} - \mathbf{H}_{\alpha\gamma}\mathbf{H}_{\gamma\gamma}^{-1}\mathbf{H}_{\gamma\alpha}$. Since $\mathbf{w}^* = \mathbf{H}_{\gamma\gamma}^{-1}\mathbf{H}_{\gamma\alpha}$, we have $||\mathbf{w}^*||_0 = ||\mathbf{\Omega}_{*\alpha}||_0 - 1$. Hence, our sparsity assumption on $\mathbf{w}^*$ is implied by the sparsity of $\mathbf{\Omega}$. Moreover, our results reveal that the sparsity of $\mathbf{\Omega}_{*\gamma}$ is not needed for the inference on $\alpha$.

**Assumption 4.4.** Assume that $||\widehat{\boldsymbol{\beta}} - \boldsymbol{\beta}^*||_2 = \mathcal{O}_\mathbb{P}(s^{1/2} \cdot \sqrt{\log d/n})$ and $||\widehat{\boldsymbol{\beta}} - \boldsymbol{\beta}^*||_1 = \mathcal{O}_\mathbb{P}(s \cdot \sqrt{\log d/n})$, where $s = ||\boldsymbol{\beta}^*||_0$. It holds that $\max\{s, s_1\} \cdot \lambda_s \cdot \sqrt{\log d} = o(1)$ and $||\mathbf{w}^*||_1^4 \cdot M^2 \cdot \sqrt{\log n/n} = o(1)$, where $s_1$ is defined in Assumption 4.3, and $\lambda_s \asymp ||\mathbf{w}^*||_1 \cdot M \cdot (s + M) \cdot \sqrt{\log d/n}$.

Assumption 4.4 specifies the scaling of $s, s_1, d$ and $n$. For instance, if $M$ and $||\mathbf{w}^*||_1$ are constants, Assumption 4.4 reduces to $s^2 \cdot \log d = o(n^{1/2})$, and $s \cdot s_1 \cdot \log d = o(n^{1/2})$. Note that the rate of $\lambda_s$ is slower than the conventional $\sqrt{\log d/n}$ rate. This is because (3.10) is different from the original formulation of the Dantzig selector. Specifically, in (3.10), both $\nabla^2_{\alpha\gamma}\ell(\widehat{\boldsymbol{\beta}})$ and $\nabla^2_{\gamma\gamma}\ell(\widehat{\boldsymbol{\beta}})$ depend on the estimator $\widehat{\boldsymbol{\beta}}$, whose estimation error can accumulate and therefore inflates the magnitude of $\lambda_s$.

**Remark 4.1.** Note that Assumption 4.4 requires that the initial estimator $\widehat{\boldsymbol{\beta}}$ satisfies $||\widehat{\boldsymbol{\beta}} - \boldsymbol{\beta}^*||_2 = \mathcal{O}_\mathbb{P}(s^{1/2} \cdot \sqrt{\log d/n})$ and $||\widehat{\boldsymbol{\beta}} - \boldsymbol{\beta}^*||_1 = \mathcal{O}_\mathbb{P}(s \cdot \sqrt{\log d/n})$. In high dimensional settings, we can estimate $\boldsymbol{\beta}$ by maximizing the following penalized likelihood function with a generic penalty function $p_\lambda(\cdot)$,

$$\widehat{\boldsymbol{\beta}} \in \underset{\boldsymbol{\beta}}{\operatorname{argmax}} \left\{ \ell(\boldsymbol{\beta}) - \sum_{j=1}^{d} p_\lambda(\beta_j) \right\}, \tag{4.1}$$

where $\lambda \geq 0$ is a tuning parameter. The proof of $||\widehat{\boldsymbol{\beta}} - \boldsymbol{\beta}^*||_2 = \mathcal{O}_\mathbb{P}(s^{1/2} \cdot \sqrt{\log d/n})$ and $||\widehat{\boldsymbol{\beta}} - \boldsymbol{\beta}^*||_1 = \mathcal{O}_\mathbb{P}(s \cdot \sqrt{\log d/n})$ follows from the similar arguments in the existing literature (Fan et al., 2012; Wang et al., 2013a; Loh and Wainwright, 2013; Wang et al., 2013b). We defer the detailed results to appendix. Here, we would like to emphasize that, our inferential framework allows general regularized estimators such as nonconvex penalty functions. It is more flexible than the method proposed by van de Geer et al. (2014) based on inverting Karush-Kuhn-Tucker (KKT) condition for the Lasso estimator.

The following main theorem establishes the asymptotic normality of the maximum directional likelihood estimator $\widehat{\alpha}^P$.

**Theorem 4.1.** Under Assumptions 4.1, 4.2, 4.3 and 4.4, we have

$$n^{1/2} \cdot (\widehat{\alpha}^P - \alpha^*) \rightsquigarrow N(0, 4 \cdot \sigma^2 \cdot H_{\alpha|\gamma}^{-2}), \quad \text{where} \quad \sigma^2 = \mathbf{\Sigma}_{\alpha\alpha} - 2\mathbf{w}^{*T}\mathbf{\Sigma}_{\gamma\alpha} + \mathbf{w}^{*T}\mathbf{\Sigma}_{\gamma\gamma}\mathbf{w}^*.$$

*Proof.* The detailed proof is shown in Section 5. □

To apply this theorem in practice, one needs to estimate the asymptotic variance $\sigma^2 \cdot H_{\alpha|\gamma}^{-2}$, which depends on the unknown covariance matrix $\mathbf{\Sigma}$ and $H_{\alpha|\gamma}$. Recall that such an estimator $\widehat{\mathbf{\Sigma}}$ is given in (3.12). The following corollary shows asymptotic normality holds when $\sigma$ and $H_{\alpha|\gamma}$ are replaced by their estimators.

**Corollary 4.1.** Assume that the Assumptions 4.1, 4.2, 4.3, and 4.4 hold. In addition, we assume that $||\mathbf{\Sigma}_{\gamma\gamma}||_\infty = \mathcal{O}(1)$, $||\mathbf{\Sigma}_{\alpha\gamma}||_\infty = \mathcal{O}(1)$ and $||\mathbf{w}^*||_1^2 \cdot M^3 \cdot s \cdot \sqrt{\log d/n} = o(1)$. Then

$$\widehat{\sigma}^{-1} \cdot \widehat{H}_{\alpha|\gamma} \cdot n^{1/2} \cdot (\widehat{\alpha}^P - \alpha^*) \rightsquigarrow N(0, 4).$$

*Proof.* The detailed proof is shown in Appendix A.1. □

We note that the estimator $\widehat{\alpha}^P$ is not semiparametrically efficient, because not all information about $\boldsymbol{\beta}$ is retained in the statistical chromatography. However, under certain situations, we can show that $\widehat{\alpha}^P$ achieves the semiparametric information lower bound.

**Proposition 4.1.** When $\boldsymbol{\beta}^* = \mathbf{0}$, the estimator $\widehat{\alpha}^P$ is semiparametrically efficient.

*Proof.* The detailed proof is shown in Appendix A.4. □

This proposition suggests that when the covariates have null effects, the estimator $\widehat{\alpha}^P$ is semiparametrically efficient. Since in high dimensional settings, most of the covariates have null or weak effects, we expect that the estimator $\widehat{\alpha}^P$ gains model flexibility and computational efficiency without paying much price on the information loss. This fact is further reinforced through numerical studies.

Next, we prove the asymptotic distribution of the test statistic $\Lambda_n$ and the validity of the maximum likelihood ratio test under the same conditions in Theorem 4.1 and Corollary 4.1.

**Theorem 4.2.** Under Assumptions 4.1, 4.2, 4.3 and 4.4, if $\alpha^* = \alpha_0$, then we have

$$(4 \cdot \sigma^2)^{-1} \cdot H_{\alpha|\gamma} \cdot \Lambda_n \rightsquigarrow \chi_1^2.$$

*Proof.* The detailed proof is shown in Appendix A.3. □

As before, to apply the theorem in practice, we replace $\sigma^2$ and $H_{\alpha|\gamma}$ with their estimators. The following corollary shows that under $H_0$, type I error of the test $\psi_{\text{DLRT}}(\omega)$ converges to the desired significance level $\omega$ and the p-value is asymptotically uniform.

**Corollary 4.2.** Suppose the conditions in Corollary 4.1 hold. Then

$$\lim_{n\to\infty} \mathbb{P}(\psi_{\text{DLRT}}(\omega) = 1 \mid H_0) = \omega \quad \text{and} \quad P_{\text{DLRT}} \rightsquigarrow \text{Uniform}[0, 1] \quad \text{under } H_0,$$

where $\psi_{\text{DLRT}}(\omega)$ is defined in (3.13) and $P_{\text{DLRT}} = 1 - \chi_1^2((4 \cdot \widehat{\sigma}^2)^{-1} \cdot \widehat{H}_{\alpha|\gamma} \cdot \Lambda_n)$ is the associated p-value.

*Proof.* The detailed proof is shown in Appendix A.1. □

Note that, our assumptions do not contain any type of minimal signal strength condition on the nonzero components of $\boldsymbol{\beta}^*$. Therefore, unlike the oracle-type results in Fan and Li (2001), variable selection consistency is not a priori for our approach and a valid p-value can be produced even if a covariate is not selected in the model.

**Remark 4.2** (Misspecified Model). One additional advantage of our likelihood based inference over the methods in Zhang and Zhang (2014); van de Geer et al. (2014) and Javanmard and Montanari (2013) is that we can provide theoretical justifications even if the model is misspecified. In the misspecified setting, let $\boldsymbol{\beta}^o$ denote the oracle parameter (i.e., least false parameter) that minimizes the rank Kullback-Leibler divergence, i.e., $\boldsymbol{\beta}^o = \arg\min_{\boldsymbol{\beta}} \text{RKL}(\ell^*, \ell(\boldsymbol{\beta}))$, where $\text{RKL}(\ell^*, \ell(\boldsymbol{\beta}))$ stands for the rank Kullback-Leibler divergence (RKL) between the assumed semiparametric GLM and the true model; see appendix for the detailed definition. Suppoe $\boldsymbol{\beta}^o$ is unique. Given an initial estimator, one can follow the similar idea in Section 3 to test the hypothesis for the oracle parameter $H_0^o : \alpha^o = \alpha_0$ under the misspecified model. Additional details can be found in the appendix.

## 5 Proof of Theorem 4.1

In this section, we give the proof of Theorem 4.1. The proofs of the remaining results, including Corollary 4.1, Theorem 4.2 and Proposition 4.1 are deferred to appendix.

We define an unbiased score function as $S(\boldsymbol{\beta}^*) := \nabla_\alpha \ell(\boldsymbol{\beta}^*) - \mathbf{w}^{*T}\nabla_\gamma \ell(\boldsymbol{\beta}^*)$, which plays an important role in the proof. The proof of Theorem 4.1 has three steps. First, we show that the first derivative of $\widehat{\ell}(\alpha)$ approximates $S(\boldsymbol{\beta}^*)$. Second, we apply the central limit theorem for a linear combination of increasing dimensional U-statistics to conclude the asymptotic normality of $S(\boldsymbol{\beta}^*)$. Finally, we show that the negative Hessian of $\widehat{\ell}(\alpha)$ approximates $H_{\alpha|\gamma}$.

**Step 1: Show the convergence of $\widehat{\ell}'(\alpha^*)$.** Define $\widehat{\boldsymbol{\gamma}}(\alpha) := \widehat{\boldsymbol{\gamma}} + (\widehat{\alpha} - \alpha)\widehat{\mathbf{w}}$ and $\widehat{\boldsymbol{\Delta}}_{\boldsymbol{\gamma}} = \widehat{\boldsymbol{\gamma}}(\alpha^*) - \boldsymbol{\gamma}^*$. Moreover, recall that $S(\boldsymbol{\beta}^*) := \nabla_\alpha \ell(\boldsymbol{\beta}^*) - \mathbf{w}^{*T}\nabla_\gamma \ell(\boldsymbol{\beta}^*)$. By the chain rule and mean value theorem, we have

$$\begin{aligned}\widehat{\ell}'(\alpha^*) &= \nabla_\alpha \ell(\alpha^*, \widehat{\boldsymbol{\gamma}}(\alpha^*)) - \widehat{\mathbf{w}}^T \nabla_\gamma \ell(\alpha^*, \widehat{\boldsymbol{\gamma}}(\alpha^*)) \\ &= \nabla_\alpha \ell(\boldsymbol{\beta}^*) + \nabla^2_{\alpha\gamma}\ell(\alpha^*, \bar{\boldsymbol{\gamma}})\widehat{\boldsymbol{\Delta}}_{\boldsymbol{\gamma}} - \left\{\widehat{\mathbf{w}}^T \nabla_\gamma \ell(\boldsymbol{\beta}^*) + \widehat{\mathbf{w}}^T \nabla^2_{\gamma\gamma}\ell(\alpha^*, \widetilde{\boldsymbol{\gamma}})\widehat{\boldsymbol{\Delta}}_{\boldsymbol{\gamma}}\right\} \\ &= S(\boldsymbol{\beta}^*) + \underbrace{(\mathbf{w}^* - \widehat{\mathbf{w}})^T \nabla_\gamma \ell(\boldsymbol{\beta}^*)}_{I_1} + \underbrace{\left[\{\nabla^2_{\alpha\gamma}\ell(\alpha^*, \bar{\boldsymbol{\gamma}}) - \widehat{\mathbf{w}}^T \nabla^2_{\gamma\gamma}\ell(\alpha^*, \widetilde{\boldsymbol{\gamma}})\}\widehat{\boldsymbol{\Delta}}_{\boldsymbol{\gamma}}\right]}_{I_2}, \quad (5.1)\end{aligned}$$

where $\bar{\boldsymbol{\gamma}}$ and $\widetilde{\boldsymbol{\gamma}}$ are intermediate values between $\boldsymbol{\gamma}^*$ and $\widehat{\boldsymbol{\gamma}}(\alpha^*)$. Thus, the first step of the proof reduces to controlling the two terms $I_1$ and $I_2$ in (5.1). In particular, to bound $I_1$, we need the following Lemma 5.1 to bound $\|\widehat{\mathbf{w}} - \mathbf{w}^*\|_1$ and Lemma 5.2 to bound $\|\nabla \ell(\boldsymbol{\beta}^*)\|_\infty$, respectively.

**Lemma 5.1.** Under the conditions in Theorem 4.1,

$$\|\widehat{\mathbf{w}} - \mathbf{w}^*\|_1 = \mathcal{O}_{\mathbb{P}}\Big(\|\mathbf{w}^*\|_1 \cdot s_1 \cdot M \cdot (s + M) \cdot \sqrt{\frac{\log d}{n}}\Big).$$

*Proof.* To prove this Lemma, we first verify that

$$\|\nabla^2_{\alpha\gamma}\ell(\widehat{\boldsymbol{\beta}}) - \mathbf{w}^{*T}\nabla^2_{\gamma\gamma}\ell(\widehat{\boldsymbol{\beta}})\|_\infty = \mathcal{O}_{\mathbb{P}}\Big(\|\mathbf{w}^*\|_1 \cdot M \cdot (s + M) \cdot \sqrt{\frac{\log d}{n}}\Big).$$

This implies that $\mathbf{w}^*$ is in the feasible region of the constrained optimization problem (3.10), if we take $\lambda_s \asymp \|\mathbf{w}^*\|_1 \cdot M \cdot (s + M) \cdot \sqrt{\log d/n}$. In addition, we can prove that for $n$ large enough,

$$\inf_{\mathbf{v}\in\mathcal{C}} \frac{s_1 \cdot (-\mathbf{v}^T \nabla^2_{\gamma\gamma}\ell(\widehat{\boldsymbol{\beta}})\mathbf{v})}{\|\mathbf{v}_{S_{\mathbf{w}^*}}\|_1^2} \geq \rho', \text{ where } \mathcal{C} = \{\mathbf{v} \in \mathbb{R}^{d-1} : \|\mathbf{v}_{S_{\mathbf{w}^*}^c}\|_1 \leq \|\mathbf{v}_{S_{\mathbf{w}^*}}\|_1\},$$

where $\rho'$ is a positive constant and $S_{\mathbf{w}^*} = \{j : w_j^* \neq 0\}$ is the support set for $\mathbf{w}^*$. This is known as the compatibility factor condition which ensures the strong convexity of the loss function within the cone $\mathcal{C}$. Following the similar strategy to the proof of Theorem 7.1 of Bickel et al. (2009), we can prove that $||\widehat{\mathbf{w}} - \mathbf{w}^*||_1 = \mathcal{O}_{\mathbb{P}}(s_1 \cdot \lambda_s)$. The detailed proof is shown in Appendix B. □

**Lemma 5.2.** Assume that Assumption 4.1 holds. Then, for any $C'' > 0$, we have $\|\nabla \ell(\boldsymbol{\beta}^*)\|_\infty \leq C'' \cdot \sqrt{\frac{\log d}{n}}$, with probability at least

$$1 - 2 \cdot d \cdot \exp\left[-\min\left\{\frac{C^2 \cdot C''^2}{2^9 \cdot C'^2 \cdot m^2} \cdot \frac{\log d}{n}, \frac{C \cdot C''}{2^5 \cdot C' \cdot m} \cdot \sqrt{\frac{\log d}{n}}\right\} \cdot k\right], \quad (5.2)$$

where $k = \lfloor n/2 \rfloor$, $m = \max_{1 \leq i \leq n} \max_{1 \leq j \leq d} |x_{ij}|$, and $C, C'$ are defined in Assumption 4.1.

*Proof.* To prove Lemma 5.2, the key is to prove a new concentration inequality for U-statistics with sub-exponential kernel functions. In particular, the following lemma allows the kernel function to be unbounded, which is more general than most of existing concentration results for U-statistics with bounded kernels, such as Theorem 4.1.13 in de la Pena and Giné (1999). The following result can be of independent interest, whose proof is shown in Appendix B.

**Lemma 5.3.** Let $X_1, ..., X_n$ be independent random variables. Consider the following U-statistics of order $m$,

$$U_n = \binom{n}{m}^{-1} \sum_{i_1 < ... < i_m} u(X_{i_1}, ..., X_{i_m}),$$

where the summation is over all $i_1 < ... < i_m$ selected from $\{1, ..., n\}$ and $\mathbb{E}[u(X_{i_1}, ..., X_{i_m})] = 0$ for all $i_1 < ... < i_m$. We assume that the kernel function $u(X_{i_1}, ..., X_{i_m})$ is symmetric in the sense that $u(X_{i_1}, ..., X_{i_m})$ is independent of the order of $X_{i_1}, ..., X_{i_m}$. If there exist constants $L_1$ and $L_2$, such that

$$\mathbb{P}(|u(X_{i_1}, ..., X_{i_m})| \geq x) \leq L_1 \cdot \exp(-L_2 \cdot x), \quad (5.3)$$

for all $i_1 < ... < i_m$ and all $x \geq 0$, then

$$\mathbb{P}(|U_n| \geq x) \leq 2 \cdot \exp\left[-\min\left\{\frac{L_2^2 \cdot x^2}{8 \cdot L_1^2}, \frac{L_2 \cdot x}{4 \cdot L_1}\right\} \cdot k\right],$$

where $k = \lfloor n/m \rfloor$ is the largest integer less than $n/m$.

Given the above Lemma, we need to verify that the kernel function $\mathbf{h}_{ij}(\boldsymbol{\beta}^*)$ has mean 0, where

$$\mathbf{h}_{ij}(\boldsymbol{\beta}) = \frac{R_{ij}(\boldsymbol{\beta}) \cdot (y_i - y_j) \cdot (\boldsymbol{x}_i - \boldsymbol{x}_j)}{1 + R_{ij}(\boldsymbol{\beta})}, \quad (5.4)$$

and it satisfies (5.3). To show $\mathbb{E}\{\mathbf{h}_{ij}(\boldsymbol{\beta}^*)\} = 0$, let $\Xi_{ij}$ denote the event $\{(Y_{(i)}^L, Y_{(j)}^L) = (y_i, y_j), \boldsymbol{X}_i = \boldsymbol{x}_i, \boldsymbol{X}_j = \boldsymbol{x}_j\}$. By (3.3), we can show that the conditional distribution of $Y_i$ and $Y_j$ given $\Xi_{ij}$ follows a binomial distribution,

$$\mathbb{P}(Y_i = y_i, Y_j = y_j \mid \Xi_{ij}; \boldsymbol{\beta}) = \frac{1}{1 + R_{ij}(\boldsymbol{\beta})}, \quad \text{and} \quad \mathbb{P}(Y_i = y_j, Y_j = y_i \mid \Xi_{ij}; \boldsymbol{\beta}) = \frac{R_{ij}(\boldsymbol{\beta})}{1 + R_{ij}(\boldsymbol{\beta})}. \quad (5.5)$$

According to this binomial distribution, the conditional expectation of $\mathbf{h}_{ij}(\boldsymbol{\beta}^*)$ given $\Xi_{ij}$ is

$$\mathbb{E}\{\mathbf{h}_{ij}(\boldsymbol{\beta}^*) \mid \Xi_{ij}; \boldsymbol{\beta}^*\} = \frac{R_{ij}(\boldsymbol{\beta}^*) \cdot (y_i - y_j) \cdot (\boldsymbol{x}_i - \boldsymbol{x}_j)}{1 + R_{ij}(\boldsymbol{\beta}^*)} \cdot \mathbb{P}(Y_i = y_i, Y_j = y_j \mid \Xi_{ij}; \boldsymbol{\beta}^*)$$
$$+ \frac{R_{ij}^{-1}(\boldsymbol{\beta}^*) \cdot (y_j - y_i) \cdot (\boldsymbol{x}_i - \boldsymbol{x}_j)}{1 + R_{ij}^{-1}(\boldsymbol{\beta})} \cdot \mathbb{P}(Y_i = y_j, Y_j = y_i \mid \Xi_{ij}; \boldsymbol{\beta}^*).$$

By plugging (5.5) into the above equation, it is easy to verify that $\mathbb{E}\{\mathbf{h}_{ij}(\boldsymbol{\beta}^*) \mid \Xi_{ij}\} = 0$. Finally, the double expectation rule yields $\mathbb{E}\{\mathbf{h}_{ij}(\boldsymbol{\beta}^*)\} = \mathbb{E}\big[\mathbb{E}\{\mathbf{h}_{ij}(\boldsymbol{\beta}^*)|\Xi_{ij}\}\big] = 0$. Next, we verify that the kernel function satisfies (5.3). Since $R_{ij}(\boldsymbol{\beta}) > 0$ and $\max_{ij}|x_{ij}| \leq m$, we have

$$\|\mathbf{h}_{ij}(\boldsymbol{\beta}^*)\|_\infty \leq \|(y_i - y_j) \cdot (\boldsymbol{x}_i - \boldsymbol{x}_j)\|_\infty \leq 2 \cdot m \cdot |y_i - y_j|.$$

By the sub-exponential tail condition on $y_i$, for any $x > 0$ and $k = 1, ..., d$,

$$\mathbb{P}\big(|[\mathbf{h}_{ij}(\boldsymbol{\beta}^*)]_k| > x\big) \leq \mathbb{P}\big(|y_i - y_j| > (2 \cdot m)^{-1} \cdot x\big) \leq 2 \cdot C' \cdot \exp\{-C \cdot (4 \cdot m)^{-1} \cdot x\}.$$

Then we apply Lemma 5.3 with $k = \lfloor n/2 \rfloor$. This completes the proof of Lemma 5.2. □

Hence, by Lemma 5.1 and Lemma 5.2, we can show that

$$|I_1| \leq \|\mathbf{w}^* - \widehat{\mathbf{w}}\|_1 \cdot \|\nabla_\gamma \ell(\boldsymbol{\beta}^*)\|_\infty = \mathcal{O}_{\mathbb{P}}\big(\lambda_s \cdot s_1\big) \cdot \mathcal{O}_{\mathbb{P}}\Big(\sqrt{\frac{\log d}{n}}\Big) = o_{\mathbb{P}}\Big(\frac{1}{\sqrt{n}}\Big),$$

where the last step follows by Assumption 4.4. We further separate $I_2$ into the following terms,

$$|I_2| \leq \underbrace{\big|\{\nabla^2_{\alpha\gamma}\ell(\widehat{\boldsymbol{\beta}}) - \widehat{\mathbf{w}}^T \nabla^2_{\gamma\gamma}\ell(\widehat{\boldsymbol{\beta}})\}\widehat{\boldsymbol{\Delta}}_\gamma\big|}_{I_{21}} + \underbrace{\big|\{\nabla^2_{\alpha\gamma}\ell(\widehat{\boldsymbol{\beta}}) - \nabla^2_{\alpha\gamma}\ell(\alpha^*, \bar{\boldsymbol{\gamma}})\}\widehat{\boldsymbol{\Delta}}_\gamma\big|}_{I_{22}}$$
$$+ \underbrace{\big|\widehat{\mathbf{w}}^T\{\nabla^2_{\gamma\gamma}\ell(\widehat{\boldsymbol{\beta}}) - \nabla^2_{\gamma\gamma}\ell(\alpha^*, \widetilde{\boldsymbol{\gamma}})\}\widehat{\boldsymbol{\Delta}}_\gamma\big|}_{I_{23}}. \quad (5.6)$$

To control the three terms, we first need to bound $\|\widehat{\boldsymbol{\Delta}}_\gamma\|_1$. By Assumption 4.4, we have $\|\widehat{\boldsymbol{\gamma}} - \boldsymbol{\gamma}^*\|_1 = \mathcal{O}_{\mathbb{P}}(s \cdot \sqrt{\log d/n})$ and $|\widehat{\alpha} - \alpha^*| = \mathcal{O}_{\mathbb{P}}(s^{1/2} \cdot \sqrt{\log d/n})$. Moreover, by Cauchy-Schwarz inequality, it holds that $\|\mathbf{w}^*\|_1 \leq \sqrt{s_1}\|\mathbf{w}^*\|_2 \leq \sqrt{s_1}\|\mathbf{H}_{\gamma\gamma}^{-1}\mathbf{H}_{\alpha\gamma}^T\|_2 \leq \sqrt{s_1}\lambda_{\min}(\mathbf{H})^{-1}\lambda_{\max}(\mathbf{H}) \leq \sqrt{s_1}c^{-1}c'$, where the last inequality is by Assumption 4.2. Therefore,

$$\|\widehat{\boldsymbol{\Delta}}_\gamma\|_1 \leq \|\widehat{\boldsymbol{\gamma}} - \boldsymbol{\gamma}^*\| + |\widehat{\alpha} - \alpha^*|\|\widehat{\mathbf{w}}\|_1 = \mathcal{O}_{\mathbb{P}}\Big(s \cdot \sqrt{\frac{\log d}{n}} + \sqrt{ss_1} \cdot \sqrt{\frac{\log d}{n}}\Big) = \mathcal{O}_{\mathbb{P}}\Big(\max\{s, s_1\}\sqrt{\frac{\log d}{n}}\Big),$$

where we used the fact that $\|\widehat{\mathbf{w}}\|_1 = \|\mathbf{w}^*\|_1 + o_{\mathbb{P}}(1) = \mathcal{O}_{\mathbb{P}}(s_1^{1/2})$. Now, we consider these three terms in (5.6) one by one. For $I_{21}$, by the definition of $\widehat{\mathbf{w}}$,

$$I_{21} \leq \|\nabla^2_{\alpha\gamma}\ell(\widehat{\boldsymbol{\beta}}) - \widehat{\mathbf{w}}^T\nabla^2_{\gamma\gamma}\ell(\widehat{\boldsymbol{\beta}})\|_\infty \cdot \|\widehat{\boldsymbol{\Delta}}_\gamma\|_1 = \mathcal{O}_{\mathbb{P}}\Big(\lambda_s \cdot \max\{s, s_1\} \cdot \sqrt{\frac{\log d}{n}}\Big) = o_{\mathbb{P}}\Big(\frac{1}{\sqrt{n}}\Big).$$

Now, we consider $I_{22}$. The key step is to quantify the smoothness of the Hessian matrix $\nabla^2\ell(\alpha^*, \boldsymbol{\gamma})$ in a small neighborhood of $\boldsymbol{\gamma}^*$. It is shown in the following Lemma.

**Lemma 5.4.** Under the conditions in Theorem 4.1, for any deterministic sequence $\delta_n$ such that $M \cdot \delta_n = o(1)$,
$$\sup_{\|\boldsymbol{\beta}-\boldsymbol{\beta}^*\|_1 \leq \delta_n} \|\nabla^2 \ell(\boldsymbol{\beta}) - \nabla^2 \ell(\boldsymbol{\beta}^*)\|_\infty = \mathcal{O}_\mathbb{P}(M \cdot \delta_n).$$

*Proof.* Let $w_{ij} = \exp\{-(y_i - y_j) \cdot \boldsymbol{\Delta}^T(\boldsymbol{x}_i - \boldsymbol{x}_j)\}$, where $\boldsymbol{\Delta} = \boldsymbol{\beta} - \boldsymbol{\beta}^*$. By definition, $R_{ij}(\boldsymbol{\beta}) = R_{ij}(\boldsymbol{\beta}^*) \cdot w_{ij}$. Thus

$$\nabla^2 \ell(\boldsymbol{\beta}) = -\binom{n}{2}^{-1} \sum_{1 \leq i < j \leq n} \frac{u_{ij} \cdot R_{ij}(\boldsymbol{\beta}^*) \cdot (y_i - y_j)^2 \cdot (\boldsymbol{x}_i - \boldsymbol{x}_j)^{\otimes 2}}{(1 + R_{ij}(\boldsymbol{\beta}^*))^2},$$

where $u_{ij} = w_{ij} \cdot (1 + R_{ij}(\boldsymbol{\beta}^*))^2 (1 + w_{ij} \cdot R_{ij}(\boldsymbol{\beta}^*))^{-2}$. Note that if $w_{ij} \geq 1$, then $(1 + R_{ij}(\boldsymbol{\beta}^*))^2/(1 + w_{ij} \cdot R_{ij}(\boldsymbol{\beta}^*))^2 \leq 1$. On the other hand, if $w_{ij} \leq 1$,

$$\frac{(1 + R_{ij}(\boldsymbol{\beta}^*))^2}{(1 + w_{ij} \cdot R_{ij}(\boldsymbol{\beta}^*))^2} \leq \frac{(1 + R_{ij}(\boldsymbol{\beta}^*))^2}{w_{ij}^2 \cdot (1 + R_{ij}(\boldsymbol{\beta}^*))^2} = \frac{1}{w_{ij}^2}.$$

Thus $u_{ij} \leq \max\{w_{ij}, w_{ij}^{-1}\}$. Therefore, for any $1 \leq s, t \leq d$,

$$\begin{aligned}
|\nabla^2_{st} \ell(\boldsymbol{\beta}) - \nabla^2_{st} \ell(\boldsymbol{\beta}^*)| &= \binom{n}{2}^{-1} \sum_{1 \leq i < j \leq n} \frac{R_{ij}(\boldsymbol{\beta}) \cdot (y_i - y_j)^2 \cdot (x_{is} - x_{js}) \cdot (x_{it} - x_{jt}) \cdot (u_{ij} - 1)}{(1 + R_{ij}(\boldsymbol{\beta}))^2} \\
&\leq 2^{-1} |\nabla^2_{ss} \ell(\boldsymbol{\beta}^*) + \nabla^2_{tt} \ell(\boldsymbol{\beta}^*)| \cdot \max_{1 \leq i < j \leq n} |\max\{w_{ij}, w_{ij}^{-1}\} - 1|. \quad (5.7)
\end{aligned}$$

By Holder's inequality, we have

$$\sup_{\|\boldsymbol{\beta}-\boldsymbol{\beta}^*\|_1 \leq \delta_n} \max_{i<j} |(y_i - y_j) \cdot \boldsymbol{\Delta}^T(\boldsymbol{x}_i - \boldsymbol{x}_j)| \leq M \cdot \|\boldsymbol{\Delta}\|_1 = \mathcal{O}_\mathbb{P}(M \cdot \delta_n) = o_\mathbb{P}(1),$$

and $\sup_{\|\boldsymbol{\beta}-\boldsymbol{\beta}^*\|_1 \leq \delta_n} \max_{1 \leq i < j \leq n} |\max\{w_{ij}, w_{ij}^{-1}\} - 1| = \mathcal{O}_\mathbb{P}(M \cdot \delta_n)$. Thus, by (5.7),

$$\sup_{\|\boldsymbol{\beta}-\boldsymbol{\beta}^*\|_1 \leq \delta_n} \|\nabla^2 \ell(\boldsymbol{\beta}) - \nabla^2 \ell(\boldsymbol{\beta}^*)\|_\infty \leq \left\{\|\nabla^2 \ell(\boldsymbol{\beta}^*) + \mathbf{H}\|_\infty + \|\mathbf{H}\|_\infty\right\} \cdot \mathcal{O}_\mathbb{P}(M \cdot \delta_n). \quad (5.8)$$

By Assumption 4.2, $\|\mathbf{H}\|_\infty = \mathcal{O}(1)$. It remains to control $\|\nabla^2 \ell(\boldsymbol{\beta}^*) + \mathbf{H}\|_\infty$. Let $\bar{\mathbf{r}}_{ij} = \mathbf{T}_{ij} - \mathbb{E}(\mathbf{T}_{ij})$, where

$$\mathbf{T}_{ij} = \frac{R_{ij}(\boldsymbol{\beta}^*) \cdot (y_i - y_j)^2 \cdot (\boldsymbol{x}_i - \boldsymbol{x}_j)^{\otimes 2}}{(1 + R_{ij}(\boldsymbol{\beta}^*))^2}.$$

Then $\nabla^2 \ell(\boldsymbol{\beta}^*) + \mathbf{H} = -\frac{2}{n(n-1)} \cdot \sum_{i<j} \bar{\mathbf{r}}_{ij}$ is a mean-zero second order U-statistic with kernel function $\bar{\mathbf{r}}_{ij}$. For any $1 \leq a, b \leq d$, $\bar{\mathbf{r}}_{ij}$ satisfies $[\bar{\mathbf{r}}_{ij}]_{(a,b)} \leq 2 \cdot M^2$. The Hoeffding inequality yields, for any $x > 0$,

$$\mathbb{P}\left(\left|\nabla^2_{ab} \ell(\boldsymbol{\beta}^*) + \mathbf{H}_{a,b}\right| > x\right) \leq 2 \cdot \exp\left(-\frac{k \cdot x^2}{8 \cdot M^4}\right),$$

where $k = \lfloor n/2 \rfloor$. Taking $x = M^2 \cdot \sqrt{\log d/n}$, by the union bound, we get with high probability, $\|\nabla^2 \ell(\boldsymbol{\beta}^*) + \mathbf{H}\|_\infty \leq M^2 \cdot \sqrt{\log d/n}$. Together with (5.8), we complete the proof of Lemma 5.4. $\square$

By Lemma 5.4 and the fact that $\|\bar{\boldsymbol{\gamma}} - \boldsymbol{\gamma}^*\|_1 \leq \|\widehat{\boldsymbol{\Delta}}_{\boldsymbol{\gamma}}\|_1 = \mathcal{O}_{\mathbb{P}}(\max\{s, s_1\} \cdot \sqrt{\log d/n})$, we have

$$\begin{aligned}
\|\nabla^2_{\alpha\gamma}\ell(\widehat{\boldsymbol{\beta}}) - \nabla^2_{\alpha\gamma}\ell(\alpha^*, \bar{\boldsymbol{\gamma}})\|_\infty &\leq \|\nabla^2_{\alpha\gamma}\ell(\widehat{\boldsymbol{\beta}}) - \nabla^2_{\alpha\gamma}\ell(\boldsymbol{\beta}^*)\|_\infty + \|\nabla^2_{\alpha\gamma}\ell(\alpha^*, \bar{\boldsymbol{\gamma}}) - \nabla^2_{\alpha\gamma}\ell(\boldsymbol{\beta}^*)\|_\infty \\
&= \mathcal{O}_{\mathbb{P}}\Big(M \cdot \max\{s, s_1\}\sqrt{\frac{\log d}{n}}\Big).
\end{aligned} \quad (5.9)$$

Now, we can bound $I_{22}$ by using (5.9):

$$I_{22} \leq \|\nabla^2_{\alpha\gamma}\ell(\widehat{\boldsymbol{\beta}}) - \nabla^2_{\alpha\gamma}\ell(\alpha^*, \bar{\boldsymbol{\gamma}})\|_\infty \cdot \|\widehat{\boldsymbol{\Delta}}_{\boldsymbol{\gamma}}\|_1 = \mathcal{O}_{\mathbb{P}}\Big(M \cdot \max\{s, s_1\}^2 \cdot \frac{\log d}{n}\Big) = o_{\mathbb{P}}\Big(\frac{1}{\sqrt{n}}\Big).$$

Following the similar arguments as in (5.9), we can prove that

$$\|\nabla^2_{\gamma\gamma}\ell(\widehat{\boldsymbol{\beta}}) - \nabla^2_{\gamma\gamma}\ell(\alpha^*, \widetilde{\boldsymbol{\gamma}})\|_\infty = \mathcal{O}_{\mathbb{P}}\Big(M \cdot \max\{s, s_1\} \cdot \sqrt{\frac{\log d}{n}}\Big). \quad (5.10)$$

In addition, Lemma 5.1 implies that $\|\widehat{\mathbf{w}} - \mathbf{w}^*\|_1 = \mathcal{O}_{\mathbb{P}}(s_1 \lambda_s) = o_{\mathbb{P}}(1)$. Together with (5.10), we have

$$\begin{aligned}
I_{23} &\leq \|\widehat{\mathbf{w}}\|_1 \cdot \|\nabla^2_{\gamma\gamma}\ell(\widehat{\boldsymbol{\beta}}) - \nabla^2_{\gamma\gamma}\ell(\alpha^*, \widetilde{\boldsymbol{\gamma}})\|_\infty \cdot \|\widehat{\boldsymbol{\Delta}}_{\boldsymbol{\gamma}}\|_1 \\
&= (\|\mathbf{w}^*\|_1 + o_{\mathbb{P}}(1)) \cdot \mathcal{O}_{\mathbb{P}}\Big(M \cdot \max\{s, s_1\}\sqrt{\frac{\log d}{n}}\Big) \cdot \mathcal{O}_{\mathbb{P}}\Big(\max\{s, s_1\} \cdot \sqrt{\frac{\log d}{n}}\Big) = o_{\mathbb{P}}\Big(\frac{1}{\sqrt{n}}\Big).
\end{aligned}$$

Thus, we have proved the rate of convergence of $n^{1/2} \cdot |\widehat{\ell}'(\alpha^*) - S(\boldsymbol{\beta}^*)|$, i.e.,

$$n^{1/2} \cdot |\widehat{\ell}'(\alpha^*) - S(\boldsymbol{\beta}^*)| = \mathcal{O}_{\mathbb{P}}\Big(\lambda_s \cdot \max\{s, s_1\} \cdot \sqrt{\log d}\Big) = o_{\mathbb{P}}(1). \quad (5.11)$$

**Step 2: Characterize the limiting distribution of $S(\boldsymbol{\beta}^*)$.** We provide the following Lemma on the central limit theorem for U-statistics with increasing dimensions.

**Lemma 5.5.** Under Assumption 4.2, for any $\mathbf{b} \in \mathbb{R}^d$ with $\|\mathbf{b}\|_0 \leq \widetilde{s}$ and $\|\mathbf{b}\|_2 = 1$, if $\widetilde{s}^{3/2} \cdot n^{-1/2} \cdot M^3 = o_{\mathbb{P}}(1)$ where $M := \max_{1 \leq i < j \leq n} \|(y_i - y_j) \cdot (\boldsymbol{x}_i - \boldsymbol{x}_j)\|_\infty$, then

$$\frac{\sqrt{n}}{2} \cdot (\mathbf{b}^T \boldsymbol{\Sigma} \mathbf{b})^{-1/2} \cdot \mathbf{b}^T \nabla \ell(\boldsymbol{\beta}^*) \rightsquigarrow N(0, 1).$$

*Proof.* The Lemma is proved by applying the Hoeffding's decomposition,

$$\begin{aligned}
&\frac{\sqrt{n}}{2} \cdot (\mathbf{b}^T \boldsymbol{\Sigma} \mathbf{b})^{-1/2} \cdot \mathbf{b}^T \nabla \ell(\boldsymbol{\beta}^*) \\
&= (\mathbf{b}^T \boldsymbol{\Sigma} \mathbf{b})^{-1/2} \cdot \frac{1}{\sqrt{n}} \cdot \sum_{i=1}^n \mathbf{b}^T \mathbf{g}(y_i, \boldsymbol{x}_i, \boldsymbol{\beta}^*) + \frac{\sqrt{n}}{2} \cdot (\mathbf{b}^T \boldsymbol{\Sigma} \mathbf{b})^{-1/2} \cdot \mathbf{b}^T \{\nabla \ell(\boldsymbol{\beta}^*) - \widehat{\mathbf{U}}_n\},
\end{aligned}$$

where $\mathbf{g}(y_i, \boldsymbol{x}_i, \boldsymbol{\beta}^*)$ and $\widehat{\mathbf{U}}_n$ are defined in (3.11). We can verify that the Lyapunov central limit theorem for independent random variables can be applied for the first term under the assumption that $\widetilde{s}^{3/2} \cdot n^{-1/2} \cdot M^3 = o_{\mathbb{P}}(1)$. The remaining proof requires more careful calculation of the moment of approximation error $\mathbf{b}^T(\nabla \ell(\boldsymbol{\beta}^*) - \widehat{\mathbf{U}}_n)$ in the Hájek projection, because here we allow the intrinsic dimension $\widetilde{s}$ to scale with $n$. We defer the detailed proof to Appendix B. □

Since $S(\boldsymbol{\beta}^*)$ is a sparse linear combination of the U-statistic $\nabla\ell(0, \boldsymbol{\gamma}^*)$ and $\|\mathbf{w}^*\|_0 = s_1$, with $\mathbf{b} = (1, -\mathbf{w}^{*T})^T$, Lemma 5.5 implies that

$$n^{1/2} \cdot S(\boldsymbol{\beta}^*)/(2\sigma) \rightsquigarrow N(0,1), \quad \text{where} \quad \sigma^2 = \boldsymbol{\Sigma}_{\alpha\alpha} - 2\mathbf{w}^{*T}\boldsymbol{\Sigma}_{\gamma\alpha} + \mathbf{w}^{*T}\boldsymbol{\Sigma}_{\gamma\gamma}\mathbf{w}^*. \tag{5.12}$$

**Step3: Show the convergence of $\widehat{\ell}''(\bar{\alpha})$ for any $\bar{\alpha}$ between 0 and $\widehat{\alpha}^P$.** We now show that $|\widehat{\ell}''_n(\bar{\alpha}) + H_{\alpha|\gamma}| = o_{\mathbb{P}}(1)$. By chain rule, we have

$$\begin{aligned}\widehat{\ell}''_n(\bar{\alpha}) &= \nabla^2_{\alpha\alpha}\ell(\bar{\alpha}, \widehat{\boldsymbol{\gamma}}(\bar{\alpha})) - 2\nabla^2_{\alpha\gamma}\ell(\bar{\alpha}, \widehat{\boldsymbol{\gamma}}(\bar{\alpha}))^T\widehat{\mathbf{w}} + \widehat{\mathbf{w}}^T\nabla^2_{\gamma\gamma}\ell(\bar{\alpha}, \widehat{\boldsymbol{\gamma}}(\bar{\alpha}))^T\widehat{\mathbf{w}} \\ &= (1, -\widehat{\mathbf{w}}^T)\nabla^2\ell(\bar{\alpha}, \widehat{\boldsymbol{\gamma}}(\bar{\alpha}))(1, -\widehat{\mathbf{w}}^T)^T. \end{aligned} \tag{5.13}$$

We then decompose $\widehat{\ell}''_n(\bar{\alpha}) + H_{\alpha|\gamma}$ into two terms, namely,

$$\widehat{\ell}''_n(\bar{\alpha}) + H_{\alpha|\gamma} = \underbrace{\widehat{\ell}''_n(\bar{\alpha}) - (1, -\widehat{\mathbf{w}}^T)\nabla^2\ell(\boldsymbol{\beta}^*)(1, -\widehat{\mathbf{w}}^T)^T}_{I_3} + \underbrace{(1, -\widehat{\mathbf{w}}^T)\nabla^2\ell(\boldsymbol{\beta}^*)(1, -\widehat{\mathbf{w}}^T)^T + H_{\alpha|\gamma}}_{I_4}. \tag{5.14}$$

For the first term, by (5.13) and Hölder's inequality, we get

$$|I_3| \leq \|\nabla^2\ell(\bar{\alpha}, \widehat{\boldsymbol{\gamma}}(\bar{\alpha})) - \nabla^2\ell(\boldsymbol{\beta}^*)\|_\infty (\|\widehat{\mathbf{w}}\|_1 + 1)^2.$$

To control the difference between the Hessian matrices, we apply Lemma 5.4. Let $\bar{\boldsymbol{\Delta}} = (\bar{\alpha}, \widehat{\boldsymbol{\gamma}}(\bar{\alpha})^T)^T - \boldsymbol{\beta}^*$. We have

$$\|\bar{\boldsymbol{\Delta}}\|_1 = |\bar{\alpha} - \alpha^*| + \|\widehat{\boldsymbol{\gamma}} - \boldsymbol{\gamma}^* + (\widehat{\alpha} - \bar{\alpha})\widehat{\mathbf{w}}\|_1 \leq |\bar{\alpha} - \alpha^*| + \|\widehat{\boldsymbol{\gamma}} - \boldsymbol{\gamma}^*\|_1 + |\widehat{\alpha} - \bar{\alpha}|\|\widehat{\mathbf{w}}\|_1. \tag{5.15}$$

To control $|\bar{\alpha} - \alpha^*|$, we need a rough bound on the rate of convergence of the post-regularization estimator $\widehat{\alpha}^P - \alpha^*$. The following lemma serves our purpose:

**Lemma 5.6.** *Under the conditions in Theorem 4.1, we have*

$$|\widehat{\alpha}^P - \alpha^*| = \mathcal{O}_{\mathbb{P}}\Big(\|\mathbf{w}^*\|_1 \cdot M \cdot \sqrt{\frac{\log n}{n}}\Big).$$

By Lemma 5.6, we have $|\bar{\alpha} - \alpha^*| \leq |\widehat{\alpha}^P - \alpha^*| = \mathcal{O}_{\mathbb{P}}(\|\mathbf{w}^*\|_1 \cdot M \cdot \sqrt{\log n/n}) = \mathcal{O}_{\mathbb{P}}(\sqrt{s_1} \cdot M \cdot \sqrt{\log n/n})$. Moreover, we have $\|\widehat{\boldsymbol{\gamma}} - \boldsymbol{\gamma}^*\|_1 = \mathcal{O}_{\mathbb{P}}(s \cdot \sqrt{\log d/n})$, $\|\widehat{\mathbf{w}}\|_1 = \|\mathbf{w}^*\|_1 + o_{\mathbb{P}}(1)$ and that

$$|\widehat{\alpha} - \bar{\alpha}| \leq |\widehat{\alpha} - \alpha^*| + |\bar{\alpha} - \alpha^*| = \mathcal{O}_{\mathbb{P}}\Big(M \cdot \max\{s^{1/2}, s_1^{1/2}\} \cdot \sqrt{\frac{\log(d \vee n)}{n}}\Big).$$

Putting together the above results and by (5.15), we conclude that $\|\bar{\boldsymbol{\Delta}}\|_1 = \mathcal{O}_{\mathbb{P}}(\sqrt{s_1} \cdot M \cdot \sqrt{\log n/n} + s \cdot \sqrt{\log d/n} + M \cdot \max\{\sqrt{s \cdot s_1}, s_1\} \cdot \sqrt{\log d/n}) = \mathcal{O}_{\mathbb{P}}(M \cdot \max\{s, s_1\} \cdot \sqrt{\log(d \vee n)/n})$. Therefore, applying Lemma 5.4, we have $\|\nabla^2\ell(\bar{\alpha}, \widehat{\boldsymbol{\gamma}}(\bar{\alpha})) - \nabla^2\ell(\boldsymbol{\beta}^*)\|_\infty = \mathcal{O}_{\mathbb{P}}(M^2 \cdot \max\{s, s_1\} \cdot \sqrt{\log(d \vee n)/n})$. Therefore, we conclude that

$$|I_3| = \mathcal{O}_{\mathbb{P}}\big(\|\mathbf{w}^*\|_1^2 \cdot M^2 \cdot \max\{s, s_1\} \cdot \sqrt{\log(d \vee n)/n}\big) = o_{\mathbb{P}}(1). \tag{5.16}$$

We now focus on $I_4$, which can be decomposed into the following terms:

$$I_4 = \underbrace{\{\nabla^2_{\alpha\alpha}\ell(\boldsymbol{\beta}^*) + \mathbf{H}_{\alpha\alpha}\}}_{I_{41}} - 2\underbrace{\{\widehat{\mathbf{w}}^T\nabla^2_{\alpha\gamma}\ell(\boldsymbol{\beta}^*) + \mathbf{w}^{*T}\mathbf{H}_{\alpha\gamma}\}}_{I_{42}} + \underbrace{\{\widehat{\mathbf{w}}^T\nabla^2_{\gamma\gamma}\ell(\boldsymbol{\beta}^*)\widehat{\mathbf{w}} + \mathbf{w}^{*T}\mathbf{H}_{\gamma\gamma}\mathbf{w}^*\}}_{I_{43}}.$$

By the proof of Lemma 5.4, we have $\|\nabla^2\ell(\boldsymbol{\beta}^*)+\mathbf{H}\|_\infty = \mathcal{O}_\mathbb{P}(M^2 \cdot \sqrt{\log d/n})$. Hence $I_{41} = \mathcal{O}_\mathbb{P}(M^2 \cdot \sqrt{\log d/n}) = o_\mathbb{P}(1)$. For the second term, it holds that $I_{42} = \widehat{\mathbf{w}}^T(\nabla^2_{\alpha\boldsymbol{\gamma}}\ell(\boldsymbol{\beta}^*)+\mathbf{H}_{\alpha\boldsymbol{\gamma}}) - (\widehat{\mathbf{w}}-\mathbf{w}^*)^T\mathbf{H}_{\alpha\boldsymbol{\gamma}}$. We have $|\widehat{\mathbf{w}}^T(\nabla^2_{\alpha\boldsymbol{\gamma}}\ell(\boldsymbol{\beta}^*)+\mathbf{H}_{\alpha\boldsymbol{\gamma}})| \leq \|\widehat{\mathbf{w}}\|_1\|\nabla^2_{\alpha\boldsymbol{\gamma}}\ell(\boldsymbol{\beta}^*)+\mathbf{H}_{\alpha\boldsymbol{\gamma}}\|_\infty = \mathcal{O}_\mathbb{P}(\|\mathbf{w}^*\|_1 \cdot M^2 \cdot \sqrt{\log d/n})$, and $|(\widehat{\mathbf{w}}-\mathbf{w}^*)^T\mathbf{H}_{\alpha\boldsymbol{\gamma}}| \leq \|\widehat{\mathbf{w}}-\mathbf{w}^*\|_1\|\mathbf{H}_{\alpha\boldsymbol{\gamma}}\|_\infty = o_\mathbb{P}(1)$. Therefore, we conclude that $|I_{42}| = o_\mathbb{P}(1)$. For the term $I_{43}$, we apply similar arguments to get $I_{43} = \mathcal{O}_\mathbb{P}\big(\|\mathbf{w}^*\|_1^2 \cdot M^2 \cdot \sqrt{\log d/n} + \|\mathbf{w}^*\|_1 \cdot s_1 \cdot \sqrt{\log d/n}\big) = o_\mathbb{P}(1)$. Hence, we conclude that $I_4 = o_\mathbb{P}(1)$. Together with (5.16), this implies

$$|\widehat{\ell}''_n(\bar{\alpha}) + H_{\alpha|\boldsymbol{\gamma}}| = o_\mathbb{P}(1). \tag{5.17}$$

Given (5.11), (5.12), (5.17), we now wrap up the whole proof. By first-order optimality condition, we have $\widehat{\ell}'(\widehat{\alpha}^P) = 0$. Applying mean-value theorem, we get $\widehat{\ell}'(\widehat{\alpha}^P) = \widehat{\ell}'(\alpha^*) + \widehat{\ell}''(\bar{\alpha})(\widehat{\alpha}^P - \alpha^*)$, where $\bar{\alpha}$ is an intermediate value between $\widehat{\alpha}^P$ and $\alpha^*$. This implies

$$\widehat{\alpha}^P - \alpha^* = \widehat{\ell}''(\bar{\alpha})^{-1}\widehat{\ell}'(\alpha^*). \tag{5.18}$$

Finally, combining (5.18), (5.11), (5.12), (5.17) and applying Slutsky's theorem, we have $n^{1/2}(\widehat{\alpha}^P - \alpha^*) = -H_{\alpha|\boldsymbol{\gamma}}^{-1} \cdot n^{1/2}S(\boldsymbol{\beta}^*) + o_\mathbb{P}(1)$. We complete the proof of Theorem 4.1.

## 6 Extensions to Missing Data and Selection Bias

In this section, we will illustrate how the semiparametric GLM is useful for handling high dimensional data with missing values and selection bias.

Assume that $Y$ given $\boldsymbol{X}$ follows the GLMs in equation (1.1) and the missing data process is decomposable. As shown in equation (2.3), $Y$ given $\boldsymbol{X}$ and $\delta = 1$ satisfies the semiparametric GLM with the same finite dimensional parameter $\boldsymbol{\beta}$ and unknown function $f^m(\cdot)$. Following the same regularized statistical chromatography arguments in Section 3, the conditional probability of $\mathbf{R}_{ij}^L = \boldsymbol{r}_{ij}^L$ given the order statistic $\boldsymbol{Y}_{(i,j)}^L = (\min(Y_i, Y_j), \max(Y_i, Y_j))$, the covariate $(\boldsymbol{x}_i, \boldsymbol{x}_j)$ and the selection indicator $\delta_i = \delta_j = 1$ is

$$\mathbb{P}(\mathbf{R}_{ij}^L = \boldsymbol{r}_{ij}^L \mid \boldsymbol{y}_{(i,j)}^L, \boldsymbol{x}_i, \boldsymbol{x}_j, \delta_i = \delta_j = 1; \boldsymbol{\beta}) = \{1 + R_{ij}(\boldsymbol{\beta})\}^{-1},$$

where $R_{ij}(\boldsymbol{\beta})$ is given by (3.4). Thus, the composite likelihood function becomes

$$\ell^m(\boldsymbol{\beta}) = -\binom{n}{2}^{-1} \sum_{1 \leq i \leq j \leq n} \delta_i \cdot \delta_j \cdot \log\Big(1 + R_{ij}(\boldsymbol{\beta})\Big).$$

Note that the subjects $i$ and $j$ contribute to the loss function if and only if they are both completely observed, i.e., $\delta_i = \delta_j = 1$. Hence, $\ell^m(\boldsymbol{\beta})$ is expressed in terms of the observed data and is computable in practice. The initial estimator is given by

$$\widehat{\boldsymbol{\beta}}_m \in \underset{\boldsymbol{\beta}}{\text{argmax}} \;\; \ell^m(\boldsymbol{\beta}) - \sum_{j=1}^d p_{\lambda_m}(\beta_j), \tag{6.1}$$

where $\lambda_m \geq 0$ is a tuning parameter and $p_\lambda(\cdot)$ is a generic penalty function (which could be nonconvex). Further discussions on the parameter estimation in the presence of missing data can be found in appendix. In the following, we apply the main results in Section 4 to establish the limiting distribution of the maximum directional likelihood estimator and directional likelihood ratio test statistic under the null hypothesis $H_0 : \alpha^* = 0$.

Let $\widehat{\boldsymbol{\beta}}_m = (\widehat{\alpha}_m, \widehat{\boldsymbol{\gamma}}_m)$. Consider the following directional likelihood function $\widehat{\ell}_m(\alpha) = \ell^m(\alpha, \widehat{\boldsymbol{\gamma}}_m + (\widehat{\alpha}_m - \alpha)\widehat{\mathbf{w}}_m)$, where

$$\widehat{\mathbf{w}}_m = \arg\min ||\mathbf{w}||_1 \quad \text{subject to} \quad ||\nabla^2_{\alpha\gamma}\ell^m(\widehat{\boldsymbol{\beta}}_m) - \mathbf{w}^T \nabla^2_{\gamma\gamma}\ell^m(\widehat{\boldsymbol{\beta}}_m)||_\infty \leq \lambda_{ms}. \tag{6.2}$$

Let $\alpha_m^P = \arg\max_{\alpha \in \mathbb{R}} \widehat{\ell}_m(\alpha)$, and $\Lambda_n^m = 2n\{\widehat{\ell}_m(\alpha_m^P) - \widehat{\ell}_m(\alpha_0)\}$. We now establish the asymptotic properties of $\widehat{\alpha}_m^P - \alpha^*$ and $\Lambda_n^m$ under the null hypothesis $H_0 : \alpha^* = 0$. The following assumptions, analogous to Assumptions 4.2, 4.3, and 4.4, are adopted for the missing data setting, and we refer to Section 4 for more detailed discussions.

**Assumption 6.1.** Let $\mathbf{H}_m = -\mathbb{E}(\nabla^2 \ell^m(\boldsymbol{\beta}^*))$, $(H_m)_{\alpha|\gamma} = (H_m)_{\alpha\alpha} - (\mathbf{H}_m)_{\alpha\gamma}[(\mathbf{H}_m)_{\gamma\gamma}]^{-1}(\mathbf{H}_m)_{\gamma\alpha}$ and $\boldsymbol{\Sigma}_m = \mathbb{E}\{\mathbf{g}_m^{\otimes 2}(y_i, \boldsymbol{x}_i, \boldsymbol{\beta}^*)\}$, where $\mathbf{g}_m(y_i, \boldsymbol{x}_i, \boldsymbol{\beta}) = \frac{n}{2} \cdot \mathbb{E}\{\nabla \ell^m(\boldsymbol{\beta}) \mid y_i, \boldsymbol{x}_i, \delta_i = 1\}$. Assume that $\lambda_{\min}(\boldsymbol{\Sigma}_m) \geq c$ and $\lambda_{\min}(\mathbf{H}_m) \geq c$, $\lambda_{\max}(\mathbf{H}_m) \leq c'$ for some constants $c, c' > 0$, $||\mathbf{H}_m||_\infty = \mathcal{O}(1)$, $(H_m)_{\alpha|\gamma} = \mathcal{O}(1)$ and $(H_m)_{\alpha|\gamma}^{-1} = \mathcal{O}(1)$.

**Assumption 6.2.** Let $\mathbf{w}_m^* = [(\mathbf{H}_m)_{\gamma\gamma}]^{-1}(\mathbf{H}_m)_{\gamma\alpha}$. Assume that $s_1 = ||\mathbf{w}_m^*||_0$ and $s_1^{3/2} \cdot n^{-1/2} \cdot M^3 = o_\mathbb{P}(1)$, where $M := \max_{1 \leq i < j \leq n} ||\delta_i \cdot \delta_j \cdot (y_i - y_j) \cdot (\boldsymbol{x}_i - \boldsymbol{x}_j)||_\infty$.

**Assumption 6.3.** Assume that $||\widehat{\boldsymbol{\beta}}_m - \boldsymbol{\beta}^*||_1 = \mathcal{O}_\mathbb{P}(s \cdot \sqrt{\log d/n})$ and $||\widehat{\boldsymbol{\beta}}_m - \boldsymbol{\beta}^*||_2 = \mathcal{O}_\mathbb{P}(s^{1/2} \cdot \sqrt{\log d/n})$, where $s = ||\boldsymbol{\beta}^*||_0$. It holds that $\max\{s, s_1\} \cdot \lambda_{ms} \cdot \sqrt{\log d} = o(1)$, where $s_1$ is defined in Assumption 6.2, and $\lambda_{ms} \asymp ||\mathbf{w}_m^*||_1 \cdot M \cdot (s + M) \cdot \sqrt{\log d/n}$.

The following hypothesis testing result is a direct corollary of Theorems 4.1 and 4.2.

**Corollary 6.1.** Under Assumptions 4.1, 6.1, 6.2 and 6.3, we have

$$n^{1/2} \cdot (\widehat{\alpha}^P - \alpha^*) \rightsquigarrow N(0, 4\sigma_m^2 \cdot (H_m)_{\alpha|\gamma}^{-2}), \quad \text{where} \quad \sigma_m^2 = (\boldsymbol{\Sigma}_m)_{\alpha\alpha} - 2\mathbf{w}_m^{*T}(\boldsymbol{\Sigma}_m)_{\gamma\alpha} + \mathbf{w}_m^{*T}(\boldsymbol{\Sigma}_m)_{\gamma\gamma}\mathbf{w}_m^*.$$

Also, under the null hypothesis, it holds $(4 \cdot \sigma_m^2)^{-1} \cdot (H_m)_{\alpha|\gamma} \cdot \Lambda_n \rightsquigarrow \chi_1^2$. Moreover, assume that the following conditions hold,

$$||(\boldsymbol{\Sigma}_m)_{\gamma\gamma}||_\infty = \mathcal{O}(1), \quad ||(\boldsymbol{\Sigma}_m)_{\alpha\gamma}||_\infty = \mathcal{O}(1), \quad \text{and} \quad ||\mathbf{w}_m^*||_1^2 \cdot M^3 \cdot s \cdot \sqrt{\log d/n} = o(1).$$

Then the directional likelihood ratio test statistic satisfies $(4 \cdot \widehat{\sigma}_m^2)^{-1} \cdot (\widehat{H}_m)_{\alpha|\gamma} \cdot \Lambda_n \rightsquigarrow \chi_1^2$, where

$$\widehat{\sigma}_m^2 := (\widehat{\boldsymbol{\Sigma}}_m)_{\alpha\alpha} - 2\widehat{\mathbf{w}}_m^T(\widehat{\boldsymbol{\Sigma}}_m)_{\gamma\alpha} + \widehat{\mathbf{w}}_m^T(\widehat{\boldsymbol{\Sigma}}_m)_{\gamma\gamma}\widehat{\mathbf{w}}_m,$$

$$\widehat{\boldsymbol{\Sigma}}_m := \frac{1}{n} \cdot \sum_{i=1}^n \left\{ \frac{1}{n-1} \cdot \sum_{j=1, j \neq i}^n \frac{\delta_i \cdot \delta_j \cdot R_{ij}(\widehat{\boldsymbol{\beta}}_m) \cdot (y_i - y_j) \cdot (\boldsymbol{x}_i - \boldsymbol{x}_j)}{1 + R_{ij}(\widehat{\boldsymbol{\beta}}_m)} \right\}^{\otimes 2},$$

and $(\widehat{H}_m)_{\alpha|\gamma} := -\nabla^2_{\alpha\alpha}\ell^m(\widehat{\boldsymbol{\beta}}_m) + \widehat{\mathbf{w}}_m^T \nabla^2_{\gamma\alpha}\ell^m(\widehat{\boldsymbol{\beta}}_m)$.

The proof of Corollary 6.1 is provided in Supplementary Appendix D.

# 7 Numerical Results

In this section we provide synthetic and real data examples to back up the theoretical results.

## 7.1 Simulation Studies

We conduct simulation studies to assess the finite sample performance of the proposed methods. We generate the outcomes from (1) the linear regression with the standard Gaussian noise or (2) the logistic regression, and the covariates from $N(0, \boldsymbol{\Sigma})$, where $\Sigma_{ij} = 0.6^{|i-j|}$. The true values of $\boldsymbol{\beta}$ are $\boldsymbol{\beta}_j^* = \mu$ for $j = 1, 2, 3$ and $\boldsymbol{\beta}_j^* = 0$ for $j = 4, ..., d$. Thus, the cardinality of the support set of $\boldsymbol{\beta}^*$ is $s = 3$. The sample size is $n = 100$, the number of covariates is $d = 200$, and the number of simulation replications is 500.

We calculate the $\ell_1$-regularized estimator $\widehat{\boldsymbol{\beta}}$ by using the `glmnet` package in R. In particular, we determine the regularization parameter $\lambda$ by minimizing the K-fold cross validated loss function,

$$\mathrm{CV}(\lambda) = \sum_{k=1}^{K} \{\ell(\widehat{\boldsymbol{\beta}}_\lambda^{(-k)}) - \ell^{(-k)}(\widehat{\boldsymbol{\beta}}_\lambda^{(-k)})\},$$

where $\ell^{(-k)}$ stands for the loss function evaluated without the $k$th subset and similarly $\widehat{\boldsymbol{\beta}}_\lambda^{(-k)}$ stands for the regularized estimator derived without using the $k$th subset. In the simulation studies, we use 5-fold cross validation. The tuning parameter for the Dantzig selector $\lambda_s$ in (3.10) is chosen by $4\sqrt{\log(nd)/n}$. We find that the simulation results are not sensitive to the choice of $\lambda_s$. We only present the results with the Lasso penalty. Similar results are observed by using the folded concave penalty based on the LLA algorithm (Fan et al., 2012).

For the linear regression, we consider the directional likelihood ratio test (DLRT) and the Wald test based on the asymptotic normality of $\widehat{\alpha}^P$, as well as the desparsifying method in van de Geer et al. (2014); Zhang and Zhang (2014) and debias method in Javanmard and Montanari (2013). Both of these two methods are tailored for the linear regression with the $L_2$ loss and are optimal for confidence intervals and hypothesis testing. To examine the validity of our tests, we report their type I errors for the null hypothesis $H_0 : \beta_1 = \mu$ with various choices of $\mu \in [0, 1]$ at the 0.05 significance level. The results are summarized in Table 7.1. We find that, our Wald test and DLRT yield accurate type I errors, which are comparable to the desparsifying and debias methods. In addition, we also compare the powers of these tests. In particular, we test the null hypothesis $H_0 : \beta_1 = 0$, but increase $\mu$ from 0 to 1 in the data generating procedure. As shown in the left panel of Figure 7.1, our Wald test and DLRT based on the semiparametric GLM are nearly as efficient as the desparsifying and debias methods. Such results show that the semiparametric GLM gains model flexibility by losing little inferential efficiency.

For the logistic model, we only consider the desparsifying method, because the debias method is not defined. As shown in Table 7.1, our proposed tests yield well controlled type I errors. Similarly, the power comparison for testing $H_0 : \beta_1 = 0$ in Figure 7.1 reveals that our tests under the more flexible semiparametric model are comparable to the desparsifying method. Moreover, the DLRT is more powerful than the remaining two tests, which demonstrates the numerical advantages of the likelihood ratio inference over the Wald type tests. This observation is also consistent with the literature for low dimensional inference.

To further demonstrate the advantage of the proposed methods, we consider the data with missing values. Similar to the previous data generating procedures, we first simulate the original data $Y_i$ and $\boldsymbol{X}_i$. Then, for the linear regression, we consider the following two scenarios to create missing values: (1) the response $Y_i$ is observed (i.e., $\delta_i = 1$) if and only if $Y_i \leq 0$; and (2) $Y_i$ is always observed if $Y_i \leq 0$ and observed with probability 0.2 if $Y_i > 0$, i.e., $\mathbb{P}(\delta_i = 1 \mid Y_i, \boldsymbol{X}_i) =$ $1 - 0.8I(Y_i > 0)$. For the logistic regression, we also consider two scenarios to create missing values: (1) $\mathbb{P}(\delta_i = 1 \mid Y_i, \boldsymbol{X}_i) = 0.2 + 0.6Y_i$; and (2) $\mathbb{P}(\delta_i = 1 \mid Y_i, \boldsymbol{X}_i) = 0.2 + 0.8Y_i$. Since the desparsifying and debias methods are developed based on the assumption that no missing values exist, we consider the following two practical procedures for handling missing data on $Y$. The first approach is that we apply the desparsifying and debias methods directly to samples with $Y$ observed, which is known as the complete-case analysis. The second approach is that we apply these two methods to an imputed dataset. More specifically, for those samples with missing values on $Y$, we impute $Y$ by using the k-nearest neighbors method in Troyanskaya et al. (2001), implemented by the R function `impute.knn`. The type I errors are shown in Table 7.2. As expected, for the desparsifying and debias methods, the type I errors of the complete-case analysis are far from the 0.05 significance level. Although the imputation method shows some advantages over the complete-case analysis, similar patterns are observed. Therefore, in the presence of missing data, the existing methods cannot produce any result that is statistically reliable. In contrast, the type I errors based on the proposed tests are well controlled, and they are robust to the missing data and selection bias. The same conclusion holds under all simulation scenarios.

Table 7.1: Type I errors of the Wald test and directional likelihood ratio test (DLRT), the desparsifying and debias methods for linear and logistic regressions for $H_0 : \alpha = \mu$, at the 0.05 significance level, where $\mu = 0.00, ..., 1.00$.

| Model | Method | 0.00 | 0.10 | 0.20 | 0.40 | 0.60 | 0.80 | 1.00 |
|---|---|---|---|---|---|---|---|---|
| Linear | Wald | 0.048 | 0.066 | 0.060 | 0.052 | 0.054 | 0.046 | 0.054 |
|  | DLRT | 0.040 | 0.052 | 0.064 | 0.042 | 0.032 | 0.034 | 0.040 |
|  | Desparsity | 0.044 | 0.054 | 0.058 | 0.044 | 0.058 | 0.058 | 0.056 |
|  | Debias | 0.034 | 0.030 | 0.036 | 0.024 | 0.028 | 0.028 | 0.028 |
| Logistic | Wald | 0.054 | 0.060 | 0.054 | 0.054 | 0.066 | 0.068 | 0.038 |
|  | DLRT | 0.052 | 0.048 | 0.058 | 0.056 | 0.054 | 0.050 | 0.038 |
|  | Desparsity | 0.052 | 0.044 | 0.058 | 0.046 | 0.050 | 0.058 | 0.058 |

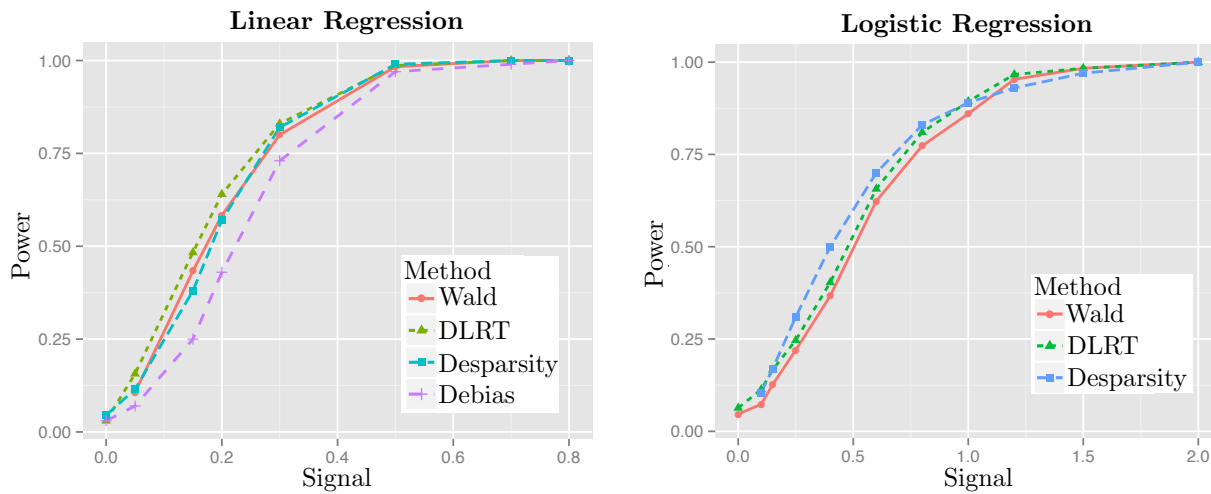

Figure 7.1: Power curves for testing $H_0 : \beta_1 = 0$ for the linear (left panel) and logistic (right panel) regressions at the 0.05 significance level.

In summary, our proposed testing procedures under the semiparametric GLM are as competitive as the existing methods even if the assumed model is correct. More importantly, in the presence of missing data or selection bias, the proposed methods significantly outperform the existing ones.

Table 7.2: Type I errors of the Wald test and directional likelihood ratio test (DLRT), the desparsifying method and debias method based on complete-case analysis (CC-) and imputation (Imp-) for linear and logistic regressions with missing data (selection bias) for $H_0 : \alpha = \mu$, at the 0.05 significance level, where $\mu = 0.10, ..., 0.25$.

| Scenario | Model | Method | 0.10 | 0.15 | 0.20 | 0.25 | 0.30 | 0.35 |
|---|---|---|---|---|---|---|---|---|
| 1 | Linear | Wald | 0.062 | 0.048 | 0.064 | 0.046 | 0.064 | 0.050 |
| | | DLRT | 0.056 | 0.042 | 0.060 | 0.036 | 0.056 | 0.048 |
| | | CC-Desparsity | 0.076 | 0.156 | 0.214 | 0.278 | 0.334 | 0.580 |
| | | Imp-Desparsity | 0.068 | 0.128 | 0.176 | 0.198 | 0.270 | 0.448 |
| | | CC-Debias | 0.126 | 0.322 | 0.488 | 0.662 | 0.820 | 0.900 |
| | | Imp-Debias | 0.108 | 0.260 | 0.306 | 0.438 | 0.470 | 0.624 |
| 1 | Logistic | Wald | 0.058 | 0.064 | 0.060 | 0.070 | 0.078 | 0.054 |
| | | DLRT | 0.044 | 0.052 | 0.044 | 0.054 | 0.052 | 0.042 |
| | | CC-Desparsity | 0.296 | 0.698 | 0.956 | 0.988 | 1.000 | 1.000 |
| | | Imp-Desparsity | 0.214 | 0.582 | 0.902 | 0.980 | 1.000 | 1.000 |
| 2 | Linear | Wald | 0.060 | 0.068 | 0.048 | 0.060 | 0.072 | 0.052 |
| | | DLRT | 0.060 | 0.062 | 0.040 | 0.048 | 0.052 | 0.046 |
| | | CC-Desparsity | 0.086 | 0.098 | 0.164 | 0.370 | 0.524 | 0.660 |
| | | Imp-Desparsity | 0.080 | 0.088 | 0.146 | 0.236 | 0.268 | 0.362 |
| | | CC-Debias | 0.072 | 0.152 | 0.334 | 0.530 | 0.728 | 0.804 |
| | | Imp-Debias | 0.070 | 0.096 | 0.148 | 0.308 | 0.376 | 0.442 |
| 2 | Logistic | Wald | 0.078 | 0.032 | 0.050 | 0.052 | 0.052 | 0.060 |
| | | DLRT | 0.074 | 0.022 | 0.040 | 0.044 | 0.042 | 0.046 |
| | | CC-Desparsity | 0.156 | 0.422 | 0.546 | 0.656 | 0.768 | 0.846 |
| | | Imp-Desparsity | 0.124 | 0.234 | 0.340 | 0.338 | 0.466 | 0.514 |

## 7.2 Analysis of Gene Expression Data

In this section, we apply the proposed tests to analyze the AGEMAP (Atlas of Gene Expression in Mouse Aging Project) gene expression data (Zahn et al., 2007). The dataset contains the expression values for 296 genes belonging to the mouse vascular endothelial growth factor (VEGF) signaling pathway. The sample size is $n = 40$. Among these 296 genes, we are interested in identifying genes that are significantly associated with the target gene Casp9. Thus, we treat the gene Casp9 as the response and the remaining 295 genes as covariates.

Since no missing value presents, we directly apply the desparsifying and debias methods to test $H_0 : \beta_j = 0$ for each $1 \leq j \leq 295$, under the linear model assumption. Similarly, we can assume that the gene Casp9 given the remaining variables follows the semiparametric GLM and the proposed Wald and likelihood ratio tests can be applied. To take into account of the multiplicity of tests, we use the method of Holm (1979) in the R function `p.adjust` to adjust the p-values. At the 0.05 significance level, all these four methods claim that gene Cdc42 is significant; see the first row of Table 7.3. This suggests that our tests are as effective as those existing procedures when there are no missing values.

To further illustrate the advantage of our methods in the presence of missing data, we create missing values for the outcome variable $Y_i$. More specifically, if $Y_i$ is among the top $M\%$ samples, where $M = 0, 15, 25$ and $35$, we remove the values of $Y_i$. Here, $M = 0$ means no missing data is created. This corresponds to the analysis of the original complete data. Similar to that in the simulation studies, the considered missing data mechanism depends on the unobserved values, which makes the analysis challenging.

The results are shown in Table 7.3, where the results based on the original complete data ($M = 0$) can be used as a benchmark. Based on the incomplete dataset, after the same adjustment for p-values, our Wald and likelihood ratio tests still select gene Cdc42, which are consistent with the results based on the original data. This pattern is preserved, even after 35% data are removed. For the desparsifying and debias methods, similar to the simulation studies, we can either apply them to those samples with only complete data (complete-case analysis) or the full data created by the imputation method (Troyanskaya et al., 2001). In particular, the CC-Desparsity and the CC-Debias methods consistently select no genes, when there exist missing data. This seems to suggest a lack of power for the existing methods based on the complete-case analysis. In addition, Imp-Desparsity tends to select very different genes at different levels of missing data percentage. They are all different from the benchmark gene Cdc42. Our analysis suggests that, the presence of missing values can dramatically change the results of Imp-Desparsity. Finally, Imp-Debias performs similarly to CC-Debias and tends to have low powers.

In conclusion, the existing methods based on the imputation methods or complete cases are either very sensitive to the missing data or have low powers. On the other hand, the proposed tests are quite robust and potentially more reliable in the presence of missing data.

Table 7.3: Significant genes selected by the Wald and directional likelihood ratio tests under the semiparametric GLM, the desparsifying method and debias method based on complete-case analysis (CC-) and imputation (Imp-) for the gene expression data. Here, $M\%$ samples are missing.

| M  | Wald  | DLRT  | CC-Desparsity | CC-Debias | Imp-Desparsity | Imp-Debias |
|----|-------|-------|---------------|-----------|----------------|------------|
| 0  | Cdc42 | Cdc42 | Cdc42         | Cdc42     | Cdc42          | Cdc42      |
| 15 | Cdc42 | Cdc42 | -             | -         | Mapk13         | -          |
| 25 | Cdc42 | Cdc42 | -             | -         | Ppp3cb         | -          |
| 35 | Cdc42 | Cdc42 | -             | -         | Nfatc3,Ppp3cb  | -          |

# Appendix

## A  Proofs of Remaining Results in Section 4

In this appendix, we present the proofs of Corollaries 4.1, 4.2, Theorem 4.2 and Proposition 4.1.

### A.1  Proof of Corollary 4.1

By Theorem 4.1, it is suffices to show $|\hat{\sigma}^2 - \sigma^2| = o_{\mathbb{P}}(1)$ and $|\hat{H}_{\alpha|\gamma} - H_{\alpha|\gamma}| = o_{\mathbb{P}}(1)$. We provide the following Lemma that shows the concentration of $\hat{\Sigma}$.

**Lemma A.1.** Under the same conditions as in Corollary 4.1, it hods that

$$||\hat{\Sigma} - \Sigma||_\infty = \mathcal{O}_{\mathbb{P}}\Big(M^3 \cdot s \cdot \sqrt{\frac{\log d}{n}}\Big).$$

*Proof.* The detailed proof is shown in Supplementary Appendix E. □

Given this lemma, we now prove Corollary 4.1. Recall that

$$\sigma^2 = \Sigma_{\alpha\alpha} - 2\mathbf{w}^{*T}\Sigma_{\gamma\alpha} + \mathbf{w}^{*T}\Sigma_{\gamma\gamma}\mathbf{w}^* \text{ and } \hat{\sigma}^2 = \hat{\Sigma}_{\alpha\alpha} - 2\hat{\mathbf{w}}^T\hat{\Sigma}_{\gamma\alpha} + \hat{\mathbf{w}}^T\hat{\Sigma}_{\gamma\gamma}\hat{\mathbf{w}}.$$

We now rearrange these terms and group them in the following way,

$$|\hat{\sigma}^2 - \sigma^2| \leq \underbrace{|\hat{\Sigma}_{\alpha\alpha} - \Sigma_{\alpha\alpha}|}_{I_1} - 2\cdot\underbrace{|\hat{\mathbf{w}}^T\hat{\Sigma}_{\gamma\alpha} - \mathbf{w}^{*T}\Sigma_{\gamma\alpha}|}_{I_2} + \underbrace{|\hat{\mathbf{w}}^T\hat{\Sigma}_{\gamma\gamma}\hat{\mathbf{w}} - \mathbf{w}^{*T}\Sigma_{\gamma\gamma}\mathbf{w}^*|}_{I_3}. \quad (A.1)$$

We first consider $I_1$. Applying Lemma A.1, we have

$$I_1 \leq ||\hat{\Sigma} - \Sigma||_\infty = \mathcal{O}_{\mathbb{P}}\Big(M^3 \cdot s \cdot \sqrt{\frac{\log d}{n}}\Big). \quad (A.2)$$

For $I_2$, using the triangle inequality, we have

$$I_2 \leq \underbrace{|(\hat{\mathbf{w}} - \mathbf{w}^*)^T(\hat{\Sigma}_{\gamma\alpha} - \Sigma_{\gamma\alpha})|}_{I_{21}} + \underbrace{|(\hat{\mathbf{w}} - \mathbf{w}^*)^T\Sigma_{\gamma\alpha})|}_{I_{22}} + \underbrace{|\mathbf{w}^{*T}(\hat{\Sigma}_{\gamma\alpha} - \Sigma_{\gamma\alpha})|}_{I_{23}}.$$

By Lemmas A.1 and 5.1, we can bound $I_{21}$, $I_{22}$ and $I_{23}$ respectively as follows,

$$I_{21} \leq ||\hat{\mathbf{w}} - \mathbf{w}^*||_1 \cdot ||\hat{\Sigma}_{\gamma\alpha} - \Sigma_{\gamma\alpha}||_\infty = \mathcal{O}_{\mathbb{P}}\Big(\lambda_s \cdot s_1\Big) \cdot \mathcal{O}_{\mathbb{P}}\Big(M^3 \cdot s \cdot \sqrt{\frac{\log d}{n}}\Big),$$

$$I_{22} \leq ||\hat{\mathbf{w}} - \mathbf{w}^*||_1 \cdot ||\Sigma_{\gamma\alpha}||_\infty = \mathcal{O}_{\mathbb{P}}\Big(\lambda_s \cdot s_1\Big),$$

$$I_{23} \leq ||\mathbf{w}^*||_1 \cdot ||\hat{\Sigma}_{\gamma\alpha} - \Sigma_{\gamma\alpha}||_\infty = \mathcal{O}_{\mathbb{P}}\Big(||\mathbf{w}^*||_1 \cdot M^3 \cdot s \cdot \sqrt{\frac{\log d}{n}}\Big).$$

It follows that

$$I_2 = \mathcal{O}_{\mathbb{P}}\Big(||\mathbf{w}^*||_1 \cdot M^3 \cdot s \cdot \sqrt{\frac{\log d}{n}} + s_1 \cdot \lambda_s\Big). \quad (A.3)$$

Following the similar arguments, we can bound $I_3$ as follows,

$$I_3 \le \underbrace{|\widehat{\mathbf{w}}^T(\widehat{\mathbf{\Sigma}}_{\gamma\gamma} - \mathbf{\Sigma}_{\gamma\gamma})\widehat{\mathbf{w}}|}_{I_{31}} + \underbrace{|\widehat{\mathbf{w}}^T \mathbf{\Sigma}_{\gamma\gamma}\widehat{\mathbf{w}} - \mathbf{w}^{*T}\mathbf{\Sigma}_{\gamma\gamma}\mathbf{w}^*|}_{I_{32}}.$$

It holds that

$$I_{31} \le \|\widehat{\mathbf{w}}\|_1^2 \cdot \|\widehat{\mathbf{\Sigma}}_{\gamma\gamma} - \mathbf{\Sigma}_{\gamma\gamma}\|_\infty = \mathcal{O}_\mathbb{P}\Big(\|\mathbf{w}^*\|_1^2 \cdot M^3 \cdot s \cdot \sqrt{\frac{\log d}{n}}\Big).$$

To control $I_{32}$, we apply the following lemma.

**Lemma A.2.** Let $\mathbf{W}$ be a symmetric $(d \times d)$-matrix and $\widehat{\mathbf{v}}$ and $\mathbf{v} \in \mathbb{R}^d$. Then

$$|\widehat{\mathbf{v}}^T \mathbf{W} \widehat{\mathbf{v}} - \mathbf{v}^T \mathbf{W} \mathbf{v}| \le \|\mathbf{W}\|_\infty \cdot \|\widehat{\mathbf{v}} - \mathbf{v}\|_1^2 + 2 \cdot \|\mathbf{W}\mathbf{v}\|_\infty \cdot \|\widehat{\mathbf{v}} - \mathbf{v}\|_1.$$

*Proof of Lemma A.2.* Note that

$$\begin{aligned}|\widehat{\mathbf{v}}^T \mathbf{W} \widehat{\mathbf{v}} - \mathbf{v}^T \mathbf{W} \mathbf{v}| &\le |(\widehat{\mathbf{v}} - \mathbf{v})^T \mathbf{W}(\widehat{\mathbf{v}} - \mathbf{v})| + 2 \cdot |\mathbf{v}^T \mathbf{W}(\widehat{\mathbf{v}} - \mathbf{v})| \\ &\le \|\mathbf{W}\|_\infty \cdot \|\widehat{\mathbf{v}} - \mathbf{v}\|_1^2 + 2 \cdot \|\mathbf{W}\mathbf{v}\|_\infty \cdot \|\widehat{\mathbf{v}} - \mathbf{v}\|_1.\end{aligned}$$

The proof is complete. $\square$

By Lemma A.2, we can show that

$$I_{32} \le \|\mathbf{\Sigma}_{\gamma\gamma}\|_\infty \cdot \|\widehat{\mathbf{w}} - \mathbf{w}^*\|_1^2 + \|\mathbf{\Sigma}_{\gamma\alpha}\|_\infty \cdot \|\widehat{\mathbf{w}} - \mathbf{w}^*\|_1 = \mathcal{O}_\mathbb{P}(\lambda_s \cdot s_1).$$

It follows that

$$I_3 = \mathcal{O}_\mathbb{P}\Big(\|\mathbf{w}^*\|_1^2 \cdot M^3 \cdot s \cdot \sqrt{\frac{\log d}{n}} + s_1 \cdot \lambda_s\Big). \tag{A.4}$$

Combining (A.1), (A.2), (A.3) and (A.4), we obtain the convergence rate

$$|\widehat{\sigma}^2 - \sigma^2| = \mathcal{O}_\mathbb{P}\Big(\|\mathbf{w}^*\|_1^2 \cdot M^3 \cdot s \cdot \sqrt{\frac{\log d}{n}} + s_1 \cdot \lambda_s\Big) = o_\mathbb{P}(1).$$

Now we prove $|\widehat{H}_{\alpha|\gamma} - H_{\alpha|\gamma}| = o_\mathbb{P}(1)$. By definition, we have

$$\begin{aligned}|\widehat{H}_{\alpha|\gamma} - H_{\alpha|\gamma}| &= |-\nabla^2_{\alpha\alpha}\ell(\widehat{\boldsymbol{\beta}}) + \widehat{\mathbf{w}}^T \nabla^2_{\alpha\gamma}\ell(\widehat{\boldsymbol{\beta}}) - H_{\alpha\alpha} + \mathbf{w}^{*T} \mathbf{H}_{\alpha\gamma}| \\ &\le |\nabla^2_{\alpha\alpha}\ell(\widehat{\boldsymbol{\beta}}) + H_{\alpha\alpha}| + \|\widehat{\mathbf{w}}\|_1 \|\nabla^2_{\alpha\gamma}\ell(\widehat{\boldsymbol{\beta}}) + \mathbf{H}_{\alpha\gamma}\|_\infty + \|\mathbf{H}_{\alpha\gamma}\|_\infty \|\widehat{\mathbf{w}} - \mathbf{w}^*\|_1.\end{aligned} \tag{A.5}$$

Applying the argument in Step 3 of the proof of Theorem 4.1, we get

$$\begin{aligned}\|\nabla^2 \ell(\widehat{\boldsymbol{\beta}}) + \mathbf{H}\|_\infty &= \|\nabla^2 \ell(\widehat{\boldsymbol{\beta}}) - \nabla^2 \ell(\boldsymbol{\beta}^*)\|_\infty + \|\nabla^2 \ell(\boldsymbol{\beta}^*) + \mathbf{H}\|_\infty \\ &= \mathcal{O}_\mathbb{P}\Big(M \cdot s \cdot \sqrt{\frac{\log d}{n}} + M^2 \cdot \sqrt{\frac{\log d}{n}}\Big).\end{aligned} \tag{A.6}$$

Therefore, $|\nabla^2_{\alpha\alpha}\ell(\widehat{\boldsymbol{\beta}}) + H_{\alpha\alpha}| + \|\widehat{\mathbf{w}}\|_1 \|\nabla^2_{\alpha\gamma}\ell(\widehat{\boldsymbol{\beta}}) + \mathbf{H}_{\alpha\gamma}\|_\infty = \mathcal{O}_\mathbb{P}\big(\|\mathbf{w}^*\|_1 M \cdot (M+s) \cdot \sqrt{\log d/n}\big) = o_\mathbb{P}(1)$. Moreover, $\|\mathbf{H}_{\alpha\gamma}\|_\infty \|\widehat{\mathbf{w}} - \mathbf{w}^*\|_1 = \mathcal{O}_\mathbb{P}(s_1 \cdot \lambda_s)$. Hence by (A.5), we conclude that $|\widehat{H}_{\alpha|\gamma} - H_{\alpha|\gamma}| = o_\mathbb{P}(1)$.

Applying the result of Theorem 4.1 and Slusky's theorem, we obtain the conclusion of the corollary.

## A.2 Proof of Corollary 4.2

In the previous section we proved that $|\widehat{\sigma}^2 - \sigma^2| = o_{\mathbb{P}}(1)$ and $|\widehat{H}_{\alpha|\boldsymbol{\gamma}} - H_{\alpha|\boldsymbol{\gamma}}| = o_{\mathbb{P}}(1)$. Therefore, by applying Theorem 4.2 and Slusky's theorem, we obtain

$$(4 \cdot \widehat{\sigma}^2)^{-1} \cdot \widehat{H}_{\alpha|\boldsymbol{\gamma}} \cdot \Lambda_n \rightsquigarrow \chi_1^2.$$

Thus, we have $\lim_{n\to\infty} \mathbb{P}(\psi_{\text{DLRT}}(\omega) = 1 \mid H_0) = \lim_{n\to\infty} \mathbb{P}\big((4 \cdot \widehat{\sigma}^2)^{-1} \cdot \widehat{H}_{\alpha|\boldsymbol{\gamma}} \cdot \Lambda_n > \chi_{1\omega}^2\big) = \omega$. Similarly, for any $t \in (0,1)$, we have

$$\lim_{n\to\infty} \mathbb{P}(P_{\text{DLRT}} < t) = \lim_{n\to\infty} \mathbb{P}\Big(\chi_1^2\big((4 \cdot \widehat{\sigma}^2)^{-1} \cdot \widehat{H}_{\alpha|\boldsymbol{\gamma}} \cdot \Lambda_n\big) > 1 - t\Big)$$
$$= \lim_{n\to\infty} \mathbb{P}\Big((4 \cdot \widehat{\sigma}^2)^{-1} \cdot \widehat{H}_{\alpha|\boldsymbol{\gamma}} \cdot \Lambda_n > \chi_{1t}^2\Big) = t.$$

This completes the proof.

## A.3 Proof of Theorem 4.2

By the first order KKT condition, we have $\widehat{\ell}'_n(\widehat{\alpha}^P) = 0$. Hence, using Taylor expansion, we have for some $\bar{\alpha}_1$ lying between $\alpha_0$ and $\widehat{\alpha}$ that

$$\widehat{\ell}_n(\alpha_0) - \widehat{\ell}_n(\widehat{\alpha}^P) = \widehat{\ell}'_n(\widehat{\alpha}^P)(\alpha_0 - \widehat{\alpha}^P) + \frac{1}{2}\widehat{\ell}''_n(\bar{\alpha}_1)(\widehat{\alpha}^P - \alpha_0)^2 = \frac{1}{2}\widehat{\ell}''_n(\bar{\alpha}_1)(\widehat{\alpha}^P - \alpha_0)^2. \quad (A.7)$$

Under the null hypothesis, $\alpha^* = \alpha_0$. Therefore,

$$\Lambda_n = -2n\big\{\widehat{\ell}_n(\alpha_0) - \widehat{\ell}_n(\widehat{\alpha}^P)\big\} = -\widehat{\ell}''_n(\bar{\alpha}_1)\big\{\sqrt{n} \cdot (\widehat{\alpha}^P - \alpha^*)\big\}^2.$$

By Theorem 4.1, we have $\sqrt{n} \cdot (\widehat{\alpha}^P - \alpha^*) \rightsquigarrow N\big(0, 4\sigma^2 \cdot H_{\alpha|\boldsymbol{\gamma}}^{-2}\big)$. Moreover, applying the exact same argument as Step 3 in the proof of Theorem 4.1, we obtain $-\widehat{\ell}''_n(\bar{\alpha}_1) = H_{\alpha|\boldsymbol{\gamma}} + o_{\mathbb{P}}(1)$. Therefore, applying Slusky and continuous mapping theorem, we have

$$(4\sigma^2)^{-1} H_{\alpha|\boldsymbol{\gamma}} \Lambda_n \rightsquigarrow \chi_1^2.$$

This completes the proof.

## A.4 Proof of Proposition 4.1

When $\boldsymbol{\beta}^* = \mathbf{0}$, without loss of generality, we assume that $\mathbb{E}(y) = 0$, $\mathbb{E}(y^2) = \sigma_y^2$, $\mathbb{E}(\boldsymbol{x}) = 0$ and $\text{Cov}(\boldsymbol{x}) = \boldsymbol{\Sigma}_x$. By the definition of $\mathbf{H}$, we have that

$$\mathbf{H} = \mathbb{E}\bigg\{\frac{(y_i - y_j)^2 \cdot (\boldsymbol{x}_i - \boldsymbol{x}_j)^{\otimes 2}}{4}\bigg\} = \sigma_y^2 \cdot \boldsymbol{\Sigma}_x.$$

Similarly, we can show that

$$\frac{1}{n} \cdot \sum_{i=1}^{n} \frac{1}{(n-1)^2} \sum_{j=1,k=1,j\neq i,k\neq i} \mathbb{E}\bigg\{\frac{(y_i - y_j) \cdot (y_i - y_k) \cdot (\boldsymbol{x}_i - \boldsymbol{x}_j)(\boldsymbol{x}_i - \boldsymbol{x}_k)^T}{4}\bigg\}$$
$$= \frac{1}{n} \cdot \sum_{i=1}^{n} \frac{1}{(n-1)^2} \cdot \frac{(n-1)^2 \cdot \sigma_y^2}{4} \cdot \boldsymbol{\Sigma}_x + o(1) = \frac{1}{4} \cdot \sigma_y^2 \cdot \boldsymbol{\Sigma}_x + o(1).$$

This implies that, $\boldsymbol{\Sigma} = \frac{1}{4} \cdot \sigma_y^2 \cdot \boldsymbol{\Sigma}_x$. Note that, by definition,

$$\sigma^2 = \frac{\sigma_y^2}{4} \cdot (\boldsymbol{\Sigma}_{x\alpha\alpha} - \boldsymbol{\Sigma}_{x\alpha\gamma}\boldsymbol{\Sigma}_{x\gamma\gamma}^{-1}\boldsymbol{\Sigma}_{x\gamma\alpha}), \text{ and } H_{\alpha|\gamma} = \boldsymbol{\Sigma}_{x\alpha\alpha} - \boldsymbol{\Sigma}_{x\alpha\gamma}\boldsymbol{\Sigma}_{x\gamma\gamma}^{-1}\boldsymbol{\Sigma}_{x\gamma\alpha}.$$

Hence, by Theorem 4.2, the asymptotic variance of $\alpha$ is $\sigma_y^2 \cdot (\boldsymbol{\Sigma}_{x\alpha\alpha} - \boldsymbol{\Sigma}_{x\alpha\gamma}\boldsymbol{\Sigma}_{x\gamma\gamma}^{-1}\boldsymbol{\Sigma}_{x\gamma\alpha})^{-1}$.

To show that $\widehat{\alpha}^P$ is semiparametrically efficient, consider the Gaussian linear regression model

$$y = \boldsymbol{x}^T\boldsymbol{\delta} + \epsilon, \quad \text{where} \quad \epsilon \sim N(0, \sigma_y^2),$$

which is a parametric sub-model of the semiparametric GLM with $\boldsymbol{\beta} = (\alpha, \boldsymbol{\gamma}) = \boldsymbol{\delta}/\sigma_y^2$. Assume that the variance $\sigma_y^2$ is known. Since $\boldsymbol{\beta}^* = \mathbf{0}$ is equivalent to $\boldsymbol{\delta}^* = \mathbf{0}$, the Fisher information for estimating $\boldsymbol{\delta}$ is $\sigma_y^{-2}\boldsymbol{\Sigma}_x$. Hence, the Fisher information for estimating $\boldsymbol{\beta}$ is $\sigma_y^2\boldsymbol{\Sigma}_x$. By the block matrix inversion, the Fisher information for estimating $\alpha$ is $\sigma_y^2 \cdot (\boldsymbol{\Sigma}_{x\alpha\alpha} - \boldsymbol{\Sigma}_{x\alpha\gamma}\boldsymbol{\Sigma}_{x\gamma\gamma}^{-1}\boldsymbol{\Sigma}_{x\gamma\alpha})^{-1}$.

Hence, the inverse of the asymptotic variance of $\widehat{\alpha}^P$ is identical to the Fisher information under the Gaussian linear regression model. This implies that it is the least favorable parametric sub-model, and thus $\widehat{\alpha}^P$ achieves the semiparametric information bound.

# B  Proofs of Lemma 5.1, 5.3, 5.5 and Lemma 5.6

In this appendix, we present the proofs of Lemma 5.1, 5.3 and Lemma 5.5, 5.6 used in the proof of Theorem 4.1.

## B.1  Proof of Lemma 5.1

Let $\widehat{\boldsymbol{\Delta}} = \widehat{\mathbf{w}} - \mathbf{w}^*$. We first invoke the following lemma to conclude that $\mathbf{w}^*$ is in the feasible set of optimization problem.

**Lemma B.1.** Under the same conditions as in Theorem 4.1, we obtain

$$||\nabla^2_{\alpha\gamma}\ell(\widehat{\boldsymbol{\beta}}) - \mathbf{w}^{*T}\nabla^2_{\gamma\gamma}\ell(\widehat{\boldsymbol{\beta}})||_\infty = \mathcal{O}_{\mathbb{P}}\Big(||\mathbf{w}^*||_1 \cdot M \cdot (s+M) \cdot \sqrt{\frac{\log d}{n}}\Big).$$

Thus, to ensure $\mathbf{w}^*$ is in the feasible set of optimization problem, we take,

$$\lambda_s \geq C\Big(||\mathbf{w}^*||_1 \cdot M \cdot (s+M) \cdot \sqrt{\frac{\log d}{n}}\Big), \text{ for some constant } C.$$

*Proof of Lemma B.1.* Since by definition $\mathbf{H}_{\alpha\gamma} = \mathbf{w}^{*T}\mathbf{H}_{\gamma\gamma}$, it holds that

$$||\nabla^2_{\alpha\gamma}\ell(\widehat{\boldsymbol{\beta}}) - \mathbf{w}^{*T}\nabla^2_{\gamma\gamma}\ell(\widehat{\boldsymbol{\beta}})||_\infty \leq \underbrace{||\mathbf{H}_{\alpha\gamma} + \nabla^2_{\alpha\gamma}\ell(\widehat{\boldsymbol{\beta}})||_\infty}_{I_1} + \underbrace{||-\mathbf{w}^{*T}\{\mathbf{H}_{\gamma\gamma} + \nabla^2_{\gamma\gamma}\ell(\widehat{\boldsymbol{\beta}})\}||_\infty}_{I_2}.$$

Now, we consider $I_1$ and $I_2$ separately. By (A.6), we have

$$I_1 = \mathcal{O}_{\mathbb{P}}\Big(M \cdot (s+M) \cdot \sqrt{\frac{\log d}{n}}\Big).$$

We complete the proof by invoking (A.6) again for $I_2$,

$$I_2 \leq ||\mathbf{H}_{\gamma\gamma} + \nabla^2_{\gamma\gamma}\ell(\widehat{\boldsymbol{\beta}})||_\infty \cdot ||\mathbf{w}^*||_1 = \mathcal{O}_{\mathbb{P}}\Big(||\mathbf{w}^*||_1 \cdot M \cdot (s + M) \cdot \sqrt{\frac{\log d}{n}}\Big).$$

$\square$

By the definition of the Dantzig selector and Lemma B.1,

$$||\nabla^2_{\gamma\gamma}\ell(\widehat{\boldsymbol{\beta}})\widehat{\boldsymbol{\Delta}}||_\infty \leq ||\nabla^2_{\gamma\alpha}\ell(\widehat{\boldsymbol{\beta}}) - \nabla^2_{\gamma\gamma}\ell(\widehat{\boldsymbol{\beta}})\widehat{\mathbf{w}}||_\infty + ||\nabla^2_{\gamma\alpha}\ell(\widehat{\boldsymbol{\beta}}) - \nabla^2_{\gamma\gamma}\ell(\widehat{\boldsymbol{\beta}})\mathbf{w}^*||_\infty \leq 2\cdot\lambda_s. \quad (B.1)$$

Note that by Hölder inequality and (B.1), we have the following bound,

$$-\widehat{\boldsymbol{\Delta}}^T \nabla^2_{\gamma\gamma}\ell(\widehat{\boldsymbol{\beta}})\widehat{\boldsymbol{\Delta}} \leq ||\widehat{\boldsymbol{\Delta}}||_1 \cdot ||\nabla^2_{\gamma\gamma}\ell(\widehat{\boldsymbol{\beta}})\widehat{\boldsymbol{\Delta}}||_\infty \leq 2\cdot\lambda_s \cdot ||\widehat{\boldsymbol{\Delta}}||_1. \quad (B.2)$$

Let $S_w = \{j : w_j^* \neq 0\}$ denote the support set of $\mathbf{w}^*$. Note that $\mathbf{w}^*_{S_w^c} = \mathbf{0}$. By the definition of the Dantzig selector,

$$\sum_{j \in S_w} |w_j^*| \geq \sum_{j \in S_w} |\widehat{w}_j| + \sum_{j \in S_w^c} |\widehat{w}_j|. \quad (B.3)$$

By the triangle inequality, we have

$$\sum_{j \in S_w} |\widehat{w}_j| \geq \sum_{j \in S_w} |w_j^*| - \sum_{j \in S_w} |\widehat{w}_j - w_j^*|. \quad (B.4)$$

Adding inequalities (B.3) and (B.4) together, we get $||\widehat{\boldsymbol{\Delta}}_{S_w^c}|| \leq ||\widehat{\boldsymbol{\Delta}}_{S_w}||_1$. This further implies that $||\widehat{\boldsymbol{\Delta}}|| \leq 2 \cdot ||\widehat{\boldsymbol{\Delta}}_{S_w}||_1$ and plugging it into (B.2), we have $-\widehat{\boldsymbol{\Delta}}^T \nabla^2_{\gamma\gamma}\ell(\widehat{\boldsymbol{\beta}})\widehat{\boldsymbol{\Delta}} \leq 4 \cdot \lambda_s \cdot ||\widehat{\boldsymbol{\Delta}}_{S_w}||_1$. Next, we need to verify that for $n$ large enough

$$\inf_{\mathbf{v} \in \mathcal{C}} \frac{-s_1 \cdot (\mathbf{v}^T \nabla^2_{\gamma\gamma}\ell(\widehat{\boldsymbol{\beta}})\mathbf{v})}{||\mathbf{v}_{S_w}||_1^2} \geq \rho', \quad \text{where } \mathcal{C} = \{\mathbf{v} \in \mathbb{R}^{d-1} : ||\mathbf{v}_{S_w^c}||_1 \leq ||\mathbf{v}_{S_w}||_1\},$$

where $\rho'$ is a positive constant and $S_w = \{j : w_j^* \neq 0\}$ is the support set for $\mathbf{w}^*$. For any $\mathbf{v} \in \mathcal{C}$, by the proof of Lemma 5.4, we obtain

$$\frac{-s_1 \cdot \mathbf{v}^T \nabla^2_{\gamma\gamma}\ell(\widehat{\boldsymbol{\beta}})\mathbf{v}}{||\mathbf{v}_{S_w}||_1^2} \geq \frac{-s_1 \cdot \mathbf{v}^T \nabla^2_{\gamma\gamma}\ell(\boldsymbol{\beta}^*)\mathbf{v}}{||\mathbf{v}_{S_w}||_1^2} \cdot \exp(-b), \quad (B.5)$$

where $b = \max_{i<j} |(y_i - y_j) \cdot (\widehat{\boldsymbol{\beta}} - \boldsymbol{\beta}^*)^T (\boldsymbol{x}_i - \boldsymbol{x}_j)|$. Note that, for $n$ large enough,

$$b \leq \max_k \max_{i<j} |(y_i - y_j) \cdot (x_{ik} - x_{jk})| \cdot ||\widehat{\boldsymbol{\beta}} - \boldsymbol{\beta}^*||_1 \leq \log 2.$$

By (B.5) and $\lambda_{\min}(\mathbf{H}) \geq c > 0$, it yields

$$\begin{aligned}
\frac{-s_1 \cdot \mathbf{v}^T \nabla^2_{\gamma\gamma}\ell(\widehat{\boldsymbol{\beta}})\mathbf{v}}{||\mathbf{v}_{S_w}||_1^2} &\geq \frac{1}{2} \cdot \frac{-s_1 \cdot \mathbf{v}^T \nabla^2_{\gamma\gamma}\ell(\boldsymbol{\beta}^*)\mathbf{v}}{||\mathbf{v}_{S_w}||_1^2} \geq \frac{1}{2} \cdot \frac{s_1 \cdot (c \cdot ||\mathbf{v}||_2^2 - ||\mathbf{H}_{\gamma\gamma} + \nabla^2_{\gamma\gamma}\ell(\boldsymbol{\beta}^*)||_\infty \cdot ||\mathbf{v}||_1^2)}{||\mathbf{v}_{S_w}||_1^2} \\
&\geq \frac{c}{2} - \mathcal{O}_{\mathbb{P}}(s_1 \cdot M^2 \cdot \sqrt{\log d/n}) = \frac{c}{2} + o_{\mathbb{P}}(1),
\end{aligned}$$

where the last line follows from the proof of Lemma 5.4. With $\rho' = c/2$, thus, the left hand side of (B.2) can be lower bounded by

$$-\widehat{\boldsymbol{\Delta}}^T \nabla^2_{\gamma\gamma} \ell(\widehat{\boldsymbol{\beta}}) \widehat{\boldsymbol{\Delta}} \geq \frac{1}{2} \cdot \rho' \cdot s_1^{-1} \cdot ||\widehat{\boldsymbol{\Delta}}_{S_w}||_1^2.$$

Combining the upper and lower bounds, we finally obtain the claimed results

$$||\widehat{\boldsymbol{\Delta}}_{S_w}||_1 \leq \frac{8}{\rho'} \cdot \lambda_s \cdot s_1, \text{ and } ||\widehat{\boldsymbol{\Delta}}||_1 \leq 2||\widehat{\boldsymbol{\Delta}}_{S_w}||_1 \leq \frac{16}{\rho'} \cdot \lambda_s \cdot s_1.$$

## B.2 Proof of Lemma 5.3

By the symmetry of the kernel function, $U_n$ can be rewritten as follows,

$$U_n = \frac{1}{k} \cdot \frac{1}{n!} \cdot \sum v_n(X_{i_1}, ..., X_{i_n}),$$

where the summation is over all $n!$ permutations of $\{1, ..., n\}$, and

$$v_n(X_{i_1}, ..., X_{i_n}) = u(X_{i_1}, ..., X_{i_m}) + u(X_{i_{m+1}}, ..., X_{i_{2m}})... + u(X_{i_{km-m+1}}, ..., X_{i_{km}}).$$

Note that $v_n(X_1, ..., X_n)$ is a sum of $k$ independent random variables. Then, for any $x \geq 0$ and $t > 0$, by the Markov inequality, we obtain that

$$\begin{aligned}
\mathbb{P}(U_n \geq x) &= \mathbb{P}\Big[\exp\Big\{t \cdot \frac{1}{n!} \cdot \sum v_n(X_{i_1}, ..., X_{i_n})\Big\} \geq \exp(t \cdot k \cdot x)\Big] \\
&\leq \exp(-t \cdot k \cdot x) \cdot \mathbb{E}\Big[\exp\Big\{\frac{1}{n!} \cdot \sum t \cdot v_n(X_{i_1}, ..., X_{i_n})\Big\}\Big], \quad \text{(B.6)}
\end{aligned}$$

where the summation is over all $n!$ permutations of $\{1, ..., n\}$. By Jensen inequality, (B.6) yields,

$$\begin{aligned}
\mathbb{P}(U_n \geq x) &\leq \exp(-t \cdot k \cdot x) \cdot \mathbb{E}\Big[\frac{1}{n!} \cdot \sum \exp\{t \cdot v_n(X_{i_1}, ..., X_{i_n})\}\Big] \\
&= \exp(-t \cdot k \cdot x) \cdot \frac{1}{n!} \cdot \sum \prod_{s=1}^{k} \mathbb{E}\Big[\exp\{t \cdot u(X_{i_{sm-m+1}}, ..., X_{i_{sm}})\}\Big],
\end{aligned}$$

where the last equality follows by the independence of $u(X_{i_{sm-m+1}}, ..., X_{i_{sm}})$ for $s = 1, ..., k$. For notational simplicity, we write $u$ for $u(X_{i_{sm-m+1}}, ..., X_{i_{sm}})$. By (5.3), for all $j \geq 1$ and $L_1 > 1$,

$$\mathbb{E}|u|^j = \int_0^\infty \mathbb{P}(|u|^j > x) \cdot dx \leq L_1 \cdot \int_0^\infty \exp(-L_2 \cdot x^{1/j}) \cdot dx = \frac{L_1}{L_2^j} \cdot j! \leq \Big(\frac{L_1}{L_2}\Big)^j \cdot j!. \quad \text{(B.7)}$$

Next, we apply the Taylor theorem for $\exp(t \cdot u)$. By $\mathbb{E}u = 0$, it yields,

$$\mathbb{E}\{\exp(t \cdot u)\} = 1 + t \cdot \mathbb{E}u + \sum_{j > 1} \frac{t^j \cdot \mathbb{E}u^j}{j!} = 1 + \sum_{j > 1} \frac{t^j \cdot \mathbb{E}u^j}{j!}.$$

Together with (B.7), it follows that for $t \leq L_2/(2 \cdot L_1)$,

$$\begin{aligned}
\mathbb{E}\{\exp(t \cdot u)\} &\leq 1 + \sum_{j > 1} \Big(\frac{t \cdot L_1}{L_2}\Big)^j = 1 + \Big(\frac{t \cdot L_1}{L_2}\Big)^2 \cdot \sum_{j \geq 0} \Big(\frac{t \cdot L_1}{L_2}\Big)^j \\
&\leq 1 + 2 \cdot \Big(\frac{t \cdot L_1}{L_2}\Big)^2 \leq \exp\Big\{2 \cdot \Big(\frac{t \cdot L_1}{L_2}\Big)^2\Big\}. \quad \text{(B.8)}
\end{aligned}$$

Combining (B.6) and (B.8), we conclude that if $t \leq L_2/(2 \cdot L_1)$

$$\mathbb{P}(U_n \geq x) \leq \exp(-t \cdot k \cdot x) \cdot \exp\left(\frac{2 \cdot L_1^2 \cdot t^2}{L_2^2} \cdot k\right).$$

Then, we can optimize the upper bound with respect to $t$ for any given $x$. This yields $t = \min\{L_2^2 \cdot x/(4 \cdot L_1^2), L_2/(2 \cdot L_1)\}$. Then

$$\mathbb{P}(U_n \geq x) \leq \exp\left(-\min\left\{\frac{L_2^2 \cdot x^2}{8 \cdot L_1^2}, \frac{L_2 \cdot x}{4 \cdot L_1}\right\} \cdot k\right).$$

Applying the previous argument to $-U_n$, we obtain

$$\mathbb{P}(U_n \leq -x) \leq \exp\left(-\min\left\{\frac{L_2^2 \cdot x^2}{8 \cdot L_1^2}, \frac{L_2 \cdot x}{4 \cdot L_1}\right\} \cdot k\right).$$

The result follows by a combination of these two bounds.

### B.3  Proof of Lemma 5.5

*Proof of Lemma 5.5.* By the definition of $\mathbf{g}(y_i, \boldsymbol{x}_i, \boldsymbol{\beta}^*)$, we have

$$\widehat{\mathbf{U}}_n = \frac{2}{n(n-1)} \cdot \sum_{i=1}^{n} \sum_{j \neq i} \mathbf{h}_{ij|i}, \quad \text{and} \quad \mathbf{h}_{ij|i} = \mathbb{E}(\mathbf{h}_{ij} \mid y_i, \boldsymbol{x}_i), \tag{B.9}$$

where $\mathbf{h}_{ij} = \mathbf{h}_{ij}(\boldsymbol{\beta}^*)$ is given by (5.4) and $\mathbf{g}(y_i, \boldsymbol{x}_i, \boldsymbol{\beta}^*) = \frac{1}{n-1} \cdot \sum_{j \neq i} \mathbf{h}_{ij|i}$. Note that

$$\frac{\sqrt{n}}{2} \cdot (\mathbf{b}^T \boldsymbol{\Sigma} \mathbf{b})^{-1/2} \cdot \mathbf{b}^T \nabla \ell(\boldsymbol{\beta}^*)$$
$$= \underbrace{(\mathbf{b}^T \boldsymbol{\Sigma} \mathbf{b})^{-1/2} \cdot \frac{1}{\sqrt{n}} \cdot \sum_{i=1}^{n} \mathbf{b}^T \mathbf{g}(y_i, \boldsymbol{x}_i, \boldsymbol{\beta}^*)}_{I_1} + \underbrace{\frac{\sqrt{n}}{2} \cdot (\mathbf{b}^T \boldsymbol{\Sigma} \mathbf{b})^{-1/2} \cdot \mathbf{b}^T \{\nabla \ell(\boldsymbol{\beta}^*) - \widehat{\mathbf{U}}_n\}}_{I_2}.$$

Note that $\mathbb{E}\{\mathbf{g}(y_i, \boldsymbol{x}_i, \boldsymbol{\beta}^*)\} = \mathbb{E}(\mathbf{h}_{ij}) = \mathbf{0}$. Hence, $\mathbf{b}^T \mathbf{g}(y_i, \boldsymbol{x}_i, \boldsymbol{\beta}^*)$ in $I_1$ are mutually independent mean-zero random variables. In addition, we have $\text{Cov}(I_1) = 1$. To apply the central limit theorem for $I_1$, we now verify the Lyapunov condition for $I_1$. By Assumption 4.2, $\mathbf{b}^T \boldsymbol{\Sigma} \mathbf{b}$ is lower bounded by $\lambda_{\min}(\boldsymbol{\Sigma})$, and

$$n^{-3/2} \cdot \sum_{i=1}^{n} \mathbb{E}|(\mathbf{b}^T \boldsymbol{\Sigma} \mathbf{b})^{-1/2} \cdot \mathbf{b}^T \mathbf{g}(y_i, \boldsymbol{x}_i, \boldsymbol{\beta}^*)|^3 = O(1) \cdot n^{-3/2} \cdot \sum_{i=1}^{n} \mathbb{E}|\mathbf{b}^T \mathbf{g}(y_i, \boldsymbol{x}_i, \boldsymbol{\beta}^*)|^3.$$

Let $\mathcal{B}$ denote the support set of the vector $\mathbf{b}$. Note that $\|\mathbf{b}_{\mathcal{B}}\|_2 \leq \|\mathbf{b}\|_2 = 1$ and $\mathbf{b}^T \mathbf{g}(y_i, \boldsymbol{x}_i, \boldsymbol{\beta}^*) = \mathbf{b}_{\mathcal{B}}^T \mathbf{g}_{\mathcal{B}}(y_i, \boldsymbol{x}_i, \boldsymbol{\beta}^*)$. By the Cauchy inequality, we have

$$n^{-3/2} \cdot \sum_{i=1}^{n} \mathbb{E}|\mathbf{b}^T \mathbf{g}(y_i, \boldsymbol{x}_i, \boldsymbol{\beta}^*)|^3 \leq n^{-3/2} \cdot \sum_{i=1}^{n} \mathbb{E}\|\mathbf{g}_{\mathcal{B}}(y_i, \boldsymbol{x}_i, \boldsymbol{\beta}^*)\|_2^3.$$

By (5.4), it is easy to show that $||\mathbf{h}_{ij}||_\infty \leq M$, which implies $||\mathbf{h}_{ij|i}||_\infty \leq M$, and $||\mathbf{g}(y_i, \boldsymbol{x}_i, \boldsymbol{\beta}^*)||_\infty \leq M$. Moreover, by assumption $|\mathcal{B}| \leq \widetilde{s}$ and the Hölder inequality, we obtain

$$n^{-3/2} \cdot \sum_{i=1}^n \mathbb{E}|(\mathbf{b}^T \boldsymbol{\Sigma} \mathbf{b})^{-1/2} \cdot \mathbf{b}^T \mathbf{g}(y_i, \boldsymbol{x}_i, \boldsymbol{\beta}^*)|^3 = \mathcal{O}_\mathbb{P}(\widetilde{s}^{3/2} \cdot n^{-1/2} \cdot M^3) = o_\mathbb{P}(1).$$

Thus, the Lyapunov Central Limit Theorem implies $I_1 \rightsquigarrow N(0,1)$. In the following, we shall show that $I_2 = o_\mathbb{P}(1)$. Note that $I_2$ can be rewritten as

$$I_2 = \frac{\sqrt{n}}{2} \cdot (\mathbf{b}^T \boldsymbol{\Sigma} \mathbf{b})^{-1/2} \cdot \frac{1}{n(n-1)} \cdot \sum_{i<j} \mathbf{b}^T \mathbf{w}_{ij},$$

where $\mathbf{w}_{ij} = \mathbf{h}_{ij} - \mathbf{h}_{ij|i} - \mathbf{h}_{ij|j}$. Next, we would like to calculate the variance of $I_2$. This requires to calculate the covariance of $\mathbf{w}_{ij}$ and $\mathbf{w}_{lk}$. To this end, we have to separately consider several situations according to the equality among $i, j, l, k$. In the first case, for $i \neq l, k$ and $j \neq l, k$,

$$\begin{aligned}
\mathbb{E}(\mathbf{w}_{ij}\mathbf{w}_{lk}^T) &= \mathbb{E}(\mathbf{h}_{ij}\mathbf{h}_{lk}^T) - \mathbb{E}(\mathbf{h}_{ij}\mathbf{h}_{lk|l}^T) - \mathbb{E}(\mathbf{h}_{ij}\mathbf{h}_{lk|k}^T) - \mathbb{E}(\mathbf{h}_{ij|i}\mathbf{h}_{lk}^T) \\
&+ \mathbb{E}(\mathbf{h}_{ij|i}\mathbf{h}_{lk|l}^T) + \mathbb{E}(\mathbf{h}_{ij|j}\mathbf{h}_{lk|k}^T) - \mathbb{E}(\mathbf{h}_{ij|j}\mathbf{h}_{lk}^T) + \mathbb{E}(\mathbf{h}_{ij|j}\mathbf{h}_{lk|l}^T) + \mathbb{E}(\mathbf{h}_{ij|j}\mathbf{h}_{lk|k}^T). \quad (\text{B.10})
\end{aligned}$$

For the first term, $\mathbb{E}(\mathbf{h}_{ij}\mathbf{h}_{lk}^T) = \mathbb{E}(\mathbf{h}_{ij})\mathbb{E}(\mathbf{h}_{lk}^T) = \mathbf{0}$, followed by the independence of $\mathbf{h}_{ij}$ and $\mathbf{h}_{lk}$. Similarly, using the independence and the mean 0 results $\mathbb{E}(\mathbf{h}_{ij}) = \mathbb{E}(\mathbf{h}_{lk}) = \mathbb{E}(\mathbf{h}_{lk|k}) = \mathbb{E}(\mathbf{h}_{ij|j}) = \mathbf{0}$, all these nine terms in (B.10) are $\mathbf{0}$. This implies $\mathbb{E}(\mathbf{w}_{ij}\mathbf{w}_{lk}^T) = \mathbf{0}$. Similar to (B.10), if only one of $i, j$ is identical to one of $l, k$, say $i = l$, then

$$\mathbb{E}(\mathbf{w}_{ij}\mathbf{w}_{ik}^T) = \mathbb{E}(\mathbf{h}_{ij}\mathbf{h}_{ik}^T) - \mathbb{E}(\mathbf{h}_{ij}\mathbf{h}_{ik|i}^T) - \mathbb{E}(\mathbf{h}_{ij|i}\mathbf{h}_{ik}^T) + \mathbb{E}(\mathbf{h}_{ij|i}\mathbf{h}_{ik|i}^T),$$

where the remaining terms in (B.10) are $\mathbf{0}$ by the same arguments. Note that

$$\begin{aligned}
\mathbb{E}(\mathbf{h}_{ij}\mathbf{h}_{ik}^T) &= \mathbb{E}\{\mathbb{E}(\mathbf{h}_{ij}\mathbf{h}_{ik}^T \mid y_i, \boldsymbol{x}_i)\} = \mathbb{E}\{\mathbb{E}(\mathbf{h}_{ij} \mid y_i, \boldsymbol{x}_i)\mathbb{E}(\mathbf{h}_{ik}^T \mid y_i, \boldsymbol{x}_i)\} = \mathbb{E}(\mathbf{h}_{ij|i}\mathbf{h}_{ik|i}^T), \\
\mathbb{E}(\mathbf{h}_{ij}\mathbf{h}_{ik|i}^T) &= \mathbb{E}\{\mathbb{E}(\mathbf{h}_{ij}\mathbf{h}_{ik|i}^T \mid y_i, \boldsymbol{x}_i)\} = \mathbb{E}\{\mathbb{E}(\mathbf{h}_{ij} \mid y_i, \boldsymbol{x}_i)\mathbf{h}_{ik|i}^T\} = \mathbb{E}(\mathbf{h}_{ij|i}\mathbf{h}_{ik|i}^T), \\
\mathbb{E}(\mathbf{h}_{ij|i}\mathbf{h}_{ik}^T) &= \mathbb{E}\{\mathbb{E}(\mathbf{h}_{ik}\mathbf{h}_{ij|i}^T \mid y_i, \boldsymbol{x}_i)\} = \mathbb{E}\{\mathbb{E}(\mathbf{h}_{ik} \mid y_i, \boldsymbol{x}_i)\mathbf{h}_{ij|i}^T\} = \mathbb{E}(\mathbf{h}_{ij|i}\mathbf{h}_{ik|i}^T).
\end{aligned}$$

Therefore, $\mathbb{E}(\mathbf{w}_{ij}\mathbf{w}_{ik}^T) = \mathbf{0}$. Then, the nontriavial covariance of $\mathbf{w}_{ij}$ and $\mathbf{w}_{lk}$ must have $i = l$ and $j = k$ if $i < j, l < k$. This leads to

$$\begin{aligned}
\mathbb{E}(I_2^2) &= n \cdot (\mathbf{b}^T \boldsymbol{\Sigma} \mathbf{b})^{-1} \cdot \frac{1}{n^2(n-1)^2} \cdot \sum_{i<j}\sum_{l<k} \mathbf{b}^T \mathbb{E}(\mathbf{w}_{ij}\mathbf{w}_{lk}^T)\mathbf{b} \\
&= n \cdot (\mathbf{b}^T \boldsymbol{\Sigma} \mathbf{b})^{-1} \cdot \frac{1}{n^2(n-1)^2} \cdot \sum_{i<j} \mathbf{b}^T \mathbb{E}(\mathbf{w}_{ij}\mathbf{w}_{ij}^T)\mathbf{b}.
\end{aligned}$$

Let $\mathcal{B}$ denote the support set of the vector $\mathbf{b}$. Note that $|\mathcal{B}| \leq \widetilde{s}$ and $(\mathbf{b}^T \boldsymbol{\Sigma} \mathbf{b})^{-1} = O(1)$. Then

$$\mathbb{E}(I_2^2) = O(1) \cdot \frac{1}{n(n-1)^2} \cdot \sum_{i<j} \lambda_{\max}\left\{\mathbb{E}(\mathbf{w}_{\mathcal{B}ij}\mathbf{w}_{\mathcal{B}ij}^T)\right\}. \tag{B.11}$$

Since $||\mathbf{h}_{ij}||_\infty \leq M$, $||\mathbf{h}_{ij|i}||_\infty \leq M$, we obtain $||\mathbf{w}_{ij}\mathbf{w}_{ij}^T||_\infty = \mathcal{O}_\mathbb{P}(M^2)$, and therefore,

$$\lambda_{\max}\left\{\mathbb{E}(\mathbf{w}_{\mathcal{B}ij}\mathbf{w}_{\mathcal{B}ij}^T)\right\} \leq \widetilde{s} \cdot ||\mathbf{w}_{ij}\mathbf{w}_{ij}^T||_\infty = \mathcal{O}_\mathbb{P}(\widetilde{s} \cdot M^2). \tag{B.12}$$

Combining (B.11) and (B.12), we obtain $\mathbb{E}(I_2^2) = \mathcal{O}_\mathbb{P}(\widetilde{s} \cdot M^2/n) = o_\mathbb{P}(1)$. By the Markov inequality, we have $I_2 = o_\mathbb{P}(1)$. Applying the Slutsky's Theorem, we finish the proof. $\square$

## B.4 Proof of Lemma 5.6

*Proof of Lemma 5.6.* Define the event $\mathcal{E}_\alpha = \{|\widehat{\alpha}^P - \alpha^*| \leq C \cdot \|\mathbf{w}^*\|_1 \cdot M \cdot \sqrt{\log n/n}\}$ for some constant $C$. In the following we show that

$$\lim_{n \to \infty} \mathbb{P}(\mathcal{E}_\alpha^c) = 0,$$

for some constant $C$. Define the set $\mathcal{D} := \{\alpha : |\alpha - \alpha^*| \leq C \cdot \|\mathbf{w}^*\|_1 \cdot M \cdot \sqrt{\log n/n}\}$ and let $G(\alpha) := -\widehat{\ell}''(\alpha^*)^{-1}\widehat{\ell}'(\alpha) + \alpha$. If there exists $\alpha \in \mathcal{D}$ such that $G(\alpha) = \alpha$, then we have $\widehat{\ell}'(\alpha) = 0$. This implies $\alpha = \widehat{\alpha}$, which further implies that $\widehat{\alpha} \in \mathcal{D}$. By Brouwer Fixed Point theorem, there exists $\alpha \in \mathcal{D}$ such that $G(\alpha) = \alpha$ if $G(\mathcal{D}) \subset \mathcal{D}$. Hence, by the above chain of arguments, to show $\lim_{n \to \infty} \mathbb{P}(\mathcal{E}_\alpha^c) = 0$, it suffices to show that

$$\lim_{n \to \infty} \mathbb{P}\big(G(\mathcal{D}) \subset \mathcal{D}\big) = 1.$$

For any $\alpha \in \mathcal{D}$, by Taylor expansion, it holds for some intermediate value $\bar{\alpha}$ that

$$\begin{aligned} G(\alpha) &= -\widehat{\ell}''(\alpha^*)^{-1}\big(\widehat{\ell}'(\alpha^*) + \widehat{\ell}''(\bar{\alpha})(\alpha - \alpha^*)\big) + \alpha \\ &= -\widehat{\ell}''(\alpha^*)^{-1}\big\{\widehat{\ell}'(\alpha^*) + \big(\widehat{\ell}''(\bar{\alpha}) - \widehat{\ell}''(\alpha^*)\big)(\alpha - \alpha^*)\big\} + \alpha^* \end{aligned} \tag{B.13}$$

We first show that $-\widehat{\ell}''(\alpha^*)^{-1}$ is upper bounded with high probability. Note that

$$-\widehat{\ell}_n''(\alpha^*) = \underbrace{-\widehat{\ell}_n''(\alpha^*) + (1,\widehat{\mathbf{w}}^T)\nabla^2\ell(\boldsymbol{\beta}^*)(1,\widehat{\mathbf{w}}^T)^T}_{I_1} \underbrace{-(1,\widehat{\mathbf{w}}^T)\nabla^2\ell(\boldsymbol{\beta}^*)(1,\widehat{\mathbf{w}}^T)^T - H_{\alpha|\gamma}}_{I_2} + H_{\alpha|\gamma}. \tag{B.14}$$

The analysis of (B.14) is similar to that of (5.14) with $\bar{\alpha}$ replaced by $\alpha^*$. For $I_1$, we have

$$I_1 \leq \|\nabla^2\ell(\alpha^*, \widehat{\boldsymbol{\gamma}}(\alpha^*)) - \nabla^2\ell(\boldsymbol{\beta}^*)\|_\infty (\|\widehat{\mathbf{w}}\|_1 + 1)^2.$$

We have $\|(\alpha^*, \widehat{\boldsymbol{\gamma}}(\alpha^*)^T)^T - \boldsymbol{\beta}^*\|_1 = \|\widehat{\boldsymbol{\gamma}} - \boldsymbol{\gamma}\|_1 + |\widehat{\alpha} - \alpha^*|\|\mathbf{w}^*\|_1 = \mathcal{O}_\mathbb{P}\big(\max\{s, s_1\} \cdot \sqrt{\log d/n}\big)$. By Lemma 5.4, we obtain $\|\nabla^2\ell(\alpha^*, \widehat{\boldsymbol{\gamma}}(\alpha^*)) - \nabla^2\ell(\boldsymbol{\beta}^*)\|_\infty = \mathcal{O}_\mathbb{P}\big(M \cdot \max\{s, s_1\} \cdot \sqrt{\log d/n}\big)$. Therefore

$$I_1 = \mathcal{O}_\mathbb{P}\big(\|\mathbf{w}^*\|_1^2 \cdot M \cdot \max\{s, s_1\}\sqrt{\log d/n}\big) = o_\mathbb{P}(1).$$

Moreover, $I_2$ is the same as $I_4$ in (5.14), hence $I_2 = o_\mathbb{P}(1)$. Therefore, we have $|\widehat{\ell}_n''(\alpha^*) + H_{\alpha|\gamma}| = o_\mathbb{P}(1)$. As $H_{\alpha|\gamma}^{-1} = \mathcal{O}(1)$, we conclude that

$$-\widehat{\ell}_n''(\alpha^*)^{-1} = \mathcal{O}(1) \text{ with probability tending to 1.} \tag{B.15}$$

Next we obtain an upper bound for $\widehat{\ell}_n'(\alpha^*)$. We showed in the proof of Theorem 4.1 that $\sqrt{n}\widehat{\ell}_n'(\alpha^*) = \sqrt{n}S(\boldsymbol{\beta}^*) + o_\mathbb{P}(1)$. Moreover, by the proof of Lemma 5.5, we have

$$\sqrt{n}S(\boldsymbol{\beta}^*) = \sqrt{n}(1, -\mathbf{w}^{*T})\widehat{\boldsymbol{U}}_n + \sqrt{n}\{S(\boldsymbol{\beta}^*) - (1, -\mathbf{w}^{*T})\widehat{\boldsymbol{U}}_n\} =: I_3 + I_4. \tag{B.16}$$

By definition, we have $\widehat{\boldsymbol{U}}_n = 2n^{-1}\sum_{i=1}^n g(y_i, \boldsymbol{x}_i, \boldsymbol{\beta}^*)$, with $\|\mathbf{g}(y_i, \boldsymbol{x}_i, \boldsymbol{\beta}^*)\|_\infty \leq M$. Moreover, $\mathbf{g}(y_i, \boldsymbol{x}_i, \boldsymbol{\beta}^*)$ are i.i.d. Hence

$$\begin{aligned} \mathbb{E}[(\sqrt{n}(1, -\mathbf{w}^{*T})\widehat{\boldsymbol{U}}_n)^2] &= 4n^{-1}\mathbb{E}\bigg[\Big\{\sum_{i=1}^n (1, -\mathbf{w}^{*T})\mathbf{g}(y_i, \boldsymbol{x}_i, \boldsymbol{\beta}^*)\Big\}^2\bigg] \\ &= \mathbb{E}\big[\{(1, -\mathbf{w}^{*T})\mathbf{g}(y_i, \boldsymbol{x}_i, \boldsymbol{\beta}^*)\}^2\big] \leq (\|\mathbf{w}^*\|_1 + 1)^2 M^2 \end{aligned}$$

Applying Markov's inequality, we have

$$\mathbb{P}\Big(|I_3| \geq (\|\mathbf{w}^*\|_1 + 1) \cdot M \cdot \sqrt{\log n}\Big) \leq \frac{\mathbb{E}[(\sqrt{n}(1, -\mathbf{w}^{*T})\widehat{\boldsymbol{U}}_n)^2]}{(\|\mathbf{w}^*\|_1 + 1)^2 M^2 \log n} = \frac{1}{\log n} = o(1).$$

This implies that $I_3 \lesssim \|\mathbf{w}^*\|_1 \cdot M \cdot \sqrt{\log n}$ with probability tending to 1. Moreover, by the proof of Lemma 5.5, we have $I_4 = o_{\mathbb{P}}(1)$. Therefore, by (B.16), we have

$$|\widehat{\ell}'_n(\alpha^*)| \lesssim \|\mathbf{w}^*\|_1 \cdot M \cdot \sqrt{\log n/n} \text{ with probability tending to 1.} \tag{B.17}$$

Lastly, we bound the term $(\alpha - \alpha^*)(\widehat{\ell}''_n(\bar{\alpha}) - \widehat{\ell}''(\alpha^*))$. By the formula for $\widehat{\ell}''_n(\alpha)$, we have

$$\widehat{\ell}''_n(\bar{\alpha}) - \widehat{\ell}''(\alpha^*) = (1, -\widehat{\mathbf{w}}^T)\{\nabla^2 \ell(\bar{\alpha}, \widehat{\boldsymbol{\gamma}}(\bar{\alpha})) - \nabla^2 \ell(\alpha^*, \widehat{\boldsymbol{\gamma}}(\alpha^*))\}(1, -\widehat{\mathbf{w}}^T)^T. \tag{B.18}$$

As $\bar{\alpha} \in \mathcal{D}$, we have $\|(\bar{\alpha}, \widehat{\boldsymbol{\gamma}}(\bar{\alpha})^T) - (\alpha^*, \widehat{\boldsymbol{\gamma}}(\alpha^*)^T)\|_1 \leq |\bar{\alpha} - \alpha^*|(1 + \|\widehat{\mathbf{w}}\|_1) \lesssim \|\mathbf{w}^*\|_1^2 M \sqrt{\log n/n}$. Therefore, by Lemma 5.4,

$$\sup_{\alpha \in \mathcal{D}} \|\nabla^2 \ell(\bar{\alpha}, \widehat{\boldsymbol{\gamma}}(\bar{\alpha})) - \nabla^2 \ell(\alpha^*, \widehat{\boldsymbol{\gamma}}(\alpha^*))\|_\infty \lesssim \|\mathbf{w}^*\|_1^2 \cdot M^2 \cdot \sqrt{\frac{\log n}{n}}. \tag{B.19}$$

Therefore, by (B.18) and (B.19), we conclude that $\sup_{\alpha \in \mathcal{D}} |\widehat{\ell}''_n(\bar{\alpha}) - \widehat{\ell}''(\alpha^*)| \lesssim \|\mathbf{w}^*\|_1^4 \cdot M^2 \cdot \sqrt{\log n/n}$. which implies

$$\sup_{\alpha \in \mathcal{D}} |(\alpha - \alpha^*)(\widehat{\ell}''_n(\bar{\alpha}) - \widehat{\ell}''(\alpha^*))| \lesssim \|\mathbf{w}^*\|_1^5 \cdot M^3 \cdot \log n/n \lesssim \|\mathbf{w}^*\|_1 \cdot M \cdot \sqrt{\log n/n}, \tag{B.20}$$

where we used the scaling condition that $\|\mathbf{w}^*\|_1^4 \cdot M^2 \cdot \sqrt{\log n/n} = o(1)$.

Combining (B.13), (B.15), (B.17) and (B.20), we conclude that

$$\lim_{n \to \infty} \mathbb{P}\big(G(\mathcal{D}) \subset \mathcal{D}\big) = \lim_{n \to \infty} \mathbb{P}\Big(\sup_{\alpha \in \mathcal{D}} |G(\alpha) - \alpha^*| \lesssim \|\mathbf{w}^*\|_1 \cdot M \cdot \sqrt{\log n/n}\Big) = 1,$$

which concludes the proof. □

## C  Results for Parameter Estimation

Let $\boldsymbol{\beta}^* = (\beta_1^*, ..., \beta_d^*)^T$ denote the vector of true parameter. In this section, we establish the statistical consistency of $\widehat{\boldsymbol{\beta}}$, which is the solution to the optimization problem (4.1). To this end, one has to impose certain convexity conditions on the loss function, such that together with the nonconvex penalty, the optimization problem (4.1) remains strongly convex in a restricted set. This is guaranteed by the following assumption.

**Assumption C.1.** There exist $\rho$, $\tau$ and $r > 0$ such that for any $\mathbf{v} \in \mathbb{R}^d$ and $\|\mathbf{v}\|_1 \leq r$, it satisfies

$$-\mathbf{v}^T \nabla^2 \ell(\boldsymbol{\beta}^*) \mathbf{v} \geq \rho \cdot \|\mathbf{v}\|_2^2 - \tau \cdot \|\mathbf{v}\|_1^2 \cdot \frac{\log d}{n}. \tag{C.1}$$

Due to the convexity of the loss function $-\ell(\boldsymbol{\beta}^*)$, we have $-\mathbf{v}^T\nabla^2\ell(\boldsymbol{\beta}^*)\mathbf{v} \geq 0$, and therefore (C.1) holds trivially for $\mathbf{v} \in \mathcal{V}$, where $\mathcal{V} = \{\mathbf{v} \in \mathbb{R}^d : \frac{\|\mathbf{v}\|_2}{\|\mathbf{v}\|_1} \leq \sqrt{\frac{\tau}{\rho}} \cdot \sqrt{\frac{\log d}{n}}\}$. Note that this assumption is sufficient to imply the validity of the RSC condition in equation (4a) of Loh and Wainwright (2013). Since the loss function $-\ell(\boldsymbol{\beta})$ is convex, by Lemma 9 of Loh and Wainwright (2013), for $n$ large enough, the entire RSC condition in equations (4a) and (4b) holds. Thus, the consistency of $\widehat{\boldsymbol{\beta}}$ follows directly from Theorem 1 of Loh and Wainwright (2013), for $n$ large enough,

$$||\widehat{\boldsymbol{\beta}} - \boldsymbol{\beta}^*||_2 \leq C_2 \cdot s^{1/2} \cdot \lambda, \text{ and } ||\widehat{\boldsymbol{\beta}} - \boldsymbol{\beta}^*||_1 \leq C_1 \cdot s \cdot \lambda, \tag{C.2}$$

where $C_1$ and $C_2$ are two constants whose explicit forms are given in Loh and Wainwright (2013). In addition, Lemma 5.2 implies $\|\nabla \ell(\boldsymbol{\beta}^*)\|_\infty \leq C'' \cdot \sqrt{\log d/n}$ with high probability. Thus, with $\lambda \asymp \sqrt{\log d/n}$, we obtain

$$||\widehat{\boldsymbol{\beta}} - \boldsymbol{\beta}^*||_2 = \mathcal{O}_\mathbb{P}(s^{1/2} \cdot \sqrt{\log d/n}), \text{ and } ||\widehat{\boldsymbol{\beta}} - \boldsymbol{\beta}^*||_1 = \mathcal{O}_\mathbb{P}(s \cdot \sqrt{\log d/n}). \tag{C.3}$$

Note that this result justifies our assumption 4.4, and therefore the nonconvex estimator $\widehat{\boldsymbol{\beta}}$ can be used as an initial estimator for post-regularization inference. Since the optimization problem (4.1) is nonconvex, the obtained solution may depend on the specific algorithm for solving (4.1). Wang et al. (2013b) proposed an approximate path following algorithm, and Theorem 4.7 of Wang et al. (2013b) showed that the estimator produced by the algorithm has the same convergence rate as in (C.2) and (C.3).

Note that the rate in (C.3) is identical to that for GLMs with the nonconvex penalty (Loh and Wainwright, 2013) and the $\ell_1$ penalty (Bickel et al., 2009; van de Geer and Bühlmann, 2009). Indeed, Raskutti et al. (2011) established the minimax lower bound in the sparse linear regression with Gaussian noise, that is $\min_{\widetilde{\boldsymbol{\beta}}} \max_{\boldsymbol{\beta}^* \in \mathbb{B}_0(s)} ||\widetilde{\boldsymbol{\beta}} - \boldsymbol{\beta}^*||_2 \gtrsim \sqrt{s \cdot \log(d/s)/n}$ with positive probability. Since the linear regression with Gaussian noise is a parametric submodel of the semiparametric model, this also serves as a lower bound for it. We conclude that the rate matches the lower bounds up to a logarithmic factor. Therefore, $\widehat{\boldsymbol{\beta}}$ in (4.1) is nearly rate optimal in the minimax sense under the semiparametric GLM.

Finally, we conclude this section by showing that assumption C.1 holds for many important GLMs, such as linear regression with Gaussian noise, logistic regression and Poisson regression, with high probability. For completeness, we also verify the restricted eigenvalue condition (RE) in the following assumption C.2, which is the key assumption for analyzing the optimization problem (4.1) with the $\ell_1$ penalty.

**Assumption C.2.** There exists $\rho' > 0$ such that for any $\mathbf{v} \in \mathbb{R}^d$ and $\|\mathbf{v}_{S^c}\|_1 \leq 3 \cdot \|\mathbf{v}_S\|_1$, it satisfies

$$-\mathbf{v}^T\nabla^2\ell(\boldsymbol{\beta}^*)\mathbf{v} \geq \rho' \cdot \|\mathbf{v}\|_2^2, \tag{C.4}$$

where $S = \{j : \beta_j^* \neq 0\}$ denotes the support set of $\boldsymbol{\beta}^*$ and $s = |S|$ is the cardinality of $S$.

Given this assumption, the detailed derivation for the rate of convergence of the estimator $\widehat{\boldsymbol{\beta}}$ can be found in an earlier version of the current paper; See Ning and Liu (2014). We now prove the validity of these assumptions in the following proposition.

**Proposition C.1.** Let the mean and covariance of $\boldsymbol{x}_i$ be 0 and $\boldsymbol{\Sigma}_x = \text{Cov}(\boldsymbol{x}_i)$ and denote $m = \max_{1 \leq i \leq n} \max_{1 \leq j \leq d} |x_{ij}|$. Assume that $\boldsymbol{x}_i$ is a sub-Gaussian vector with a finite sub-Gaussian norm denoted by $C_x$, and also assume $\|\boldsymbol{\beta}^*\|_2 \leq C_\beta$ for some finite constant $C_\beta$.

(a) Assume that the linear regression with Gaussian noise holds, i.e., $Y = \boldsymbol{\beta}^{*T}\boldsymbol{X} + \epsilon$, with $\epsilon \sim N(0,1)$. Then, with probability at least $1 - 2 \cdot d^{-6}$,

$$-\mathbf{v}^T \nabla^2 \ell(\boldsymbol{\beta}^*)\mathbf{v} \geq \rho \cdot \|\mathbf{v}\|_2^2 - \tau \cdot \|\mathbf{v}\|_1^2 \cdot \frac{\log d}{k}, \quad \text{where } \rho = C_R \cdot C_1' \cdot C_R' \cdot \lambda_{\min}(\boldsymbol{\Sigma}_x), \ \tau = 4 \cdot C_\eta, \quad \text{(C.5)}$$

and $k = \lfloor n/2 \rfloor$. Here, $C_1'$ ia an absolute positive constant, $C_R' = \exp(-2 \cdot R^2)$ and

$$C_\eta = 32 \cdot C_R \cdot m^2, \quad \text{with } C_R = \frac{\exp(-4 \cdot R)}{[1 + \exp(4 \cdot R)]^2}, \quad \text{(C.6)}$$

where $R$ is a constant satisfying

$$C_1'' \cdot C_x^2 \cdot \exp\left(-\frac{C_1''' \cdot R^2}{C_\beta^2 \cdot C_x^2}\right) \leq \lambda_{\min}(\boldsymbol{\Sigma}_x),$$

for some absolute positive constants $C_1''$ and $C_1'''$. In addition, with probability at least $1 - 2 \cdot d^{-6}$,

$$-\mathbf{v}^T \nabla^2 \ell(\boldsymbol{\beta}^*)\mathbf{v} \geq \rho' \cdot \|\mathbf{v}\|_2^2, \quad \text{where } \rho' = C_R \cdot C_1' \cdot C_R' \cdot \lambda_{\min}(\boldsymbol{\Sigma}_x) - 64 \cdot C_\eta \cdot s \cdot \sqrt{\frac{\log d}{k}}. \quad \text{(C.7)}$$

(b) Assume that the logistic regression for $Y$ holds, i.e., $\mathbb{P}(Y = 0 \mid \boldsymbol{X}) = [1 + \exp(\boldsymbol{\beta}^{*T}\boldsymbol{X})]^{-1}$, and $\mathbb{P}(Y = 1 \mid \boldsymbol{X}) = 1 - \mathbb{P}(Y = 0 \mid \boldsymbol{X})$. Then, with probability at least $1 - 2 \cdot d^{-6}$, (C.5) and (C.7) hold with $C_1' = 1$, $C_R' = 2 \cdot \exp(-R) \cdot [1 + \exp(R)]^{-2}$ and $C_\eta$ is defined in (C.6).

(c) Assume that the Poisson regression, $p(y \mid \boldsymbol{x}) = \exp[y \cdot \boldsymbol{\beta}^{*T}\boldsymbol{x} - \exp(\boldsymbol{\beta}^{*T}\boldsymbol{x})]/y!$, holds. Then, with probability at least $1 - 2 \cdot d^{-6}$, (C.5) and (C.7) hold with $C_1' = 1$, $C_R' = 2 \cdot \exp[-R - 2 \cdot \exp(R)]$ and $C_\eta$ is defined in (C.6).

*Proof.* The detailed proof is shown in Supplementary Appendix E. □

## D  Proofs for Extensions to Missing Data and Selection Bias

Similar to the previous section, the following two Lemmas are sufficient to imply the validity of $\|\widehat{\boldsymbol{\beta}}_m - \boldsymbol{\beta}^*\|_1 = \mathcal{O}_{\mathbb{P}}(s \cdot \sqrt{\log d/n})$. The proof of Corollary 6.1 is similar to that of Theorem 4.1 and Theorem 4.2, and we omit the proof.

In the first Lemma, it shows that $\|\nabla \ell^m(\boldsymbol{\beta}^*)\|_\infty \leq C''' \cdot \sqrt{\log d/n}$ with high probability. In the second Lemma, it shows that the assumption C.1 for $\ell^m(\boldsymbol{\beta}^*)$ holds with high probability. These two Lemmas yield that the estimator $\widehat{\boldsymbol{\beta}}_m$ has the desired convergence rates with missing data and selection bias.

**Lemma D.1.** Assume that assumption 4.1 holds. Then, for any $C''' > 0$, we have $\|\nabla \ell^m(\boldsymbol{\beta}^*)\|_\infty \leq C''' \cdot \sqrt{\frac{\log d}{n}}$, with probability at least

$$1 - 2 \cdot d \cdot \exp\left[-\min\left\{\frac{C^2 \cdot C''^2}{2^9 \cdot C'^2 \cdot m^2} \cdot \frac{\log d}{n}, \frac{C \cdot C''}{2^5 \cdot C' \cdot m} \cdot \sqrt{\frac{\log d}{n}}\right\} \cdot k\right], \quad \text{(D.1)}$$

where $k = \lfloor n/2 \rfloor$, $m = \max_{1 \leq i \leq n} \max_{1 \leq j \leq d} |x_{ij}|$, and $C, C'$ are defined in assumption 4.1.

*Proof.* The detailed proof is shown in Supplementary Appendix E. □

**Lemma D.2.** Let the mean and covariance of $\boldsymbol{x}_i$ be 0 and $\boldsymbol{\Sigma}_x = \text{Cov}(\boldsymbol{x}_i)$ and denote $m = \max_{1 \leq i \leq n} \max_{1 \leq j \leq d} |x_{ij}|$. Assume that $\boldsymbol{x}_i$ is a sub-Gaussian vector with the finite sub-Gaussian norm denoted by $C_x$, and also assume $\|\boldsymbol{\beta}^*\|_2 \leq C_\beta$ for some finite constant $C_\beta$.

(a) Assume that the linear regression with Gaussian noise holds, i.e., $Y = \boldsymbol{\beta}^{*T}\boldsymbol{X} + \epsilon$, with $\epsilon \sim N(0,1)$. Assume that there exists an interval $I \supseteq [-1,1]$ such that for any $y \in I$ satisfies $g_1(y) > c$ for some constant $c > 0$ and $g_2(\boldsymbol{x})$ is a positive constant, where $g_1$ and $g_2$ are given in definition 2.3. Then, with probability at least $1 - 2d^{-6}$,

$$-\mathbf{v}^T \nabla^2 \ell^m(\boldsymbol{\beta}^*)\mathbf{v} \geq \rho \cdot \|\mathbf{v}\|_2^2 - \tau \cdot \|\mathbf{v}\|_1^2 \cdot \frac{\log d}{k}, \quad \text{where } \rho = C_R \cdot C_1' \cdot C_R' \cdot \lambda_{\min}(\boldsymbol{\Sigma}_x), \ \tau = 4 \cdot C_\eta, \quad \text{(D.2)}$$

and $k = \lfloor n/2 \rfloor$. Here, $C_1'$ ia an absolute positive constant, $C_R' = \exp(-2 \cdot R^2)$ and

$$C_\eta = 32 \cdot C_R \cdot m^2, \quad \text{with } C_R = \frac{\exp(-4 \cdot R)}{[1 + \exp(4 \cdot R)]^2}, \quad \text{(D.3)}$$

where $R$ is a constant satisfying

$$C_1'' \cdot C_x^2 \cdot \exp\left(-\frac{C_1''' \cdot R^2}{C_\beta^2 \cdot C_x^2}\right) \leq \lambda_{\min}(\boldsymbol{\Sigma}_x).$$

for some absolute positive constants $C_1''$ and $C_1'''$.

(b) Assume that the logistic regression for $Y$ holds, i.e., $\mathbb{P}(Y = 0 \mid \boldsymbol{X}) = [1 + \exp(\boldsymbol{\beta}^{*T}\boldsymbol{X})]^{-1}$, and $\mathbb{P}(Y = 1 \mid \boldsymbol{X}) = 1 - \mathbb{P}(Y = 0 \mid \boldsymbol{X})$. Assume that $g_1(0) > c$ and $g_1(1) > c$ for some constant $c > 0$. Then, with probability at least $1 - 2d^{-6}$, (D.2) holds with $C_1' = 1$, $C_R' = 2 \cdot c^2 \cdot \exp(-R) \cdot [1 + \exp(R)]^{-2}$ and $C_\eta$ is defined in (D.3).

(c) Assume that the Poisson regression, $p(y \mid \boldsymbol{x}) = \exp[y \cdot \boldsymbol{\beta}^{*T}\boldsymbol{x} - \exp(\boldsymbol{\beta}^{*T}\boldsymbol{x})]/y!$. Assume that there exists two positive integers $z_1$ and $z_2$, $z_1 \neq z_2$ satisfying $g_1(z_1) > c$ and $g_1(z_2) > c$ for some constant $c > 0$. Then, with probability at least $1 - 2d^{-6}$, (D.2) holds with $C_1' = 1$,

$$C_R' = c^2 \cdot (z_1 - z_2)^2 \cdot \frac{1}{z_1! z_2!} \cdot \exp\left\{-z_1 \cdot R - z_2 \cdot R - 2 \cdot \exp(R)\right\}, \quad \text{and}$$

$$C_\eta = 32 \cdot C_R \cdot m^2 \cdot [\max\{z_1, z_2\}]^2, \quad \text{with } C_R = \frac{\exp(-4 \cdot R \cdot \max\{z_1, z_2\})}{[1 + \exp(4 \cdot R \cdot \max\{z_1, z_2\})]^2}.$$

*Proof.* The detailed proof is shown in Supplementary Appendix E. □

## E  Proofs of Auxiliary Lemmas

In this appendix, we present the proofs of the auxiliary Lemmas in previous appendix.

## E.1 Proof of Lemma A.1

We first introduce the following intermediate estimator,

$$\widehat{\boldsymbol{\Sigma}}(\boldsymbol{\beta}) = \frac{1}{n(n-1)^2} \cdot \sum_{i=1}^{n} \sum_{j \neq i, k \neq i} \mathbf{r}_{ijk}(\boldsymbol{\beta}),$$

where the kernel function $\mathbf{r}_{ijk}(\boldsymbol{\beta})$ is defined as

$$\mathbf{r}_{ijk}(\boldsymbol{\beta}) = \frac{R_{ij}(\boldsymbol{\beta}) \cdot R_{ik}(\boldsymbol{\beta}) \cdot (y_i - y_j) \cdot (y_i - y_k) \cdot (\boldsymbol{x}_i - \boldsymbol{x}_j)(\boldsymbol{x}_i - \boldsymbol{x}_k)^T}{(1 + R_{ij}(\boldsymbol{\beta})) \cdot (1 + R_{ik}(\boldsymbol{\beta}))}.$$

We have $\widehat{\boldsymbol{\Sigma}} = \widehat{\boldsymbol{\Sigma}}(\widehat{\boldsymbol{\beta}})$ and the following decomposition,

$$\widehat{\boldsymbol{\Sigma}}(\widehat{\boldsymbol{\beta}}) - \boldsymbol{\Sigma} = \underbrace{\{\widehat{\boldsymbol{\Sigma}}(\widehat{\boldsymbol{\beta}}) - \widehat{\boldsymbol{\Sigma}}(\boldsymbol{\beta}^*)\}}_{I_1} + \underbrace{\{\widehat{\boldsymbol{\Sigma}}(\boldsymbol{\beta}^*) - \boldsymbol{\Sigma}\}}_{I_2}.$$

To control $I_1$, we will bound the derivative of $\widehat{\boldsymbol{\Sigma}}(\boldsymbol{\beta})$ with respect to $\boldsymbol{\beta}$. In particular, for any $(a, b)$ element of $\mathbf{r}_{ijk}(\boldsymbol{\beta})$ and any $1 \leq l \leq d$, after some simple algebra, we can show that

$$\left| \frac{\partial [\mathbf{r}_{ijk}(\boldsymbol{\beta})]_{(a,b)}}{\partial \beta_l} \right| \leq M^2 \cdot \left| \frac{\partial}{\partial \beta_l} \left( \frac{R_{ij}(\boldsymbol{\beta}) \cdot R_{ik}(\boldsymbol{\beta})}{(1 + R_{ij}(\boldsymbol{\beta})) \cdot (1 + R_{ik}(\boldsymbol{\beta}))} \right) \right|$$
$$\leq M^3 \cdot \frac{R_{ij}(\boldsymbol{\beta}) \cdot R_{ik}(\boldsymbol{\beta}) \cdot (1 + R_{ij}(\boldsymbol{\beta})) + R_{ij}(\boldsymbol{\beta}) \cdot R_{ik}(\boldsymbol{\beta}) \cdot (1 + R_{ik}(\boldsymbol{\beta}))}{(1 + R_{ij}(\boldsymbol{\beta}))^2 \cdot (1 + R_{ik}(\boldsymbol{\beta}))^2} \leq 2M^3.$$

Thus, by the mean value theorem and the assumption $\|\widehat{\boldsymbol{\beta}} - \boldsymbol{\beta}^*\|_1 = \mathcal{O}_{\mathbb{P}}(s \cdot \sqrt{\log d / n})$,

$$[\mathbf{r}_{ijk}(\widehat{\boldsymbol{\beta}}) - \mathbf{r}_{ijk}(\boldsymbol{\beta}^*)]_{(a,b)} = \frac{\partial [\mathbf{r}_{ijk}(\boldsymbol{\beta})]_{(a,b)}}{\partial \boldsymbol{\beta}} (\widehat{\boldsymbol{\beta}} - \boldsymbol{\beta}^*)$$
$$\leq \left\| \frac{\partial [\mathbf{r}_{ijk}(\boldsymbol{\beta})]_{(a,b)}}{\partial \boldsymbol{\beta}} \right\|_{\infty} \cdot \|\widehat{\boldsymbol{\beta}} - \boldsymbol{\beta}^*\|_1 = \mathcal{O}_{\mathbb{P}}\left(M^3 \cdot s \cdot \sqrt{\frac{\log d}{n}}\right),$$

where $\boldsymbol{\beta}$ lies in between $\widehat{\boldsymbol{\beta}}$ and $\boldsymbol{\beta}^*$. Thus, this implies

$$\|I_1\|_{\infty} = \mathcal{O}_{\mathbb{P}}\left(M^3 \cdot s \cdot \sqrt{\frac{\log d}{n}}\right). \tag{E.1}$$

By the definition of $\boldsymbol{\Sigma}$ and $\mathbf{h}_{ij|i}$ in (B.9), we have

$$\boldsymbol{\Sigma} = \frac{1}{n(n-1)^2} \cdot \sum_{i=1}^{n} \sum_{j \neq i, k \neq i, k \neq j} \mathbb{E}(\mathbf{h}_{ij|i} \mathbf{h}_{ik|i}^T) + \frac{1}{n(n-1)^2} \cdot \sum_{i=1}^{n} \sum_{j \neq i} \mathbb{E}(\mathbf{h}_{ij|i}^{\otimes 2})$$
$$= \frac{1}{n(n-1)^2} \cdot \sum_{i=1}^{n} \sum_{j \neq i, k \neq i, k \neq j} \mathbb{E}(\mathbf{h}_{ij} \mathbf{h}_{ik}^T) + \frac{1}{n(n-1)^2} \cdot \sum_{i=1}^{n} \sum_{j \neq i} \mathbb{E}(\mathbf{h}_{ij|i}^{\otimes 2}),$$

where the last step follows from the fact that $\mathbf{h}_{ij}$ and $\mathbf{h}_{ik}$ are independent given $y_i, \boldsymbol{x}_i$. Since $\mathbf{r}_{ijk}(\boldsymbol{\beta}^*) = \mathbf{h}_{ij}\mathbf{h}_{ik}^T$, we have

$$I_2 = \frac{n-2}{n-1} \cdot \underbrace{\left\{\frac{1}{n(n-1)(n-2)} \sum_{j\neq i, k\neq i, j\neq k} \left[r_{ijk}(\boldsymbol{\beta}^*) - \mathbb{E}\{r_{ijk}(\boldsymbol{\beta}^*)\}\right]\right\}}_{I_{21}}$$

$$+ \frac{1}{n-1} \cdot \underbrace{\left\{\frac{1}{n(n-1)} \sum_{j\neq i} \left[r_{ijj}(\boldsymbol{\beta}^*) - \mathbb{E}\{r_{ijj}(\boldsymbol{\beta}^*)\}\right]\right\}}_{I_{22}} + \frac{1}{n-1} \cdot \underbrace{\left\{\frac{1}{n(n-1)} \sum_{j\neq i} \left[\mathbb{E}\{\mathbf{h}_{ij}^{\otimes 2}\} - \mathbb{E}\{\mathbf{h}_{ij|i}^{\otimes 2}\}\right]\right\}}_{I_{23}}.$$

It is seen that $I_{21}$ is a mean-zero third order U-statistic. Note that $\mathbf{r}_{ijk}(\boldsymbol{\beta}^*)$ satisfies $[\mathbf{r}_{ijk}(\boldsymbol{\beta}^*)]_{(a,b)} \leq M^2$. The Hoeffding inequality yields that for any $(a,b)$ element of $I_{21}$,

$$\mathbb{P}\left(\left|[I_{21}]_{(a,b)}\right| > x\right) \leq 2 \cdot \exp\left(-\frac{k \cdot x^2}{8 \cdot M^4}\right),$$

where $k = \lfloor n/3 \rfloor$. By the union bound inequality,

$$\|I_{21}\|_\infty = \mathcal{O}_\mathbb{P}\left(M^2 \cdot \sqrt{\frac{\log d}{n}}\right).$$

Similarly, $I_{22}$ is a mean-zero second order U-statistic with the kernel function $\mathbf{r}_{ijj}(\boldsymbol{\beta}^*)$ satisfying $[\mathbf{r}_{ijj}(\boldsymbol{\beta}^*)]_{(a,b)} \leq M^2$. By using the same arguments, we can show that

$$\|I_{22}\|_\infty = \mathcal{O}_\mathbb{P}\left(M^2 \cdot \sqrt{\frac{\log d}{n}}\right).$$

For $I_{23}$, note that $\|\mathbf{h}_{ij}\|_\infty \leq M$, which implies that $\|\mathbb{E}\mathbf{h}_{ij}^{\otimes 2}\|_\infty \leq M^2$. In addition, by the definition of $\mathbf{h}_{ij|i}$ in (B.9), we can show that $\|\mathbb{E}\mathbf{h}_{ij|i}^{\otimes 2}\|_\infty \leq M^2$. Hence, $\|I_{23}\|_\infty = \mathcal{O}_\mathbb{P}(M^2)$. Combining the error bounds for $I_{21}$, $I_{22}$ and $I_{23}$, we finally obtain

$$\|I_2\|_\infty = \mathcal{O}_\mathbb{P}\left(M^2 \cdot \sqrt{\frac{\log d}{n}}\right). \tag{E.2}$$

Combining the error bounds for $I_1$ and $I_2$ in (E.1) and (E.2), we can conclude the proof,

$$\|\widehat{\boldsymbol{\Sigma}}(\widehat{\boldsymbol{\beta}}) - \boldsymbol{\Sigma}\|_\infty = \mathcal{O}_\mathbb{P}\left(M^3 \cdot s \cdot \sqrt{\frac{\log d}{n}}\right) + \mathcal{O}_\mathbb{P}\left(M^2 \cdot \sqrt{\frac{\log d}{n}}\right) = \mathcal{O}_\mathbb{P}\left(M^3 \cdot s \cdot \sqrt{\frac{\log d}{n}}\right).$$

### E.2 Proof of Proposition C.1

Denote $F_{ij} = \{|y_i| \leq \eta\} \cap \{|y_j| \leq \eta\}$ and $F'_{ij} = \{|\boldsymbol{\beta}^{*T}\boldsymbol{x}_i| \leq R\} \cap \{|\boldsymbol{\beta}^{*T}\boldsymbol{x}_j| \leq R\}$, where $\eta$ and $R$ are positive constants to be chosen later. We first apply a truncation argument for the Hessian matrix.

$$-\nabla^2 \ell(\boldsymbol{\beta}^*) \geq \frac{2}{n(n-1)} \cdot \sum_{i<j} \frac{R_{ij}(\boldsymbol{\beta}^*) \cdot (y_i - y_j)^2 \cdot (\boldsymbol{x}_i - \boldsymbol{x}_j)^{\otimes 2}}{(1 + R_{ij}(\boldsymbol{\beta}^*))^2} \cdot \mathbb{1}(F_{ij}) \cdot \mathbb{1}(F'_{ij})$$

$$\geq \frac{2 \cdot C_R}{n(n-1)} \cdot \sum_{i<j} (y_i - y_j)^2 \cdot (\boldsymbol{x}_i - \boldsymbol{x}_j)^{\otimes 2} \cdot \mathbb{1}(F_{ij}) \cdot \mathbb{1}(F'_{ij}),$$

where $C_R = \exp(-4 \cdot R \cdot \eta) \cdot (1 + \exp(4 \cdot R \cdot \eta))^{-2}$. Consider the following U-statistic

$$\mathbf{W} = \frac{2 \cdot C_R}{n(n-1)} \cdot \sum_{i<j}(y_i - y_j)^2 \cdot (\boldsymbol{x}_i - \boldsymbol{x}_j)^{\otimes 2} \cdot \mathbb{1}(F_{ij}) \cdot \mathbb{1}(F'_{ij}).$$

We first verify assumption C.1. For any $\mathbf{v} \in \mathbb{R}^d$, we get

$$-\mathbf{v}^T \nabla^2 \ell(\boldsymbol{\beta}^*) \mathbf{v} \geq \mathbf{v}^T \mathbf{W} \mathbf{v} = \mathbf{v}^T \mathbb{E}(\mathbf{W}) \mathbf{v} + \mathbf{v}^T [\mathbf{W} - \mathbb{E}(\mathbf{W})] \mathbf{v}.$$

By the Hölder inequality, we get

$$|\mathbf{v}^T \mathbf{W} \mathbf{v} - \mathbf{v}^T \mathbb{E}(\mathbf{W}) \mathbf{v}| \leq ||\mathbf{v}||_1^2 \cdot ||\mathbf{W} - \mathbb{E}(\mathbf{W})||_\infty,$$

and it further implies that

$$-\mathbf{v}^T \nabla^2 \ell(\boldsymbol{\beta}^*) \mathbf{v} \geq \mathbf{v}^T \mathbb{E}(\mathbf{W}) \mathbf{v} - ||\mathbf{v}||_1^2 \cdot ||\mathbf{W} - \mathbb{E}(\mathbf{W})||_\infty. \tag{E.3}$$

Next, we establish the concentration of $\mathbf{W}$ to its mean. Note that after the truncation argument, the kernel function of $\mathbf{W}$ is bounded, i.e.,

$$||C_R \cdot (y_i - y_j)^2 \cdot (\boldsymbol{x}_i - \boldsymbol{x}_j)^{\otimes 2} \cdot \mathbb{1}(F_{ij})||_\infty \leq 16 \cdot C_R \cdot m^2 \cdot \eta^2.$$

The Hoeffding inequality can be applied to the centered U-statistic $W_{jk} - \mathbb{E}(W_{jk})$. For some constant $t > 0$ to be chosen, and any $1 \leq j, k \leq d$,

$$\mathbb{P}\Big(|W_{jk} - \mathbb{E}(W_{jk})| \geq t\Big) \leq 2 \cdot \exp\Big(-\frac{k \cdot t^2}{2 \cdot C_\eta^2}\Big).$$

where $k = \lfloor n/2 \rfloor$ and $C_\eta = 32 \cdot C_R \cdot m^2 \cdot \eta^2$. By the union bound inequality,

$$\mathbb{P}\Big(||\mathbf{W} - \mathbb{E}(\mathbf{W})||_\infty \geq t\Big) \leq \sum_{1 \leq j,k \leq d} \mathbb{P}\Big(|W_{jk} - \mathbb{E}(W_{jk})| \geq t\Big) \leq 2 \cdot d^2 \cdot \exp\Big(-\frac{k \cdot t^2}{2 \cdot C_\eta^2}\Big). \tag{E.4}$$

Taking $t = 4 \cdot C_\eta \cdot \sqrt{\log d/k}$, we obtain that $||\mathbf{W} - \mathbb{E}(\mathbf{W})||_\infty \leq 4 \cdot C_\eta \cdot \sqrt{\log d/k}$, with probability at least $1 - 2d^{-6}$. As seen from (E.3), it remains to find a lower bound for $\mathbf{v}^T \mathbb{E}(\mathbf{W}) \mathbf{v}$. In the following, we establish the lower bounds for three important generalized linear models, including linear regressions with Gaussian errors, logistic regressions and Poisson regressions.

**Linear model:** If $y$ follows from the linear model with $N(0,1)$ error, with $\eta = 1$ under $F'_{ij}$, we get

$$\mathbb{E}\big[(y_i - y_j)^2 \cdot \mathbb{1}\{|y_i| \leq 1, |y_j| \leq 1\} \mid \boldsymbol{x}_i, \boldsymbol{x}_j\big]$$
$$= \frac{1}{\sqrt{2\pi}} \cdot \int_{-1}^{1} \int_{-1}^{1} (y_i - y_j)^2 \cdot \exp\Big[-\frac{(y_i - \boldsymbol{\beta}^{*T}\boldsymbol{x}_i)^2 + (y_j - \boldsymbol{\beta}^{*T}\boldsymbol{x}_j)^2}{2}\Big] \cdot dy_i dy_j$$
$$\geq \frac{1}{\sqrt{2\pi}} \cdot \int_{-1}^{1} \int_{-1}^{1} (y_i - y_j)^2 \cdot \exp[-(y_i^2 + y_j^2) - 2 \cdot R^2] \cdot dy_i dy_j. \tag{E.5}$$

For notational simplicity, we let

$$C'_1 = \frac{1}{\sqrt{2\pi}} \cdot \int_{-1}^{1} \int_{-1}^{1} (y_i - y_j)^2 \cdot \exp[-(y_i^2 + y_j^2)] \cdot dy_i dy_j, \quad \text{and} \quad C'_R = \exp(-2 \cdot R^2). \tag{E.6}$$

By (E.5), we have,
$$\mathbf{v}^T \mathbb{E}(\mathbf{W})\mathbf{v} = \mathbf{v}^T \mathbb{E}\big[\mathbb{E}(\mathbf{W} \mid \mathbf{x})\big]\mathbf{v} \geq C_1' \cdot C_R \cdot C_R' \cdot \mathbf{v}^T \mathbb{E}[(\boldsymbol{x}_i - \boldsymbol{x}_j)^{\otimes 2} \cdot \mathbb{1}(F_{ij}')]\mathbf{v}.$$

Let $F_{ij}^{'c}$ be the complement of $F_{ij}'$. The Cauchy inequality yields,
$$\begin{aligned}
\mathbf{v}^T \mathbb{E}[(\boldsymbol{x}_i - \boldsymbol{x}_j)^{\otimes 2} \cdot \mathbb{1}(F_{ij}^{'c})]\mathbf{v} &\leq \mathbb{E}[2(\mathbf{v}^T\boldsymbol{x}_i)^2 \cdot \mathbb{1}(F_{ij}^{'c}) + 2(\mathbf{v}^T\boldsymbol{x}_j)^2 \cdot \mathbb{1}(F_{ij}^{'c})] \\
&= 4 \cdot \mathbb{E}[(\mathbf{v}^T\boldsymbol{x}_i)^2 \cdot \mathbb{1}(F_{ij}^{'c})].
\end{aligned}$$

Since $\boldsymbol{x}_i$ is sub-Gaussian and let $\mathbf{u} = \mathbf{v}/\|\mathbf{v}\|_2$, we further have
$$\begin{aligned}
\mathbf{v}^T \mathbb{E}[(\boldsymbol{x}_i - \boldsymbol{x}_j)^{\otimes 2} \cdot \mathbb{1}(F_{ij}^{'c})]\mathbf{v} &\leq 4 \cdot \|\mathbf{v}\|_2^2 \cdot \mathbb{E}[(\mathbf{u}^T\boldsymbol{x}_i)^2 \cdot \mathbb{1}(F_{ij}^{'c})] \leq 4 \cdot \|\mathbf{v}\|_2^2 \cdot \sqrt{\mathbb{E}(\mathbf{u}^T\boldsymbol{x}_i)^4} \cdot \sqrt{\mathbb{P}(F_{ij}^{'c})} \\
&\leq 16\sqrt{2} \cdot \|\mathbf{v}\|_2^2 \cdot C_x^2 \cdot \sqrt{\mathbb{P}(|\boldsymbol{\beta}^{*T}\boldsymbol{x}_i| > R)} \\
&\leq C_1'' \cdot \|\mathbf{v}\|_2^2 \cdot C_x^2 \cdot \exp\Big(-\frac{C_1''' \cdot R^2}{C_\beta^2 \cdot C_x^2}\Big),
\end{aligned}$$
where $C_1''$ and $C_1'''$ are absolute positive constants. Now, we choose $R$ such that
$$C_1'' \cdot C_x^2 \cdot \exp\Big(-\frac{C_1''' \cdot R^2}{C_\beta^2 \cdot C_x^2}\Big) \leq \lambda_{\min}(\boldsymbol{\Sigma}_x).$$

Thus, $\mathbf{v}^T \mathbb{E}[(\boldsymbol{x}_i - \boldsymbol{x}_j)^{\otimes 2} \cdot \mathbb{1}(F_{ij}^{'c})]\mathbf{v} \leq \lambda_{\min}(\boldsymbol{\Sigma}_x) \cdot \|\mathbf{v}\|_2^2$, which implies that
$$\begin{aligned}
\mathbf{v}^T \mathbb{E}(\mathbf{W})\mathbf{v} &\geq C_1' \cdot C_R \cdot C_R' \cdot \Big\{\mathbf{v}^T \mathbb{E}(\boldsymbol{x}_i - \boldsymbol{x}_j)^{\otimes 2}\mathbf{v} - \mathbf{v}^T \mathbb{E}[(\boldsymbol{x}_i - \boldsymbol{x}_j)^{\otimes 2} \cdot \mathbb{1}(F_{ij}^{'c})]\mathbf{v}\Big\} \\
&\geq C_1' \cdot C_R \cdot C_R' \cdot \lambda_{\min}(\boldsymbol{\Sigma}_x) \cdot \|\mathbf{v}\|_2^2.
\end{aligned} \qquad (E.7)$$

By (E.3), (E.4) and (E.7), we finally obtain, with probability at least $1 - 2d^{-6}$,
$$-\mathbf{v}^T \nabla^2 \ell(\boldsymbol{\beta}^*)\mathbf{v} \geq \rho \cdot \|\mathbf{v}\|_2^2 - \tau \cdot \|\mathbf{v}\|_1^2 \cdot \frac{\log d}{k}, \qquad (E.8)$$
where $\rho = C_R \cdot C_1' \cdot C_R' \cdot \lambda_{\min}(\boldsymbol{\Sigma}_x)$ and $\tau = 4 \cdot C_\eta$.

**Logistic model:** If $y$ given $\boldsymbol{x}$ follows from the logistic regression, one can take $\eta = 1$ in above proof, since $|y| \leq 1$. In this case, (E.5) reduces to
$$\begin{aligned}
&\mathbb{E}\big[(y_i - y_j)^2 \cdot \mathbb{1}\{|y_i| \leq 1, |y_j| \leq 1\} \mid \boldsymbol{x}_i, \boldsymbol{x}_j\big] \\
={}& \mathbb{P}(y_i = 1 \mid \boldsymbol{x}_i) \cdot \mathbb{P}(y_j = 0 \mid \boldsymbol{x}_j) + \mathbb{P}(y_i = 0 \mid \boldsymbol{x}_i) \cdot \mathbb{P}(y_j = 1 \mid \boldsymbol{x}_j) \\
={}& \frac{\exp(\boldsymbol{\beta}^{*T}\boldsymbol{x}_i) + \exp(\boldsymbol{\beta}^{*T}\boldsymbol{x}_j)}{[1 + \exp(\boldsymbol{\beta}^{*T}\boldsymbol{x}_i)] \cdot [1 + \exp(\boldsymbol{\beta}^{*T}\boldsymbol{x}_j)]} \geq C_R', \quad \text{where } C_R' = \frac{2 \cdot \exp(-R)}{[1 + \exp(R)]^2}.
\end{aligned} \qquad (E.9)$$

Hence, following the same arguments, we can establish (E.8) with $\rho = C_R \cdot C_R' \cdot \lambda_{\min}(\boldsymbol{\Sigma}_x)$ and $\tau = 4 \cdot C_\eta$. Here, $C_R'$ is redefined in (E.9).

**Poisson model:** If $y$ given $\boldsymbol{x}$ follows from the Poisson regression, with $\eta = 1$, similarly, we can get
$$\begin{aligned}
&\mathbb{E}\big[(y_i - y_j)^2 \cdot \mathbb{1}\{|y_i| \leq 1, |y_j| \leq 1\} \mid \boldsymbol{x}_i, \boldsymbol{x}_j\big] \\
={}& \exp\big[\boldsymbol{\beta}^T\boldsymbol{x}_j - \exp(\boldsymbol{\beta}^T\boldsymbol{x}_i) - \exp(\boldsymbol{\beta}^T\boldsymbol{x}_j)\big] + \exp\big[y_i\boldsymbol{\beta}^T\boldsymbol{x}_i - \exp(\boldsymbol{\beta}^T\boldsymbol{x}_i) - \exp(\boldsymbol{\beta}^T\boldsymbol{x}_j)\big] \\
\geq{}& C_R', \quad \text{where } C_R' = 2 \cdot \exp[-R - 2 \cdot \exp(R)].
\end{aligned} \qquad (E.10)$$

Hence, following the same arguments, we can establish (E.8) with $\rho = C_R \cdot C_R' \cdot \lambda_{\min}(\Sigma_x)$ and $\tau = 4 \cdot C_\eta$. Here, $C_R'$ is redefined in (E.10).

Next, we will verify assumption C.2. For any $\mathbf{v} \in \mathbb{R}^d$ and $\|\mathbf{v}_{S^c}\|_1 \leq 3 \cdot \|\mathbf{v}_S\|_1$, by the Cauchy inequality, (E.3) further implies,

$$\begin{aligned} -\mathbf{v}^T \nabla^2 \ell(\boldsymbol{\beta}^*) \mathbf{v} &\geq \mathbf{v}^T \mathbb{E}(\mathbf{W}) \mathbf{v} - 16 \cdot \|\mathbf{v}_S\|_1^2 \cdot \|\mathbf{W} - \mathbb{E}(\mathbf{W})\|_\infty \\ &\geq \mathbf{v}^T \mathbb{E}(\mathbf{W}) \mathbf{v} - 16 \cdot s \cdot \|\mathbf{v}\|_2^2 \cdot \|\mathbf{W} - \mathbb{E}(\mathbf{W})\|_\infty. \end{aligned} \quad (E.11)$$

Recall that, by (E.4), we obtain that $\|\mathbf{W} - \mathbb{E}(\mathbf{W})\|_\infty \leq 4 \cdot C_\eta \cdot \sqrt{\log d/n}$, with probability at least $1 - d^{-6}$. Similar to the proof of (E.8), after some algebra, it is easy to show that, for the Gaussian model, with probability at least $1 - 2d^{-6}$,

$$-\mathbf{v}^T \nabla^2 \ell(\boldsymbol{\beta}^*) \mathbf{v} \geq \rho' \cdot \|\mathbf{v}\|_2^2, \quad \text{where} \quad \rho' = C_R \cdot C_1' \cdot C_R' \cdot \lambda_{\min}(\Sigma_x) - 64 \cdot C_\eta \cdot s \cdot \sqrt{\log d/k}, \quad (E.12)$$

where $C_R'$ is defined in (E.6). Similarly, for the logistic model, (E.12) holds with $C_1' = 1$ and $C_R'$ defined in (E.9). For the Poisson model, (E.12) holds with $C_1' = 1$ and $C_R'$ defined in (E.10).

### E.3 Proof of Lemma D.1

Note that $\nabla \ell^m(\boldsymbol{\beta}) = -\frac{2}{n(n-1)} \cdot \sum_{1 \leq i < j \leq n} \mathbf{h}_{mij}(\boldsymbol{\beta})$, where

$$\mathbf{h}_{mij}(\boldsymbol{\beta}) = -\frac{\delta_i \cdot \delta_j \cdot R_{ij}(\boldsymbol{\beta}) \cdot (y_i - y_j) \cdot (\boldsymbol{x}_i - \boldsymbol{x}_j)}{1 + R_{ij}(\boldsymbol{\beta})}. \quad (E.13)$$

Denote $\Xi_{ij} = \{(Y_{(i)}^L, Y_{(j)}^L) = (y_i, y_j), \boldsymbol{X}_i = \boldsymbol{x}_i, \boldsymbol{X}_j = \boldsymbol{x}_j\}$. We can show that $\mathbb{E}\{\mathbf{h}_{mij}(\boldsymbol{\beta}^*) \mid \Xi_{ij}, \delta_i = \delta_j = 1\} = 0$. Since $\mathbf{h}_{mij}(\boldsymbol{\beta}^*) = 0$ if either $\delta_i = 0$ or $\delta_j = 0$, we have $\mathbb{E}\{\mathbf{h}_{mij}(\boldsymbol{\beta}^*) \mid (\delta_i, \delta_j) = (a, b)\} = 0$ if $a = 0$ or $b = 0$. These together imply that

$$\begin{aligned} \mathbb{E}(\mathbf{h}_{mij}(\boldsymbol{\beta}^*)) &= \mathbb{E}(\mathbf{h}_{mij}(\boldsymbol{\beta}^*) \mid \delta_i = 1, \delta_j = 1) \cdot \mathbb{P}(\delta_i = 1, \delta_j = 1) \\ &+ \mathbb{E}(\mathbf{h}_{mij}(\boldsymbol{\beta}^*) \mid \delta_i = 0, \delta_j = 1) \cdot \mathbb{P}(\delta_i = 0, \delta_j = 1) \\ &+ \mathbb{E}(\mathbf{h}_{mij}(\boldsymbol{\beta}^*) \mid \delta_i = 1, \delta_j = 0) \cdot \mathbb{P}(\delta_i = 1, \delta_j = 0) \\ &+ \mathbb{E}(\mathbf{h}_{mij}(\boldsymbol{\beta}^*) \mid \delta_i = 0, \delta_j = 0) \cdot \mathbb{P}(\delta_i = 0, \delta_j = 0) = 0. \end{aligned}$$

Since $R_{ij}(\boldsymbol{\beta}) > 0$ and $\max_{ij} |x_{ij}| \leq m$, we have

$$\|\mathbf{h}_{mij}(\boldsymbol{\beta}^*)\|_\infty \leq 2 \cdot m \cdot |y_i - y_j|.$$

By the sub-exponential tail condition on $y_i$, for any $x > 0$ and $k = 1, ..., d$,

$$\mathbb{P}\big(|[\mathbf{h}_{mij}(\boldsymbol{\beta}^*)]_k| > x\big) \leq \mathbb{P}\big(|y_i - y_j| > (2m)^{-1} \cdot x\big) \leq 2 \cdot C' \cdot \exp\{-C \cdot (4m)^{-1} \cdot x\}.$$

Then we apply Lemma 5.3 with $k = \lfloor n/2 \rfloor$,

$$\begin{aligned} \mathbb{P}\big(\|\nabla \ell^m(\boldsymbol{\beta}^*)\|_\infty > C'' \cdot \sqrt{\log d/n}\big) &\leq \sum_{k=1}^d \mathbb{P}\big(|\nabla_k \ell^m(\boldsymbol{\beta}^*)| > C'' \cdot \sqrt{\log d/n}\big) \\ &\leq 2 \cdot d \cdot \exp\Big[-\min\Big\{\frac{C^2 \cdot C''^2}{2^9 \cdot C'^2 \cdot m^2} \cdot \frac{k \cdot \log d}{n}, \frac{C \cdot C'' \cdot k}{2^5 \cdot C' \cdot m} \cdot \sqrt{\frac{\log d}{n}}\Big\}\Big]. \end{aligned}$$

### E.4 Proof of Lemma D.2

Denote $F_{ij} = \{y_i : y_i \in [\eta_1, \eta_2]\} \cap \{y_j : y_j \in [\eta_1, \eta_2]\}$, where $[\eta_1, \eta_2] \subseteq I$. Denote $\eta = \max(|\eta_1|, |\eta_2|)$, where $\eta_1$ and $\eta_2$ are positive constants to be chosen later. Also denote $F'_{ij} = \{|\boldsymbol{\beta}^{*T}\boldsymbol{x}_i| \leq R\} \cap \{|\boldsymbol{\beta}^{*T}\boldsymbol{x}_j| \leq R\}$, where $R$ is a constant to be chosen later. We first apply a truncation argument to the Hessian matrix,

$$
\begin{aligned}
-\nabla^2 \ell^m(\boldsymbol{\beta}^*) &\geq \frac{2}{n(n-1)} \cdot \sum_{i<j} \frac{\delta_i \cdot \delta_j \cdot R_{ij}(\boldsymbol{\beta}^*) \cdot (y_i - y_j)^2 \cdot (\boldsymbol{x}_i - \boldsymbol{x}_j)^{\otimes 2}}{(1 + R_{ij}(\boldsymbol{\beta}^*))^2} \cdot \mathbb{1}(F_{ij}) \cdot \mathbb{1}(F'_{ij}) \\
&\geq \frac{2 \cdot C_R}{n(n-1)} \cdot \sum_{i<j} \delta_i \cdot \delta_j \cdot (y_i - y_j)^2 \cdot (\boldsymbol{x}_i - \boldsymbol{x}_j)^{\otimes 2} \cdot \mathbb{1}(F_{ij}) \cdot \mathbb{1}(F'_{ij}),
\end{aligned}
$$

where $C_R = \exp(-4 \cdot R \cdot \eta) \cdot (1 + \exp(4 \cdot R \cdot \eta))^{-2}$. Consider the following U-statistic

$$
\mathbf{W} = \frac{2 \cdot C_R}{n(n-1)} \cdot \sum_{i<j} \delta_i \cdot \delta_j \cdot (y_i - y_j)^2 \cdot (\boldsymbol{x}_i - \boldsymbol{x}_j)^{\otimes 2} \cdot \mathbb{1}(F_{ij}) \cdot \mathbb{1}(F'_{ij}).
$$

We now verify assumption C.1. For any $\mathbf{v} \in \mathbb{R}^d$, we get

$$
-\mathbf{v}^T \nabla^2 \ell^m(\boldsymbol{\beta}^*) \mathbf{v} \geq \mathbf{v}^T \mathbf{W} \mathbf{v} = \mathbf{v}^T \mathbb{E}(\mathbf{W}) \mathbf{v} + \mathbf{v}^T [\mathbf{W} - \mathbb{E}(\mathbf{W})] \mathbf{v}.
$$

By the Hölder inequality, it further implies that

$$
-\mathbf{v}^T \nabla^2 \ell^m(\boldsymbol{\beta}^*) \mathbf{v} \geq \mathbf{v}^T \mathbb{E}(\mathbf{W}) \mathbf{v} - ||\mathbf{v}||_1^2 \cdot ||\mathbf{W} - \mathbb{E}(\mathbf{W})||_\infty. \tag{E.14}
$$

Next, we establish the concentration of $\mathbf{W}$ to its mean. Note that after the truncation argument, the kernel function of $\mathbf{W}$ is bounded, i.e.,

$$
||C_R \cdot \delta_i \cdot \delta_j \cdot (y_i - y_j)^2 \cdot (\boldsymbol{x}_i - \boldsymbol{x}_j)^{\otimes 2} \cdot \mathbb{1}(F_{ij})||_\infty \leq 16 \cdot C_R \cdot m^2 \cdot \eta^2.
$$

The Hoeffding inequality can be applied to the centered U-statistic $W_{jk} - \mathbb{E}(W_{jk})$. For some constant $t > 0$ to be chosen, and any $1 \leq j, k \leq d$,

$$
\mathbb{P}\Big(|W_{jk} - \mathbb{E}(W_{jk})| \geq t\Big) \leq 2 \cdot \exp\Big(-\frac{k \cdot t^2}{2 \cdot C_\eta^2}\Big).
$$

where $k = \lfloor n/2 \rfloor$ and $C_\eta = 32 \cdot C_R \cdot m^2 \cdot \eta^2$. By the union bound inequality,

$$
\mathbb{P}\Big(||\mathbf{W} - \mathbb{E}(\mathbf{W})||_\infty \geq t\Big) \leq \sum_{1 \leq j, k \leq d} \mathbb{P}\Big(|W_{jk} - \mathbb{E}(W_{jk})| \geq t\Big) \leq 2 \cdot d^2 \cdot \exp\Big(-\frac{k \cdot t^2}{2 \cdot C_\eta^2}\Big). \tag{E.15}
$$

Taking $t = 4 \cdot C_\eta \cdot \sqrt{\log d/k}$, we obtain that $||\mathbf{W} - \mathbb{E}(\mathbf{W})||_\infty \leq 4 \cdot C_\eta \cdot \sqrt{\log d/k}$, with probability at least $1 - 2d^{-6}$. In the following, we establish the lower bounds for three important generalized linear models, including linear regressions with Gaussian errors, logistic regressions and Poisson regressions.

**Linear model:** If $y$ follows from the linear model with $N(0,1)$ error, with $\eta = 1$, we get

$$\mathbb{E}\big[g_1(y_i) \cdot g_1(y_j) \cdot (y_i - y_j)^2 \cdot \mathbb{1}\{|y_i| \leq 1, |y_j| \leq 1\} \mid \boldsymbol{x}_i, \boldsymbol{x}_j\big]$$
$$= \frac{1}{\sqrt{2\pi}} \cdot \int_{-1}^{1}\int_{-1}^{1} g_1(y_i) \cdot g_1(y_j) \cdot (y_i - y_j)^2 \cdot \exp\Big[-\frac{(y_i - \boldsymbol{\beta}^{*T}\boldsymbol{x}_i)^2 + (y_j - \boldsymbol{\beta}^{*T}\boldsymbol{x}_j)^2}{2}\Big] \cdot dy_i dy_j$$
$$\geq \frac{1}{\sqrt{2\pi}} \cdot \int_{-1}^{1}\int_{-1}^{1} g_1(y_i) \cdot g_1(y_j) \cdot (y_i - y_j)^2 \cdot \exp[-(y_i^2 + y_j^2) - 2 \cdot R^2] \cdot dy_i dy_j. \tag{E.16}$$

For notational simplicity, we let

$$C_1' = \frac{1}{\sqrt{2\pi}} \cdot \int_{-1}^{1}\int_{-1}^{1} g_1(y_i) \cdot g_1(y_j) \cdot (y_i - y_j)^2 \cdot \exp[-(y_i^2 + y_j^2)] \cdot dy_i dy_j, \text{ and } C_R' = \exp(-2 \cdot R^2). \tag{E.17}$$

Similar to the proof of Proposition C.1, by (E.5), we can show that,

$$\mathbf{v}^T \mathbb{E}(\mathbf{W}) \mathbf{v} \geq C_R \cdot C_1' \cdot C_R' \cdot \lambda_{\min}(\boldsymbol{\Sigma}_x) \cdot \|\mathbf{v}\|_2^2, \tag{E.18}$$

where $R$ is chosen such that for some absolute positive constants $C_1''$ and $C_1'''$,

$$C_1'' \cdot C_x^2 \cdot \exp\Big(-\frac{C_1''' \cdot R^2}{C_\beta^2 \cdot C_x^2}\Big) \leq \lambda_{\min}(\boldsymbol{\Sigma}_x).$$

By (E.14), (E.15) and (E.18), we finally obtain, with probability at least $1 - 2d^{-6}$,

$$-\mathbf{v}^T \nabla^2 \ell^m(\boldsymbol{\beta}^*) \mathbf{v} \geq \rho \cdot \|\mathbf{v}\|_2^2 - \tau \cdot \|\mathbf{v}\|_1^2 \cdot \frac{\log d}{k}, \tag{E.19}$$

where $\rho = C_R \cdot C_1' \cdot C_R' \cdot \lambda_{\min}(\boldsymbol{\Sigma}_x)$ and $\tau = 4 \cdot C_\eta$.

**Logistic model:** If $y$ given $\boldsymbol{x}$ follows from the logistic regression, one can take $\eta = 1$ in above proof, since $|y| \leq 1$. In this case, (E.16) reduces to

$$\mathbb{E}\Big[g_1(y_i) \cdot g_1(y_j) \cdot (y_i - y_j)^2 \cdot \mathbb{1}\{|y_i| \leq 1, |y_j| \leq 1\} \mid \boldsymbol{x}_i, \boldsymbol{x}_j\Big]$$
$$= c^2 \Big\{\mathbb{P}(y_i = 1 \mid \boldsymbol{x}_i) \cdot \mathbb{P}(y_j = 0 \mid \boldsymbol{x}_j) + \mathbb{P}(y_i = 0 \mid \boldsymbol{x}_i) \cdot \mathbb{P}(y_j = 1 \mid \boldsymbol{x}_j)\Big\}$$
$$= \frac{c^2\{\exp(\boldsymbol{\beta}^{*T}\boldsymbol{x}_i) + \exp(\boldsymbol{\beta}^{*T}\boldsymbol{x}_j)\}}{[1 + \exp(\boldsymbol{\beta}^{*T}\boldsymbol{x}_i)] \cdot [1 + \exp(\boldsymbol{\beta}^{*T}\boldsymbol{x}_j)]} \geq C_R', \quad \text{where } C_R' = \frac{2 \cdot c^2 \cdot \exp(-R)}{[1 + \exp(R)]^2}. \tag{E.20}$$

Hence, following the same arguments, we can establish (E.19) with $\rho = C_R \cdot C_R' \cdot \lambda_{\min}(\boldsymbol{\Sigma}_x)$ and $\tau = 4 \cdot C_\eta$. Here, $C_R'$ is redefined in (E.20).

**Poisson model:** If $y$ given $\boldsymbol{x}$ follows from the Poisson regression, taking $F_{ij} = \{y_i = z_1, y_j = z_2\}$, we can get

$$\mathbb{E}\big[g_1(y_i) \cdot g_1(y_j) \cdot (y_i - y_j)^2 \cdot \mathbb{1}(F_{ij}) \mid \boldsymbol{x}_i, \boldsymbol{x}_j\big]$$
$$= g_1(z_1) \cdot g_1(z_2) \cdot (z_1 - z_2)^2 \cdot \frac{1}{z_1! z_2!} \cdot \exp\Big\{z_1 \cdot \boldsymbol{\beta}^T \boldsymbol{x}_i + z_2 \cdot \boldsymbol{\beta}^T \boldsymbol{x}_j - \exp(\boldsymbol{\beta}^T \boldsymbol{x}_i) - \exp(\boldsymbol{\beta}^T \boldsymbol{x}_j)\Big\}$$
$$\geq C_R', \quad \text{where } C_R' = c^2 \cdot (z_1 - z_2)^2 \cdot \frac{1}{z_1! z_2!} \cdot \exp\Big\{-z_1 \cdot R - z_2 \cdot R - 2 \cdot \exp(R)\Big\}. \tag{E.21}$$

Hence, following the same arguments, we can establish (E.19) with $\rho = C_R \cdot C_R' \cdot \lambda_{\min}(\boldsymbol{\Sigma}_x)$ and $\tau = 4 \cdot C_\eta$. Here, $C_R'$ is redefined in (E.21).

# F  Extensions to Misspecified Models

In this section, we examine the theoretical properties of the penalized estimator $\widehat{\boldsymbol{\beta}}$ and the directional likelihood ratio test under model misspecification. Let $p^*(y \mid \boldsymbol{x})$ denote the true probability density function of $Y$ given $\boldsymbol{X}$. Recall that $\mathbf{R}_{ij}^L$ and $(Y_{(i)}^L, Y_{(j)}^L)$ denote the local rank and order statistics of $Y_i$ and $Y_j$. The conditional probability of $\mathbf{R}_{ij}^L = \boldsymbol{r}_{ij}^L$ given $(Y_{(i)}^L, Y_{(j)}^L)$ and $(\boldsymbol{X}_i, \boldsymbol{X}_j)$ is

$$\mathbb{P}^*(\mathbf{R}_{ij}^L = \boldsymbol{r}_{ij}^L \mid y_{(i)}^L, y_{(j)}^L, \boldsymbol{x}_i, \boldsymbol{x}_j) = \frac{p^*(y_i \mid \boldsymbol{x}_i) \cdot p^*(y_j \mid \boldsymbol{x}_j)}{p^*(y_i \mid \boldsymbol{x}_i) \cdot p^*(y_j \mid \boldsymbol{x}_j) + p^*(y_i \mid \boldsymbol{x}_j) \cdot p^*(y_j \mid \boldsymbol{x}_i)}.$$

Thus, the true pairwise rank log-likelihood is defined as

$$\ell^* = \frac{2}{n(n-1)} \cdot \sum_{1 \leq i < j \leq n} \log \mathbb{P}^*(\mathbf{R}_{ij}^L = \boldsymbol{r}_{ij}^L \mid y_{(i)}^L, y_{(j)}^L, \boldsymbol{x}_i, \boldsymbol{x}_j).$$

Define the rank Kullback-Leibler divergence (RKL) between the assumed model and the true model as

$$\text{RKL}(\ell^*, \ell(\boldsymbol{\beta})) = \mathbb{E}^*\{-\ell(\boldsymbol{\beta}) + \ell^*\},$$

where $\mathbb{E}^*(\cdot)$ represents the expectation under the true density $p^*(y \mid \boldsymbol{x})$. In addition, if the samples $(Y_1, \boldsymbol{X}_1), ..., (Y_n, \boldsymbol{X}_n)$ are i.i.d, the rank Kullback-Leibler divergence reduces to

$$\text{RKL}(\ell^*, \ell(\boldsymbol{\beta})) = \mathbb{E}^*\left\{ \log \frac{\mathbb{P}^*(\mathbf{R}_{ij}^L = \boldsymbol{r}_{ij}^L \mid y_{(i)}^L, y_{(j)}^L, \boldsymbol{x}_i, \boldsymbol{x}_j)}{\mathbb{P}(\mathbf{R}_{ij}^L = \boldsymbol{r}_{ij}^L \mid y_{(i)}^L, y_{(j)}^L, \boldsymbol{x}_i, \boldsymbol{x}_j; \boldsymbol{\beta})} \right\},$$

where $\mathbb{P}(\mathbf{R}_{ij}^L = \boldsymbol{r}_{ij}^L \mid y_{(i)}^L, y_{(j)}^L, \boldsymbol{x}_i, \boldsymbol{x}_j; \boldsymbol{\beta}) = (1 + R_{ij}(\boldsymbol{\beta}))^{-1}$ with $R_{ij}(\boldsymbol{\beta})$ given by (3.4). Hence, $\text{RKL}(\ell^*, \ell(\boldsymbol{\beta}))$ measures the distance between the rank based likelihood under the true model and the assumed model. Let $\boldsymbol{\beta}^o$ denote the oracle parameter (i.e., least false parameter) that minimizes the rank Kullback-Leibler divergence, i.e., $\boldsymbol{\beta}^o = \arg\min_{\boldsymbol{\beta}} \text{RKL}(\ell^*, \ell(\boldsymbol{\beta}))$. We assume that $\boldsymbol{\beta}^o$ is unique. Once an estimator for the oracle parameter $\boldsymbol{\beta}^o$ is given, one can follow the similar idea in Section 3 to construct hypothesis tests for $H_0^o : \alpha^o = 0$ under the misspecified model. The results are summarized in the following theorem. For notational simplicity, we use the same notations for $\boldsymbol{\Sigma}$ and $\mathbf{H}$ in this theorem. For instance, $\mathbf{H}$ and $\boldsymbol{\Sigma}$ are indeed defined as $\mathbf{H} = -\mathbb{E}^*\{\nabla^2 \ell(\boldsymbol{\beta}^o)\}$ and $\boldsymbol{\Sigma} = \mathbb{E}^*\{\mathbf{g}(y_i, \boldsymbol{x}_i, \boldsymbol{\beta}^o)^{\otimes 2}\}$, where the expectations are evaluated under the true distribution and $\boldsymbol{\beta}^*$ is replaced by $\boldsymbol{\beta}^o$, and $\mathbf{w}^o$ is defined similarly.

**Theorem F.1.** Under the Assumptions 4.1, 4.2, 4.3 and 4.4 with $\boldsymbol{\beta}^*$ replaced by $\boldsymbol{\beta}^o$, we have

$$n^{1/2} \cdot (\widehat{\alpha}^P - \alpha^*) \rightsquigarrow N(0, 4 \cdot \sigma^2 \cdot H_{\alpha|\gamma}^{-2}), \text{ and } (4 \cdot \sigma^2)^{-1} \cdot H_{\alpha|\gamma} \cdot \Lambda_n \rightsquigarrow \chi_1^2,$$

where $\sigma^2 = \boldsymbol{\Sigma}_{\alpha\alpha} - 2\mathbf{w}^{oT}\boldsymbol{\Sigma}_{\gamma\alpha} + \mathbf{w}^{oT}\boldsymbol{\Sigma}_{\gamma\gamma}\mathbf{w}^o$ and $H_{\alpha|\gamma} = H_{\alpha\alpha} - \mathbf{H}_{\alpha\gamma}\mathbf{H}_{\gamma\gamma}^{-1}\mathbf{H}_{\gamma\alpha}$.

The proof is similar to Theorem 4.1 and the details are omitted. This result can be viewed as the extension of the general theory of misspecified low dimensional models considered in White (1982) to high dimensional models.

# G   Higher Order Statistical Chromatography

In this appendix, we derive the high order chromatography. For $k = 2, .., n$, given $k$ random variables $(Y_{i_1}, ..., Y_{i_k})$, define $\mathbf{R}^L_{i_1,...,i_k}$ and $\mathbf{O}^L_{i_1,...,i_k}$ to be the local rank and order statistics of $Y_{i_1}, ..., Y_{i_k}$ among $(Y_{i_1}, ..., Y_{i_k})$. Thus

$$\mathbb{P}(\mathbf{R}^L_{i_1,...,i_k} = r^L_{i_1,...,i_k} \mid \boldsymbol{x}_{i_1}, ..., \boldsymbol{x}_{i_k}, \boldsymbol{o}^L_{i_1,...,i_k}; \boldsymbol{\beta}) = \frac{\prod_{j=1}^k p(y_{i_j} \mid \boldsymbol{x}_{i_j}; \boldsymbol{\beta}, f)}{\sum_{(j_1,...,j_k) \in \Phi_k} \prod_{t=1}^k p(y_{j_t} \mid \boldsymbol{x}_{i_t}; \boldsymbol{\beta}, f)}$$

$$= \frac{\exp(\sum_{j=1}^k \boldsymbol{\beta}^T \boldsymbol{x}_{i_j} \cdot y_{i_j})}{\sum_{(j_1,...,j_k) \in \Phi_k} \exp(\sum_{t=1}^k \boldsymbol{\beta}^T \boldsymbol{x}_{i_t} \cdot y_{j_t})},$$

where $r^L_{i_1,...,i_k}, \boldsymbol{o}^L_{i_1,...,i_k}$ are the observed values of $\mathbf{R}^L_{i_1,...,i_k}$ and $\mathbf{O}^L_{i_1,...,i_k}$ and $\Phi_k$ is the set of all permutations of $(i_1, ..., i_k)$. Hence,

$$\prod_{(i_1,...,i_k) \in \boldsymbol{\Delta}_k} \mathbb{P}(\mathbf{R}^L_{i_1,...,i_k} = r^L_{i_1,...,i_k} \mid \boldsymbol{x}_{i_1}, ..., \boldsymbol{x}_{i_k}, \boldsymbol{o}^L_{i_1,...,i_k}; \boldsymbol{\beta}) = \prod_{(i_1,...,i_k) \in \boldsymbol{\Delta}_k} \frac{\exp(\sum_{j=1}^k \boldsymbol{\beta}^T \boldsymbol{x}_{i_j} \cdot y_{i_j})}{\sum_{(j_1,...,j_k) \in \Phi_k} \exp(\sum_{t=1}^k \boldsymbol{\beta}^T \boldsymbol{x}_{i_t} \cdot y_{j_t})},$$

where $\boldsymbol{\Delta}_k$ is the combination of $k$ indexes chosen from $\{1, ..., n\}$. The likelihood function based on the $k$th order information is given by

$$\ell_k(\boldsymbol{\beta}) = \binom{n}{k}^{-1} \log\left\{ \prod_{(i_1,...,i_k) \in \boldsymbol{\Delta}_k} \mathbb{P}(\mathbf{R}^L_{i_1,...,i_k} = r^L_{i_1,...,i_k} \mid \boldsymbol{x}_{i_1}, ..., \boldsymbol{x}_{i_k}, \boldsymbol{o}^L_{i_1,...,i_k}; \boldsymbol{\beta}) \right\}$$

$$= -\binom{n}{k}^{-1} \sum_{(i_1,...,i_k) \in \boldsymbol{\Delta}_k} \log\left( D_{i_1,...,i_k}(\boldsymbol{\beta}) \right), \tag{G.1}$$

where

$$D_{i_1,...,i_k}(\boldsymbol{\beta}) = \sum_{(j_1,...,j_k) \in \Phi_k} \exp\left\{ \sum_{t=1}^k \boldsymbol{\beta}^T \boldsymbol{x}_{i_t} \cdot (y_{j_t} - y_{i_t}) \right\}.$$

In particular, when $k = 3$, we have

$$\ell_3(\boldsymbol{\beta}) = -\binom{n}{3}^{-1} \sum_{(i,j,k) \in \boldsymbol{\Delta}_3} \log\{1 + Q_{ijk}(\boldsymbol{\beta})\}, \tag{G.2}$$

where

$$Q_{ijk}(\boldsymbol{\beta}) = R_{ij}(\boldsymbol{\beta}) + R_{ik}(\boldsymbol{\beta}) + R_{jk}(\boldsymbol{\beta}) + \widetilde{R}_{ij,jk,ki}(\boldsymbol{\beta}) + \widetilde{R}_{ik,ji,kj}(\boldsymbol{\beta}).$$

Here,

$$R_{ij}(\boldsymbol{\beta}) = \exp\left\{ -(y_i - y_j) \cdot \boldsymbol{\beta}^T(\boldsymbol{x}_i - \boldsymbol{x}_j) \right\}, \quad \text{and}$$

$$\widetilde{R}_{ij,jk,ki}(\boldsymbol{\beta}) = \exp\left\{ \boldsymbol{\beta}^T \boldsymbol{x}_i \cdot (y_j - y_i) + \boldsymbol{\beta}^T \boldsymbol{x}_j \cdot (y_k - y_j) + \boldsymbol{\beta}^T \boldsymbol{x}_k \cdot (y_i - y_k) \right\}.$$

Following the similar device, we can construct the $k$th order directional likelihood function $\widehat{\ell}_k(\alpha) = \ell_k(\alpha, \widehat{\boldsymbol{\gamma}}_k + (\widehat{\alpha}_k - \alpha)\widehat{\mathbf{w}}_k)$, where $(\widehat{\alpha}_k, \widehat{\boldsymbol{\gamma}}_k) := \widehat{\boldsymbol{\beta}}_k$, $\widehat{\boldsymbol{\beta}}_k$ is a generic estimator, say $\widehat{\boldsymbol{\beta}}_k = \arg\max \ell_k(\boldsymbol{\beta}) - \sum_{j=1}^d p_\lambda(\beta_j)$, and

$$\widehat{\mathbf{w}}_k = \arg\min ||\mathbf{w}||_1 \quad \text{subject to} \quad ||\nabla^2_{\alpha\gamma}\ell_k(\widehat{\boldsymbol{\beta}}_k) - \mathbf{w}^T \nabla^2_{\gamma\gamma}\ell_k(\widehat{\boldsymbol{\beta}}_k)||_\infty \leq \lambda_{ks}.$$

It is seen that the gradient and Hessian matrix of $\ell_k(\boldsymbol{\beta})$ has the structure of $k$th order U-statistics. Similar to the proof of Theorem 4.1, by using the U-statistic theory such as Lemma 5.3, we can prove the asymptotic distributions of $\widehat{\alpha}_k^P$ and $\Lambda_n^k$ which are defined in the similar manner as the ordinary version. Although the same idea can be applied, the derivation is lengthy. We leave the rigorous development of the $k$th order chromatography theory for the future investigation.

## H  Analysis of Cancer Data

We also apply the proposed tests to analyze the data from the United Kingdom Ovarian Cancer Population Study (UKOPS) to infer differentially methylated loci between ovarian cancer cases and age matched healthy controls. The data are collected by Illumina Infinium Human Methylation27 Beadchip (Teschendorff et al., 2010). In this dataset, 96 cancer samples and 136 normal samples with methylation $\beta$-values on $22{,}951$ loci are observed, which yields a total of $n = 232$ samples with $22{,}951$ covariates. Our aim is to identify the important loci that are related to ovarian cancer.

Similar to the previous example, the original data have no missing values. We apply the desparsifying method based on the logistic regression, and the proposed Wald and likelihood ratio tests to the dataset. The same adjustment method for p-values is adopted. We find that the three methods find the same locus cg20792833. This suggests that this locus is strongly related to the ovarian cancer, which is validated by all three methods. We also examine the effect of missingness on this cancer dataset. Specifically, we generate the selection indicator $\delta_i$ for $Y_i$ by $\mathbb{P}(\delta_i = 1 \mid Y_i, \boldsymbol{X}_i) = 1 - CY_i$, for $C = 0, 0.1$ and $0.2$. This means we keep all the normal samples ($Y_i = 0$) and randomly remove the outcome variable for $C \times 100\%$ cancer samples ($Y_i = 1$). Here, $C = 0$ means no missing data is created. The results are shown in Table H.1. Similar to the previous example, in the presence of missing data, the complete-case analysis (CC-Desparsity) is too conservative and cannot identify any significant gene, and the imputation method (Imp-Desparsity) is heavily influenced by the proportion of missingness. It is seen that the proposed likelihood method can still identify the potentially important locus cg20792833 when $C = 0.1$, which shows its advantage over the competing methods. Nevertheless, it also becomes less powerful when the proprotion of missingness increases (say, $C = 0.2$).

Table H.1: Significant loci selected by the Wald and directional likelihood ratio tests, and the desparsifying method based on complete-case analysis (CC-) and imputation (Imp-) for the cancer data.

| C | Wald | DLRT | CC-Desparsity | Imp-Desparsity |
|---|---|---|---|---|
| 0 | cg20792833 | cg20792833 | cg20792833 | cg20792833 |
| 0.1 | - | cg20792833 | - | cg10636246, cg20792833 |
| 0.2 | - | - | - | - |